\documentclass{notreks_arxiv}

\providecommand{\notreksauthor}[1]{\clearauthor{#1}}

\usepackage{amsmath, amssymb}
\usepackage{thmtools}
\usepackage{xcolor}
\usepackage{verbatim}

\usepackage{booktabs}

\usepackage{graphicx}

\usepackage{diffcoeff}
\usepackage{xcolor}

\usepackage{tikz}

\usepackage[titletoc]{appendix}

\usepackage{thmtools, thm-restate} 

\usepackage{hyperref}
\usepackage[capitalise]{cleveref}

\newcommand{\E}{\mathbb{E}}
\renewcommand{\P}{\mathbb{P}}

\newcommand{\R}{\mathbb{R}}

\renewcommand{\cal}[1]{\mathcal{#1}}

\newcommand{\var}{\mathrm{Var}}

\newcommand{\abs}[1]{\left| #1 \right|}
\newcommand{\norm}[1]{\left\| #1 \right\|}

\newcommand{\inner}[2]{\langle #1, #2 \rangle}

\newcommand{\grad}{\nabla}

\newcommand{\eps}{\varepsilon}

\renewcommand{\set}[1]{\{ #1 \}}

\newcommand{\pa}[1]{\operatorname{pa}_G(#1) }
\newcommand{\ch}[1]{\operatorname{ch}_G(#1) }
\newcommand{\an}[1]{\operatorname{an}_G(#1) }
\newcommand{\de}[1]{\operatorname{de}_G(#1) }

\newcommand{\indep}{\perp\!\!\!\!\perp}

\crefname{definition}{Definition}{Definitions}
\Crefname{definition}{Definition}{Definitions}
\crefname{theorem}{Theorem}{Theorems}
\Crefname{theorem}{Theorem}{Theorems}
\crefname{corollary}{Corollary}{Corollaries}
\Crefname{corollary}{Corollary}{Corollaries}
\crefname{lemma}{Lemma}{Lemmas}
\Crefname{lemma}{Lemma}{Lemmas}
\crefname{proposition}{Proposition}{Propositions}
\Crefname{proposition}{Proposition}{Propositions}
\crefname{remark}{Remark}{Remarks}
\Crefname{remark}{Remark}{Remarks}
\crefname{figure}{Figure}{Figures}
\Crefname{figure}{Figure}{Figures}
\crefname{table}{Table}{Tables}
\Crefname{table}{Table}{Tables}

\crefname{section}{Section}{Sections}
\Crefname{section}{Section}{Sections}
\crefname{subsection}{Section}{Sections}
\Crefname{subsection}{Section}{Sections}
\crefname{subsubsection}{Section}{Sections}
\Crefname{subsubsection}{Section}{Sections}

\crefname{appendix}{Appendix}{Appendices}
\Crefname{appendix}{Appendix}{Appendices}
\crefname{subappendix}{Appendix}{Appendices}
\Crefname{subappendix}{Appendix}{Appendices}
\crefname{subsubappendix}{Appendix}{Appendices}
\Crefname{subsubappendix}{Appendix}{Appendices}
\crefname{subsubsubappendix}{Appendix}{Appendices}
\Crefname{subsubsubappendix}{Appendix}{Appendices}

\crefformat{equation}{(#2#1#3)}  

\definecolor{notrekpurple}{HTML}{8C33C0}

\SetKwInOut{Input}{Input}
\SetKw{ReturnKw}{return}

\title[NOTREKS: Marginal Independence in Score-based Structure Learning]{NOTREKS: Marginal Independence in Score-based Structure Learning}
\makeatletter
\let\savedtitle\@title
\makeatother
\usepackage{times}

\notreksauthor{%
 \Name{Fabian Bleile} \Email{f.bleile@tum.de}\\
 \addr Technical University of Munich, Germany\\
Munich Center for Machine Learning, Munich, Germany
 \AND
 \Name{Sarah Lumpp} \Email{sarah.lumpp@tum.de}\\
 \addr Technical University of Munich, Germany
 \AND
 \Name{Mathias Drton} \Email{mathias.drton@tum.de}\\
 \addr Technical University of Munich, Germany\\
Munich Center for Machine Learning, Munich, Germany
}

\begin{document}

\maketitle

\begin{abstract}%
Score-based structure learning estimates causal graphs by optimizing a data-dependent objective over structural equation models. We study how prior knowledge that selected pairs of variables are marginally independent can be incorporated into such causal discovery methods. Under faithfulness, marginally independent pairs cannot be connected by a trek, so the knowledge becomes a graphical no-trek constraint, which we call NOTREKS information. We introduce a power-series trek (PSTrek) penalty for gradient-based optimization and a trek constraint for order-based search, and we analyze why such information helps in finite samples and how it restricts the search space. Our numerical experiments illustrate to what extent NOTREKS information helps in causal discovery. In particular, using synthetic benchmarks with oracle NOTREKS information, we show that these constraints can improve both structural recovery and adjustment-based effect identification.%
\end{abstract}

\begin{keywords}%
  Causal discovery, marginal independence, prior knowledge, score-based search, structure learning%
\end{keywords}

\section{Introduction}

Score-based structure learning formulates causal discovery as an optimization problem: given a family of structural equation models (SEMs), find the model that minimizes a data-dependent score \citep{Drton:Maathuis:2017,spirtes:handbook}. Under correct model specification and in the limit of infinite data, consistent score functions allow one to recover the data-generating structure (or its Markov equivalence class). In practice, however, with finite samples and possible model misspecification, structure learning benefits from additional inductive bias in the form of regularizers or structural constraints. In this work, we consider incorporating prior knowledge of marginal independence relations into gradient- and order-based causal discovery methods.

\begin{figure*}[htbp]
    \centering
    \tikzset{
        vertex/.style={
            circle,
            draw,
            minimum size=0.5cm,
            inner sep=0pt,
            font=\scriptsize
        },
        notrek/.style={
            draw=notrekpurple,
            line width=1.1pt
        }
    }
    \begin{tikzpicture}[->, >=stealth, scale=1.2]
        \begin{scope}[xshift=0.00cm]
            \node[vertex] (a1) at (-0.40, 0.45) {1};
            \node[vertex] (a2) at ( 0.40, 0.45) {2};
            \node[vertex] (a3) at (-0.40,-0.45) {3};
            \node[vertex] (a4) at ( 0.40,-0.45) {4};

            \draw (a1) -> (a3);
            \draw (a2) -> (a3);
            \draw (a2) -> (a4);
        \end{scope}
        \begin{scope}[xshift=2.40cm]
            \node[vertex] (b1) at (-0.40, 0.45) {1};
            \node[vertex] (b2) at ( 0.40, 0.45) {2};
            \node[vertex] (b3) at (-0.40,-0.45) {3};
            \node[vertex] (b4) at ( 0.40,-0.45) {4};

            \draw[notrek, -] (b1) -- (b2);
            \draw[notrek, -] (b1) -- (b4);
        \end{scope}
        \node[font=\small] at (0,-1.00) {\(G\)};
        \node[font=\small] at (2.40,-1.00) {\(H\)};
        \node[font=\small, anchor=west] at (-0.95,-1.00) {(a)};
    \end{tikzpicture}\hspace{0.08\linewidth}%
    \begin{tikzpicture}[->, >=stealth, scale=1.2]
        \begin{scope}[xshift=0.00cm]
            \node[vertex] (c1) at (-0.40, 0.45) {1};
            \node[vertex] (c2) at ( 0.40, 0.45) {2};
            \node[vertex] (c3) at (-0.40,-0.45) {3};
            \node[vertex] (c4) at ( 0.40,-0.45) {4};

            \draw (c1) -> (c3);
            \draw (c2) -> (c3);
            \draw (c4) -> (c2);
        \end{scope}
        \begin{scope}[xshift=2.00cm]
            \node[vertex] (d1) at (-0.40, 0.45) {1};
            \node[vertex] (d2) at ( 0.40, 0.45) {2};
            \node[vertex] (d3) at (-0.40,-0.45) {3};
            \node[vertex] (d4) at ( 0.40,-0.45) {4};

            \draw (d2) -> (d3);
            \draw (d2) -> (d4);
            \draw (d3) -> (d4);
        \end{scope}
        \begin{scope}[xshift=4.00cm]
            \node[vertex] (e1) at (-0.40, 0.45) {1};
            \node[vertex] (e2) at ( 0.40, 0.45) {2};
            \node[vertex] (e3) at (-0.40,-0.45) {3};
            \node[vertex] (e4) at ( 0.40,-0.45) {4};

            \draw (e3) -> (e1);
            \draw (e2) -> (e4);
        \end{scope}
        \node[font=\small, anchor=west] at (-0.95,-1.00) {(b) Some NOTREKS-feasible DAGs};
    \end{tikzpicture}

    \vspace{10pt}
    \begin{tikzpicture}
        \node[inner sep=0pt, anchor=south west] (plot) {\includegraphics[width=0.65\linewidth]{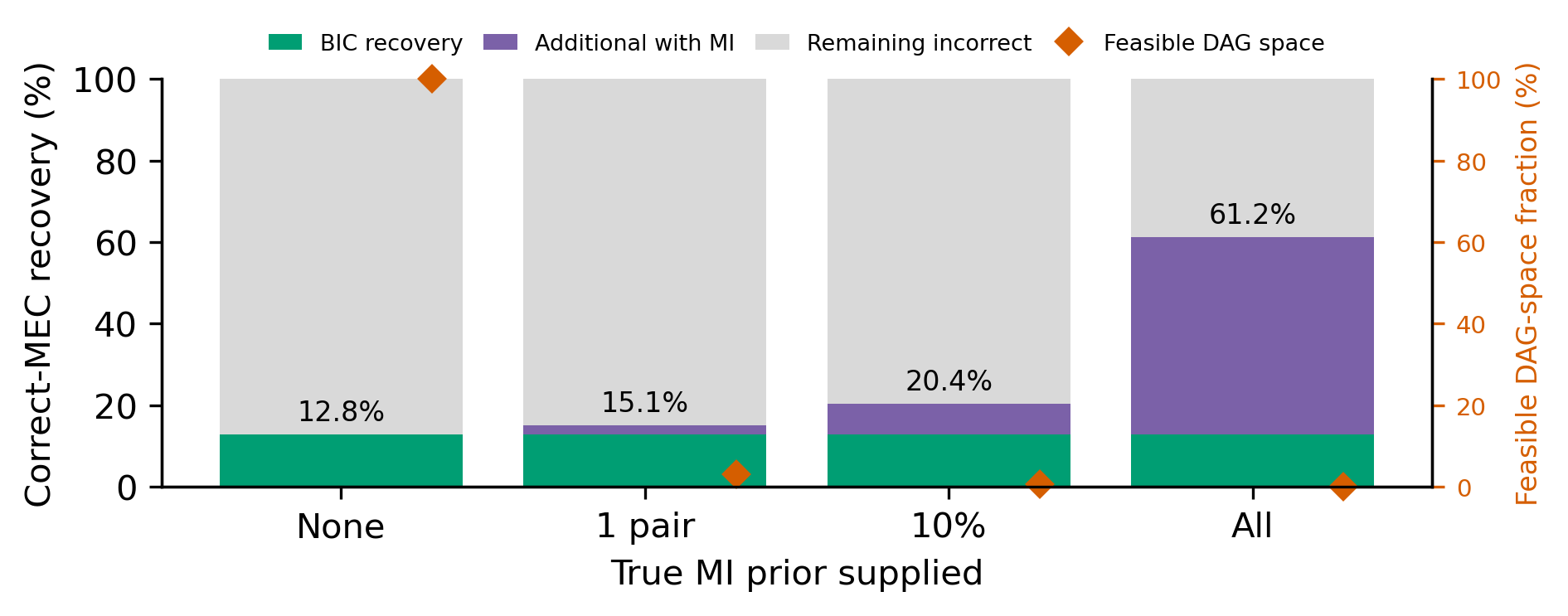}};
        \node[font=\small, anchor=south west, inner sep=0pt] at (plot.south west) {(c)};
    \end{tikzpicture}
    \caption{\textbf{(a)} True DAG $G$ and NOTREKS information $H$, encoding the marginal independencies $X_1 \indep X_2$ and $X_1 \indep X_4$. \textbf{(b)} DAGs that are NOTREKS-feasible (without treks between 1 and 2 or 1 and 4). \textbf{(c)} Exact MEC recovery when BIC is minimized globally over all DAGs, subject to increasing subsets of the true marginal independencies (linear Gaussian SEMs, $d=10$, $n=100$, averaged over five graph families; see \cref{app:notreks-motivation}).}
    \label{fig:motivation-notreks-feasible-dags}
\end{figure*}
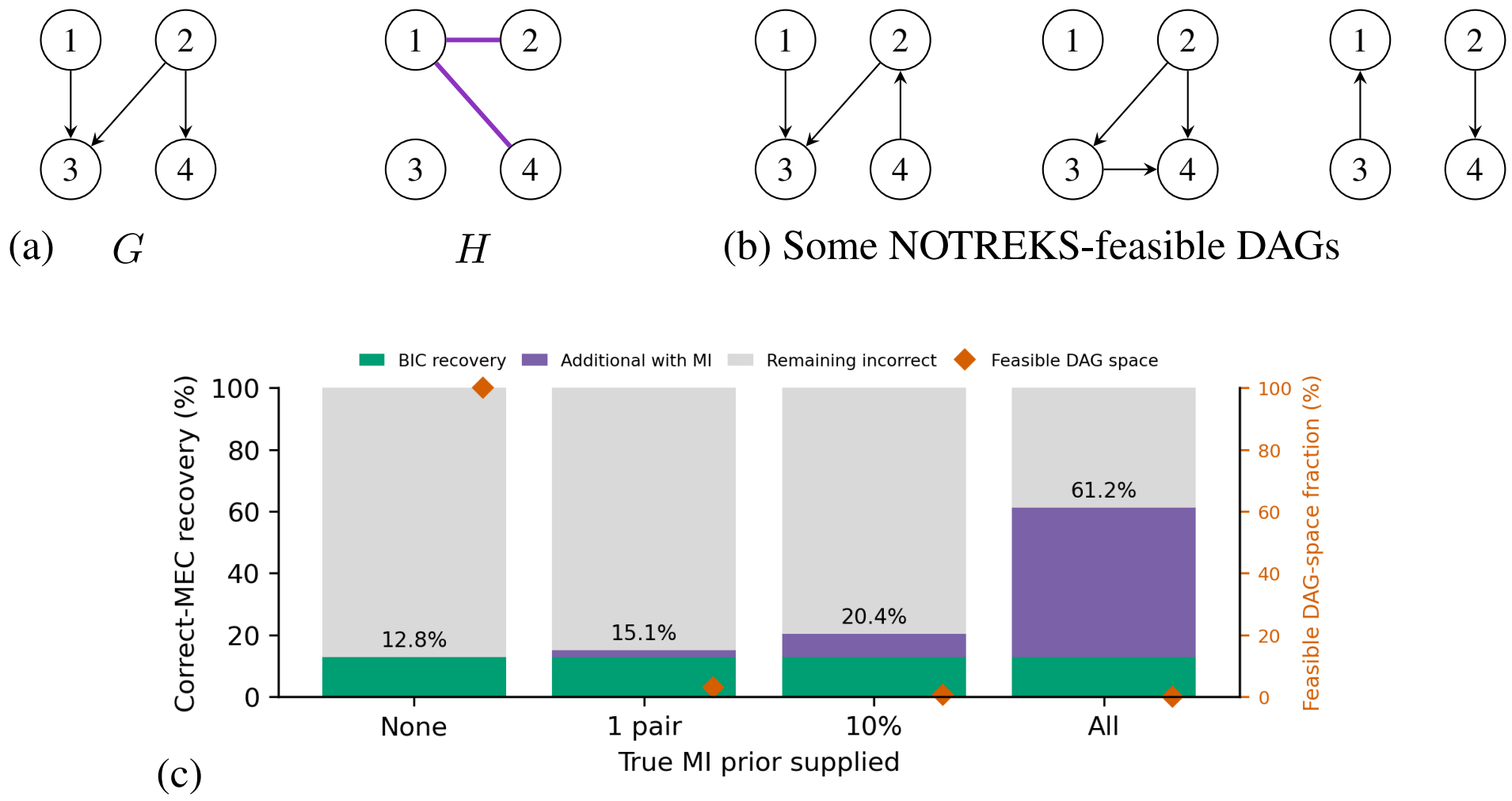

We see prior information on marginal independence as highly relevant for causal discovery, because such information could indeed be available from earlier observational studies that examined one or more pairs of variables on their own. Incorporating it into score-based methods is not redundant, since these methods do not enforce independence constraints explicitly.
The independencies implied by an estimated graph emerge only indirectly from optimizing the score and, with finite samples or under model misspecification, may contradict what is known.
In a low-dimensional setting, \cref{fig:motivation-notreks-feasible-dags}(c) shows that the true Markov equivalence class (MEC) is recovered more often when the BIC is minimized globally subject to correct marginal-independence constraints, so explicitly encoding this knowledge provides a useful inductive bias.

In this work, we introduce \emph{NOTREKS}, a strategy for enforcing known marginal independencies through the \emph{absence of treks} in a graph; see \cref{fig:motivation-notreks-feasible-dags}(a,b). Under a faithfulness assumption, marginal independence can be translated into this graphical constraint (\cref{subsec:Marginal Independence and Treks}). We propose extensions to DAGMA \citep{bello2022dagma} and FLOP \citep{wienöbst2026embracingdiscretesearchreasonable}, as representatives of gradient- and order-based causal discovery methods. For gradient-based methods, we propose the PSTrek penalty, which encodes the no-trek constraints as a differentiable function of the weight matrix, and we establish its theoretical properties. For order-based search, we restrict the parent search of FLOP to graphs that satisfy the supplied constraints. We further analyze the combinatorial consequences of no-trek constraints for the size of the search space.

We remark that in this work we do not consider constraint-based methods such as PC and FCI \citep{spirtes2000causation,zhang2008completeness}. Indeed, these methods can incorporate marginal-independence information directly, as corresponding independence tests can simply be skipped.

\subsection{Related Work}\label{subsec:Background and Related Work}

Prominent examples of score-based methods are combinatorial search over equivalence classes, as in GES \citep{chickering2002optimal}, and order-based search over topological sortings, as in BOSS \citep{andrews2023fast} and FLOP \citep{wienöbst2026embracingdiscretesearchreasonable}. Both exploit a decomposition of the score into local terms, one for each node and its parents.
Alternatives replace discrete structural constraints with smooth characterizations, enabling continuous, gradient-based optimization \citep{zheng2018dags, bello2022dagma}. Score-based approaches have also been developed for cyclic settings \citep{amendola2020structure,mooij2013cyclic,sethuraman2023nodags}.

Graphical prior knowledge can guide structure learning through edge, super-structure, or path constraints \citep{borboudakis2012incorporating,sinha2021using,clivio2025learningdefercausaldiscovery,ban2025differentiable,guo2026linear,de2025cluster}.
Power-series penalties turn structural properties into differentiable objectives, beginning with the acyclicity constraint in NOTEARS and other path constraints \citep{zheng2018dags,bello2022dagma,nazaret2023stable,lee2019scaling,ban2025differentiable,guo2026linear}. No-trek constraints are stronger than such path constraints, since the absence of a trek between $i$ and $j$ excludes directed paths in either direction as well as any common ancestor.

Marginal independencies carry substantial information about causal structure. \citet{pearl1994can} characterize when a set of marginal independencies can be represented faithfully by a DAG, and marginal and low-order conditional independencies have been used to characterize or learn graphs and their equivalence classes \citep{textor2015learning,wille2006low,wienobst2020recovering,markham2022transformational,deligeorgaki2024combinatorial}. In gradient-based structure learning, independence information has so far entered only locally, through support masks derived from the PC algorithm \citep{10.1093/bioinformatics/btae411} or by masking individual edges \citep{absar2024time}. These approaches do not generally enforce all implied no-trek constraints.

In this work we assume access to marginal-independence pairs. Alternatively, marginal independencies may also be estimated from data using statistical tests or methods for recovering unconditional dependence graphs \citep{shi:2022,wienobst2020recovering,deligeorgaki2024combinatorial,wille2006low}. Testing errors and data reuse can make such constraints unstable; robustness to data-derived information and contradictory prior information remains future work.

\section{Preliminaries}\label{sec:Preliminaries}

We adopt established concepts from graphical modeling \citep{handbook}, as reviewed next.

\subsection{Graphical Concepts}\label{para:Graphs}

Let $G=(V,E)$ be a directed graph with vertex set $V=[d]:=\{1,\dots,d\}$. We denote parents, children, ancestors, and descendants of a node $v \in V$ by $\pa{v}$, $\ch{v}$, $\an{v}$, and $\de{v}$, respectively, where we let $v\in\an{v}$ and $v\in\de{v}$. These notions naturally extend to sets of nodes by taking unions. When the graph is clear from context, the subscript $G$ may be omitted. A directed $v$-$w$ \emph{walk} $\pi$ in $G$ is an ordered sequence of directed edges whose consecutive edges are incident and which starts at $v$ and ends at $w$; nodes and edges may be visited multiple times.

A \emph{trek} in $G$ from $v$ to $w$ is an ordered pair of directed walks ($\pi_L$, $\pi_R$) where $\pi_L$ has sink $v$, $\pi_R$ has sink $w$, and both $\pi_L$ and $\pi_R$ have the same source $k$. Walks of length zero are allowed, so a directed walk (e.g., with $v=k$) also constitutes a trek $(\emptyset, \pi_R)$.
We denote the set of all \emph{treks} between $i$ and $j$ by $\cal{T}_{ij}$. The trek graph $T(G)$ is the undirected graph on $V$ in which two distinct nodes $i$ and $j$ are adjacent if and only if $G$ contains an $i$-$j$-trek, and $T(G)^c$ denotes its complement.

\subsection{Structural Equation Models}\label{subsec:SEMs and score-based learning}

Let $X = (X_1,\dots,X_d)$ be a $d$-dimensional random vector. 
A (nonparametric) structural equation model (SEM) postulates that $X$ solves a system of equations
\begin{equation}\label{eq:SEM}
    X_j = f_j(X, \varepsilon_j), \quad j \in [d],
\end{equation}
where each $f_j : \mathbb{R}^{d+1} \to \mathbb{R}$ is measurable and 
$\varepsilon_j$ are jointly independent exogenous noise variables.
Although $f_j$ may depend only on some entries of $X$ and the noise $\varepsilon_j$, we define all functions on $\mathbb{R}^{d+1}$ for notational simplicity. 
The collection $f = (f_1,\dots,f_d)$ induces a directed graph $G(f)$ on $[d]$ by taking the parent set of each node $j$ to be the set of indices $k\neq j$ such that $f_j$ depends nontrivially on $X_k$. In the linear case that will be the focus of our later experiments, $X = W^\top X + \varepsilon$ and
$G(f)$ is given by the support of the matrix $W \in \mathbb{R}^{d \times d}$, so $k \to j \in G(f)$ if and only if $W_{k j}\not=0$.
However, the ideas we put forth extend to nonlinear models, e.g., with matrix $[W(f)]_{k,j} := \|\partial_k f_j\|$  \citep{zheng2020learning, waxman2024dagma, liang2024kernel}.

We observe $n$ i.i.d.\ samples from the distribution of $X$, collected in a data matrix $\mathcal D\in\R^{n\times d}$. Score-based methods estimate $G(f)$ by minimizing a data-dependent score $L$, such as a penalized likelihood, over the considered set of graphs.


\subsection{Marginal Independence and Treks}\label{subsec:Marginal Independence and Treks}

For acyclic SEMs, the distribution $P^X$ satisfies the \emph{global Markov property} with respect to the true graph $G=G(f)$, that is, $d$-separation in $G$ implies conditional independence in $P^X$. For the empty conditioning set, a path (or walk) between two distinct nodes $i,j\in V$ is active if and only if it contains no collider, that is, if and only if it is a trek.
Hence, under the global Markov property,
\[
\text{absence of treks between } i \text{ and } j
\implies
X_i \indep X_j.
\]
The implication also holds for cyclic SEMs whose ancestral subsystems are uniquely solvable, as is the case for linear SEMs with $I-W$ invertible, since $X_i$ is then a function of the noise variables $(\varepsilon_k)_{k\in\an{i}}$ alone \citep{bongers:2021}.
If $P^X$ is faithful to $G$, the converse holds as well. Without faithfulness, a marginal independence may also arise from cancellation along existing treks, in which case the corresponding no-trek constraint excludes the true graph.

Treks constitute the structural channels through which marginal dependence arises. In linear SEMs, the \emph{trek rule} expresses each covariance $\mathrm{Cov}(X_i,X_j)$ as a sum over $i$-$j$-treks of products of edge weights and an error variance \citep{sullivant2010trek}, so that $\mathrm{Cov}(X_i,X_j)=0$ in the absence of such treks. For Gaussian SEMs, this is equivalent to $X_i \indep X_j$.
For certain continuous-time models, including stationary diffusions, marginal independencies in the stationary distribution also result from the absence of treks \citep{boege2025conditional,varando2020graphical}.

Under faithfulness, supplied marginal-independence pairs therefore translate into structural no-trek constraints for the true graph $G$. We represent these constraints by a subgraph $H=(V,\cal{I})$ of the trek-graph complement $T(G)^c$, refer to $H$ as NOTREKS information, and call the pairs in $\cal{I}$ NOTREKS pairs. When $V$ is clear from context, we also refer to the NOTREKS information only through $\cal{I}$. Every DAG $\Tilde{G}$ with $H \subseteq T(\Tilde{G})^c$, that is, every DAG without a trek between any pair in $\cal{I}$, is called NOTREKS-feasible.

\paragraph{Statistical Viewpoint.}
Under correct model specification, a consistent score recovers the true Markov equivalence class asymptotically without NOTREKS information. Correct NOTREKS information can, however, help in finite samples, by ruling out graphs that the score favors only because of sampling fluctuations, and under model misspecification.

\paragraph{Combinatorial Viewpoint.}\label{remark:Combinatorial Viewpoint}
Known marginal independencies shrink the set of feasible DAGs, but this does not generally make exact search tractable. Let $N(H)$ denote the number of NOTREKS-feasible DAGs and $a_m$ the number of labeled DAGs on $m$ nodes. If $S\subseteq V$ is an independent set of $H$, i.e., $H[S]$ has no edges, then any DAG on $S$, with all other nodes isolated, is NOTREKS-feasible, so $N(H)\geq a_{\abs{S}}$. Applying this to the color classes of a proper coloring of $H$ gives
\begin{equation*}
  \log_2 N(H)\geq \frac{d}{2}\left(\frac{d}{\chi(H)}-1\right),
\end{equation*}
where $\chi(H)$ denotes the chromatic number of $H$ (see \cref{subsec:NOTREKS Combinatorial Viewpoint}). This bound depends on $H$ only through $\chi(H)$, so the structure of $H$, and not only the number of supplied pairs, matters for how restrictive the information is. For $k$-Erdős--Rényi DAGs $G_d$ with complete NOTREKS information $H_d=T(G_d)^c$, the chromatic number $\chi(H_d)$ is $\Theta_{\P}(d)$.
Together with an upper bound on $N(H)$ in this special case, this yields $N(H_d)=2^{\Theta_{\P}(d)}$, compared with $2^{\Theta(d^2)}$ for unrestricted DAGs, so the feasible space remains exponential (\cref{subsec:NOTREKS Combinatorial Viewpoint}).

The chromatic number also has a structural interpretation. The source nodes of a DAG $G$ are pairwise not connected by treks \citep[Lemma 2]{markham2022transformational}, so they form a clique in $T(G)^c$ and $\abs{\operatorname{src}(G)} \leq \chi(T(G)^c)$; in fact, equality holds (\cref{cor:treks_and_sources}). Since $H\subseteq T(\Tilde{G})^c$ for every NOTREKS-feasible DAG $\Tilde{G}$, every feasible DAG has at least $\chi(H)$ source nodes. Our algorithms do not use this bound directly, but we examine the role of $\chi(H)$ empirically in \cref{sec:Experiments}.

\section{Trek Constraints}\label{sec:Power Series Trek constraint}

\subsection{Ancestral Characterization}
\label{subsec:Ancestral characterization}

Let $B\in\{0,1\}^{d\times d}$ be the (unweighted) adjacency matrix of a DAG. The $(\ell,i)$-entry of its path matrix $\sum_{k=0}^{d-1}B^k$ counts the directed paths from $\ell$ to $i$, including the path of length zero for $\ell=i$. From it, \citet{deligeorgaki2024combinatorial} construct the trek matrix
\begin{equation}\label{eq:binary_notreks}
    \Big(\sum\nolimits_{k=0}^{d-1}B^k\Big)^{\top} \Big(\sum\nolimits_{k=0}^{d-1}B^k\Big),
\end{equation}
whose $(i,j)$-entry glues together all pairs of paths with a common source and hence counts the treks between $i$ and $j$. In particular, the DAG contains an $i$-$j$-trek if and only if this entry is non-zero (see \cref{sec:additional-trek-constraint-material} for a $3$-node example). This characterization is directly applicable in causal discovery methods that perform discrete structure search, such as FLOP \citep{wienöbst2026embracingdiscretesearchreasonable}.


FLOP searches over topological orderings of the nodes and, for a given ordering, performs a parent search for every node, where feasible parents are those that precede the node in the ordering. This way acyclicity is ensured. NOTREKS-feasibility, however, is not ensured by the ordering alone: for every supplied pair $(i,j)\in\mathcal I$, the returned graph $G$, which in this section denotes the estimated graph rather than the true one, must satisfy $\operatorname{an}_G(i)\cap\operatorname{an}_G(j)=\emptyset$. We therefore slightly alter the parent search of FLOP, a grow--shrink procedure that first adds parents as long as this improves the score and then removes parents as long as this improves the score further. In this procedure, we ensure NOTREKS-feasibility by rejecting any proposed edge addition that would violate the supplied NOTREKS information. Edge deletions require no such check; see \cref{alg:grow-shrink-nt-main} for the pseudo-code. The method returns a DAG satisfying all supplied NOTREKS constraints. In the implementation an ancestry table is more efficient than repeated calculations of \cref{eq:binary_notreks}; maintaining this table and checking proposed additions adds computational overhead relative to vanilla FLOP.

\begin{figure}[t]
\centering
\begin{minipage}{\linewidth}
  \scriptsize
  \begin{algorithm2e}[H]
  \caption{\textsc{GrowShrinkNOTREKS}. The modification of FLOP is highlighted.}
  \label{alg:grow-shrink-nt-main}

    \Input{data $\mathcal D$, target $v$, preceding nodes $Z$, current DAG $G$, previous parents $P_{\mathrm{old}}$, NOTREKS pairs $\mathcal I$}

  $P_v \gets P_{\mathrm{old}} \cap Z$\;
  $\ell_v \gets \textsc{LocalScore}(v,P_v,\mathcal D)$\;

  \Repeat{$P_v=P_{\mathrm{pass}}$}{
    $P_{\mathrm{pass}} \gets P_v$\;
    \ForEach{$u \in Z \setminus P_v$}{
      $\ell_{\mathrm{new}} \gets
        \textsc{LocalScore}(v,P_v\cup\{u\},\mathcal D)$\;
      \If{\colorbox{notrekpurple!50}{$G+(u\mathbin{\to}v)$ satisfies $\mathcal I$} and $\ell_{\mathrm{new}}\leq\ell_v$}{
        $P_v \gets P_v\cup\{u\}$; $\ell_v \gets \ell_{\mathrm{new}}$\;
      }
    }
  }

  \Repeat{$P_v=P_{\mathrm{pass}}$}{
    $P_{\mathrm{pass}} \gets P_v$\;
    \ForEach{$u \in P_v$}{
      $\ell_{\mathrm{new}} \gets
        \textsc{LocalScore}(v,P_v\setminus\{u\},\mathcal D)$\;
      \If{$\ell_{\mathrm{new}}\leq\ell_v$}{
        $P_v \gets P_v\setminus\{u\}$; $\ell_v \gets \ell_{\mathrm{new}}$\;
      }
    }
  }

  \ReturnKw{$(P_v,\ell_v)$}\;
\end{algorithm2e}
\end{minipage}
\end{figure}

A well-known difficulty in combining decomposable scores with non-decomposable structural constraints is that parent sets can no longer be optimized separately \citep{chen2016ancestral,li2018sideconstraints}. The NOTREKS constraint has this property because a trek depends on the parent sets of all ancestors. The constrained grow--shrink procedure is therefore a feasibility-preserving heuristic without a global optimality guarantee: early parent choices may preclude better feasible graphs.

\subsection{Power Series Trek Constraint}\label{subsec:Power Series Trek Family}

Gradient-based causal discovery requires a differentiable version of the trek matrix in \cref{eq:binary_notreks}. For $W\in\R^{d\times d}$, let $G_W$ be the weighted directed graph with an edge $i\to j$ whenever $W_{ij}\neq0$. We seek a smooth function of $W$ that represents the requirement that specified pairs of vertices have no trek in $G_W$.
Non-negativity of $W$ is crucial for trek-characterization, as for signed matrices cancellation
can otherwise destroy the characterization. For instance, with
$W_c=c\left[\begin{smallmatrix}
0&-1\\1&0
\end{smallmatrix}\right]$
one has
\begin{equation*}
  \begin{aligned}
    \exp(W_c)&=\cos(c)I_2 + \frac{\sin(c)}{c}W_c.
  \end{aligned}
\end{equation*}
At $c=\pi$, $\exp(W_\pi)=-I_2$ and
$\exp(W_\pi)^\top\exp(W_\pi)=I_2$, despite the directed cycle connecting the
two vertices. Instead of restricting $W$ to have non-negative entries, we apply an \emph{admissible adjacency mapping} $\phi$ entrywise, which makes every non-zero edge contribution positive and prevents such cancellation while keeping the zero-pattern unchanged.

\begin{definition}[Admissible adjacency mapping]
\label[definition]{def:Admissible adjacency mapping}
    A function $\phi\in C^1(\R)$ is an \emph{admissible adjacency mapping} if
    $\phi(0)=0$ and $\phi(x)>0$ for every $x\neq0$.
\end{definition}
The standard choice is $\phi(x)=x^2$, for which
$\phi(W)=W\circ W$ \citep{zheng2018dags, bello2022dagma}.  Other choices allow the geometry of the smooth
constraint to be adjusted without changing its zero pattern.

Only the zero pattern of the path matrix matters for the trek characterization. We may thus generalize the path matrix $\sum_{k=0}^{d-1}B^k$ in two ways. First, we replace the binary matrix $B$ by the non-negative matrix $\phi(W)$. Second, we replace the finite sum by a matrix power series
\[
    f_a(\phi(W))=\sum_{k\geq0}a_k\,\phi(W)^k,
\]
obtained by evaluating the scalar series $f_a(x)=\sum_{k\geq0}a_kx^k$ with non-negative coefficients $a=(a_k)_{k\in\mathbb N}$ at the matrix $\phi(W)$. In \cref{eq:binary_notreks}, we have $a_k=1$ for $k\leq d-1$ and $a_k=0$ otherwise. Since every directed walk witnessing reachability contains a directed path of length at most $d-1$, $f_a(\phi(W))$ has the same zero pattern as the path matrix of $G_W$ whenever the series converges and $a_k>0$ for $k\leq d-1$. Infinite series are useful because they may admit closed forms that are convenient to compute and differentiate, such as the matrix exponential in NOTEARS or the matrix inverse underlying DAGMA. \Cref{tab:trek constraints examples} lists representative choices, following the power-series trace families of \citet[Table 2]{nazaret2023stable}. Combining such a series with an admissible adjacency mapping yields the \emph{Power Series Trek} (PSTrek) constraint.
\begin{table}[t]
    \centering
    \scriptsize
    \floatconts{tab:trek constraints examples}{\caption{Representative members of the PSTrek family.}}{%
\begin{tabular}{lccc}
        \toprule
        Name & $a_k$ & $f_a(x)$ & $R_a$ \\
        \midrule
        $h_{\mathrm{binom}}$ & $\binom{d}{k}$ & $(1+x)^d$ & $\infty$ \\
        $h_{\exp}$ & $1/k!$ & $\exp(x)$ & $\infty$ \\
        $h_{\log}$ & $1/k\;(a_0=1)$ & $1-\log(1-x)$ & $1$ \\
        $h_{\mathrm{inv},s}$ & $s^{-(k+1)}$ & $(s-x)^{-1}$ & $s$ \\
        \bottomrule
    \end{tabular}}
\end{table}

\begin{definition}[Power Series Trek Constraint]\label[definition]{def:PSTrek family}
    Let $f_a(x)=\sum_{k\geq0}a_kx^k$ be a power series with non-negative coefficients $a=(a_k)_{k\in\mathbb N}$ and convergence radius $R_a$, and let $\phi$ be an admissible adjacency mapping. On the domain $\mathcal D_{a,\phi}:=\set{W\in\R^{d\times d}\mid\rho(\phi(W))<R_a}$, the \emph{Power Series Trek} constraint matrix is
    \begin{equation}\label{eq:def of PSTrek family}
        h_{a,\phi}(W)
        :=f_a(\phi(W))^\top f_a(\phi(W))
        \in\R_{\geq0}^{d\times d}.
    \end{equation}
\end{definition}
By absolute convergence, the entries of $h_{a,\phi}(W)$ admit the expansion
\begin{equation}\label{eq:PSTrek explicit form}
    h_{a,\phi}(W)_{ij}
    =\sum_{\tau=(\pi_L,\pi_R)\in\cal T_{ij}}M_\tau(W),
    \qquad
    M_\tau(W)
    :=a_{\abs{\pi_L}}a_{\abs{\pi_R}}
      \prod_{e\in\pi_L\sqcup\pi_R}\phi(W_e),
\end{equation}
see \cref{subsec:PSTrek well-defined}. The product is over edge occurrences, so that an edge traversed repeatedly, within one walk or across both walks, contributes its factor with the total multiplicity.

The following result extends \citet[Theorem~2.2.3]{deligeorgaki2024combinatorial} from binary-valued to real-valued matrices.

\begin{theorem}[PSTrek characterization of treks]
\label[theorem]{th:PSTrek characterization of treks}
    Let $\phi$ be an admissible adjacency mapping, and let the coefficients
    $a=(a_k)_{k\in\mathbb N}\in\R_{\geq0}^{\mathbb N}$ satisfy
    $a_k>0$ for $k=0,\ldots,d-1$. Then the following hold on $\mathcal D_{a,\phi}$:
    \begin{enumerate}
        \item[(i)] $h_{a,\phi}$ is continuously differentiable; and
        \item[(ii)] for distinct $i,j\in[d]$, $h_{a,\phi}(W)_{ij}=0
            \quad\Longleftrightarrow\quad
            \text{there is no $i$-$j$-trek in $G_W$}$.
    \end{enumerate}
\end{theorem}
The proof is given in \cref{proof:th:PSTrek characterization of treks_revised}. The theorem does not require $G_W$ to be acyclic and also covers cyclic graphs on $\mathcal D_{a,\phi}$, for which the series includes walks with repeated edge occurrences.

To encode a set $\mathcal I$ of supplied NOTREKS pairs, we sum the corresponding entries and define the PSTrek penalty
\begin{equation}
\label{eq:set trek regularizer}
    h_{a,\phi}(W)_{\mathcal I}
    := \frac{2}{d-1}
        \sum_{(i,j)\in\mathcal I}h_{a,\phi}(W)_{ij}.
\end{equation}
The factor $2/(d-1)$ is a fixed rescaling that does not change which matrices satisfy the constraint. It only sets the scale on which the weight $\lambda_{\mathrm{NT}}$ is chosen. With a fixed weight $\lambda_{\mathrm{NT}}$, supplying more NOTREKS pairs increases the total penalty.

PSTrek constraints are generally non-convex (\cref{app_subsec:Non-Convexity}). Under a mild additional condition on $\phi$, positive-valued stationary points are nevertheless ruled out.

\begin{theorem}[Invexity of PSTrek constraints]
\label[theorem]{th:PSTrek constraints are invex}
    Let the assumptions of
    \cref{th:PSTrek characterization of treks} hold and suppose that $\phi$
    is \emph{radially admissible}, i.e., $x\phi'(x)>0$ for every $x\neq0$. Then, for distinct $i,j\in[d]$, the function
    $W\mapsto h_{a,\phi}(W)_{ij}$ is invex on
    $\mathcal D_{a,\phi}$, that is, each of its stationary points is a global minimizer. The same holds for any non-negative linear combination of such functions, in particular for $h_{a,\phi}(W)_{\mathcal I}$, and for non-negative linear combinations with the PSTrace acyclicity constraints of \citet{nazaret2023stable}, defined in \cref{def:Power Series Trace constraint}.
\end{theorem}
According to the theorem, which is proven in \cref{proof:PSTrek constraints are invex}, the PSTrek constraint has no stationary point with positive constraint value.  This property concerns the penalty alone. Adding a data-dependent score may destroy invexity, and the theorem does not guarantee that a local optimization method finds a globally optimal feasible graph.

\begin{proposition}[Convergence to the feasible region]\label[proposition]{cor:convergence_to_feasible_region}
    Let $a_k>0$ for $k=0,\ldots,d$, let $\phi$ be an admissible adjacency mapping, and let $\mathcal X\subseteq\mathcal D_{a,\phi}$ be compact with $0\in\mathcal X$. Let $L:\mathcal X\to\R$ be a continuous score, let $\mu^{(t)}\to0^+$, and let $W^{(t)}$ be a global minimizer on $\mathcal X$ of
    \begin{equation}\label{eq:DAGMA+NOTREKS objective}
        \begin{aligned}
        &\mu^{(t)}L(W)+g_{a,\phi}(W) +h_{a, \phi}(W)_{\mathcal I},
        \end{aligned}
    \end{equation}
    where $g_{a,\phi}$ denotes the PSTrace constraint (\cref{def:Power Series Trace constraint}). Then the sequence $(W^{(t)})$ has accumulation points, and for each accumulation point $W^\star$, the graph $G_{W^\star}$ is a NOTREKS-feasible DAG.
\end{proposition}
For simplicity, both penalties use the same $a$ and $\phi$, but the result also holds for separate choices. The proposition concerns idealized global minimizers. It does not establish convergence of practical schemes that minimize \cref{eq:DAGMA+NOTREKS objective} only approximately.

\begin{figure}[t]
\centering
\begin{minipage}{\linewidth}
  \small
  \begin{algorithm2e}[H]
  \caption{DAGMA + NOTREKS. The term added to the DAGMA objective is highlighted.}
  \label{alg:dagma-notreks}
  \Input{score $L(W;\mathcal D)$, NOTREKS pairs $\mathcal I\neq\emptyset$, stages $(\mu_t,s_t)_{t=1}^T$, weights $\lambda_1$, $\lambda_{\mathrm{NT}}$}
  Initialize $W_0\gets 0$\;
  \For{$t=1,\ldots,T$}{
    $W_t\gets$ approximate minimizer, with warm start $W_{t-1}$, of
    $\mu_t\left(L(W;\mathcal D)+\lambda_1\lVert W\rVert_1\right)+g_{\log\det,s_t,x^2}(W)+\colorbox{notrekpurple!50}{$\lambda_{\mathrm{NT}}h_{\mathrm{inv},s_t,x^2}(W)_{\mathcal I}$}$
    subject to $\rho(W\circ W)<s_t$\;
  }
  $G\gets\textsc{FeasiblePostSelection}(W_T,L,\mathcal D,\mathcal I)$\;
  \ReturnKw{$G$}\;
  \end{algorithm2e}
  \vspace{0.25ex}
  \begin{algorithm2e}[H]
  \caption{\textsc{FeasiblePostSelection}}
  \label{alg:feasible-postselection}
  \Input{weighted estimate $W_T$, score $L$, data $\mathcal D$, NOTREKS pairs $\mathcal I$}
  $\mathcal T\gets\{0\}\cup\{\abs{(W_T)_{ij}}:i\neq j,\ (W_T)_{ij}\neq0\}$\;
  \tcp{The largest threshold yields the empty, trivially feasible graph}
  For each $\tau\in\mathcal T$, form the support $E_\tau=\{i\to j:\abs{(W_T)_{ij}}>\tau\}$\;
  Retain the acyclic, NOTREKS-feasible supports among $\{E_\tau:\tau\in\mathcal T\}$ and refit the coefficients on each\;
  $G\gets$ the feasible refitted candidate with lowest ordinary BIC\;
  \ReturnKw{$G$}\;
  \end{algorithm2e}
\end{minipage}
\end{figure}

To obtain a practical method, we extend DAGMA \citep{bello2022dagma} by incorporating the PSTrek penalty into its continuous optimization objective (\cref{alg:dagma-notreks}). The extension retains DAGMA's squared adjacency map, sparsity regularization, log-determinant acyclicity constraint, continuation scheme, and warm starts. As penalty, we use $h_{\mathrm{inv},s}$ from \cref{tab:trek constraints examples} with $\phi(x)=x^2$, so that $f_a(\phi(W))=(sI-W\circ W)^{-1}$, aggregated over the NOTREKS pairs as in \cref{eq:set trek regularizer}. This choice fits DAGMA well. Its domain $\rho(W\circ W)<s$ coincides with DAGMA's feasible region, and the inverse $(sI-W\circ W)^{-1}$ already appears in the gradient of DAGMA's log-determinant constraint, so it can be reused. Evaluating the penalty and its gradient thus requires $O(d^3)$ operations, the same order as the DAGMA penalty.

In addition, we modify DAGMA's final graph selection. Thresholding the dense weighted output of the optimizer need not yield a NOTREKS-feasible graph. We therefore apply the feasible post-selection procedure in \cref{alg:feasible-postselection}, which returns an acyclic and NOTREKS-feasible graph.

\begin{remark}
The PSTrek penalty is not specific to DAGMA. It can be added to any continuous score-based method for a model class in which $X_i\indep X_j$ implies the absence of an $i$-$j$-trek.
\end{remark}

\begin{remark}[Trek-cycle constraint]
We additionally investigated a spectral trek-cycle construction. Its computational cost exceeded that of PSTrek, and it provided little benefit in our DAGMA experiments, so we present its construction, properties, and comparison with PSTrek in Appendix~\ref{subsec:Trek-Cycle Coupling constraint}.
\end{remark}

\section{Synthetic Experiments}\label{sec:Experiments}
\subsection{Experimental Setup}\label{subsec:experimental-setup}
\textit{Data generation.} We focus our experiments on linear-Gaussian SEMs with heteroskedastic noise. Causal graphs are obtained by sampling an undirected Erd\H{o}s--R\'enyi (ER) or Watts--Strogatz (WS) graph with expected degree $k\in\{2,4,8\}$ and orienting its edges according to a uniformly random ordering of the nodes. ER graphs include each edge independently with probability $p=\frac{k}{d-1}$; WS graphs \citep{watts1998collective} start from a ring on which each node is joined to its $k/2$ nearest neighbors on either side, and each edge is then rewired to a uniformly chosen node with probability $0.5$, avoiding self-loops and multiple edges.
Data are generated as $X_j=\sum_{i\in \operatorname{pa}(j)}w_{ij}X_i+\varepsilon_j$, with $w_{ij}\sim \operatorname{Unif}([-2,-0.5]\cup[0.5,2])$, $\log(\sigma_j^2)\sim \operatorname{Unif}(-\log 2,\log 2)$, and $\varepsilon_j\sim \mathcal N(0,\sigma_j^2)$, and each dataset is standardized. For each graph family and degree, we draw $20$ graphs and use the same graph and data-generating draws across the sample sizes $n\in\{100,500,2000\}$.

\medskip\noindent\textit{NOTREKS information.} We measure the amount of NOTREKS information by the fraction of trek-free pairs of $G$ that are supplied, $q := \abs{\mathcal I}/\abs{E(T(G)^c)}$, and draw $\mathcal I$ at random from the edges of $T(G)^c$. We compare no information ($q=0$), one quarter of the available pairs ($q=\frac14$, with five independent draws of $\mathcal I$ per graph), and all available pairs ($q=1$). Since dense graphs have fewer trek-free pairs, a fixed $q$ yields less NOTREKS information for dense than for sparse graphs.

\medskip\noindent\textit{Methods.} We compare FLOP+NOTREKS and DAGMA+NOTREKS with the corresponding vanilla methods. FLOP uses a decomposable linear-Gaussian BIC and $20$ random restarts of the causal order, and is evaluated for $d\in\{20,50,100\}$. DAGMA uses two optimization initializations and is evaluated in $d\in\{20,50\}$, because the continuous optimization is substantially more runtime-intensive at $d=100$.
The original DAGMA \citep{bello2022dagma} uses an $\ell_2$ score for continuous data; in all of our DAGMA experiments, both vanilla DAGMA and DAGMA+NOTREKS instead use the profiled unequal-variance Gaussian criterion, together with an $\ell_1$ penalty and log-determinant acyclicity constraint. Relative to the original DAGMA, we therefore change the data score and replace fixed thresholding of the weighted output by the feasible support-selection procedure in Algorithm~\ref{alg:feasible-postselection}; we retain DAGMA's squared adjacency map, log-determinant constraint, continuation schedule, warm starts, and optimizer.
As diagnostic baselines, we include $\mathrm{Var}$-SortnRegress and $R^2$-SortnRegress \citep{reisach2021beware,reisach2023scale}, which check whether ordering variables by marginal variance or by predictability from all other variables already explains performance. The exact objectives and production settings are given in \cref{sec:appendix_experimental_details}.

\medskip\noindent\textit{Metrics.} Since only the Markov equivalence class is identifiable, we map each estimated graph to its completed partially directed acyclic graph (CPDAG) and compare it with the true CPDAG. We report the structural Hamming distance (CPDAG SHD), which counts missing and extra adjacencies as well as incorrect edge marks, and, as a measure of causal-effect relevance, the ancestor-based adjustment identification distance (Ancestor-AID) \citep{henckel2024adjustment}. Ancestor-AID counts the ordered pairs $(i,j)$ for which the estimated CPDAG leads to a different conclusion about identifying the effect of $X_i$ on $X_j$ by ancestor adjustment than the true CPDAG. Smaller values are better for both; precise definitions are given in \cref{sec:appendix_experimental_details}.
\begin{figure}[t]
    \centering
    \includegraphics[width=1\linewidth]{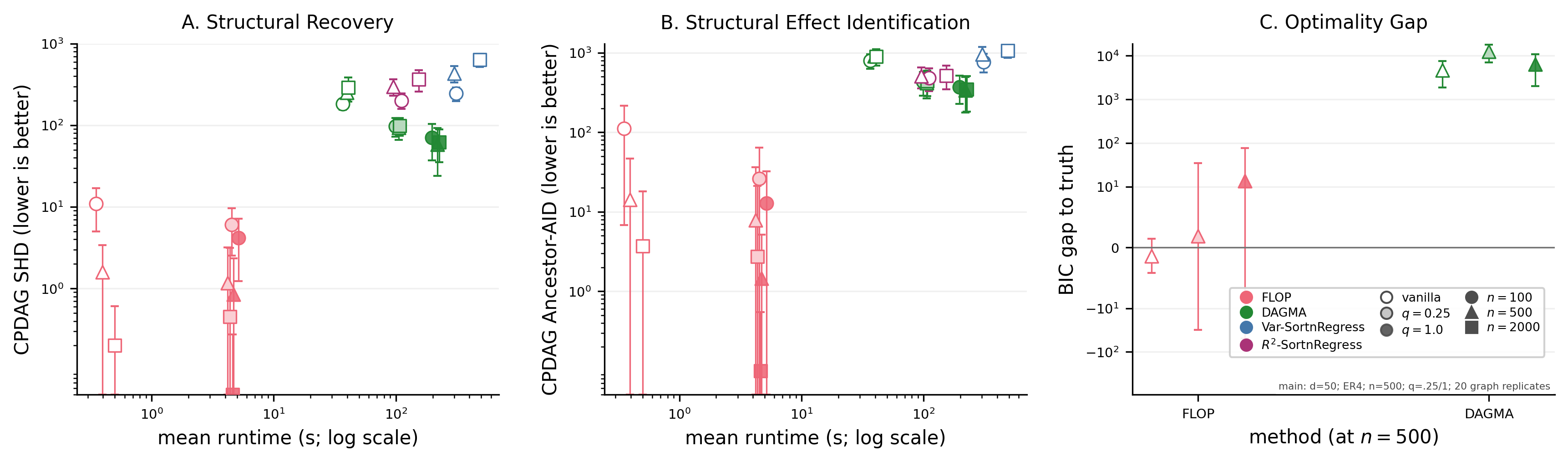}
    \caption{Results for ER$4$ graphs in $d=50$. Panels A and B show group means for CPDAG SHD and Ancestor-AID, respectively; panel C shows the mean BIC gap to the refitted ground-truth graph on a symmetric-log axis. Whiskers indicate $\pm 1$ standard deviation.}
    \label{fig:main-statistical-recovery}
\end{figure}

\begin{figure}[t]
  \centering
  \begin{minipage}{.6\textwidth}
    \centering
    \includegraphics[
      width=\linewidth,
      height=4.35cm,
      keepaspectratio
    ]{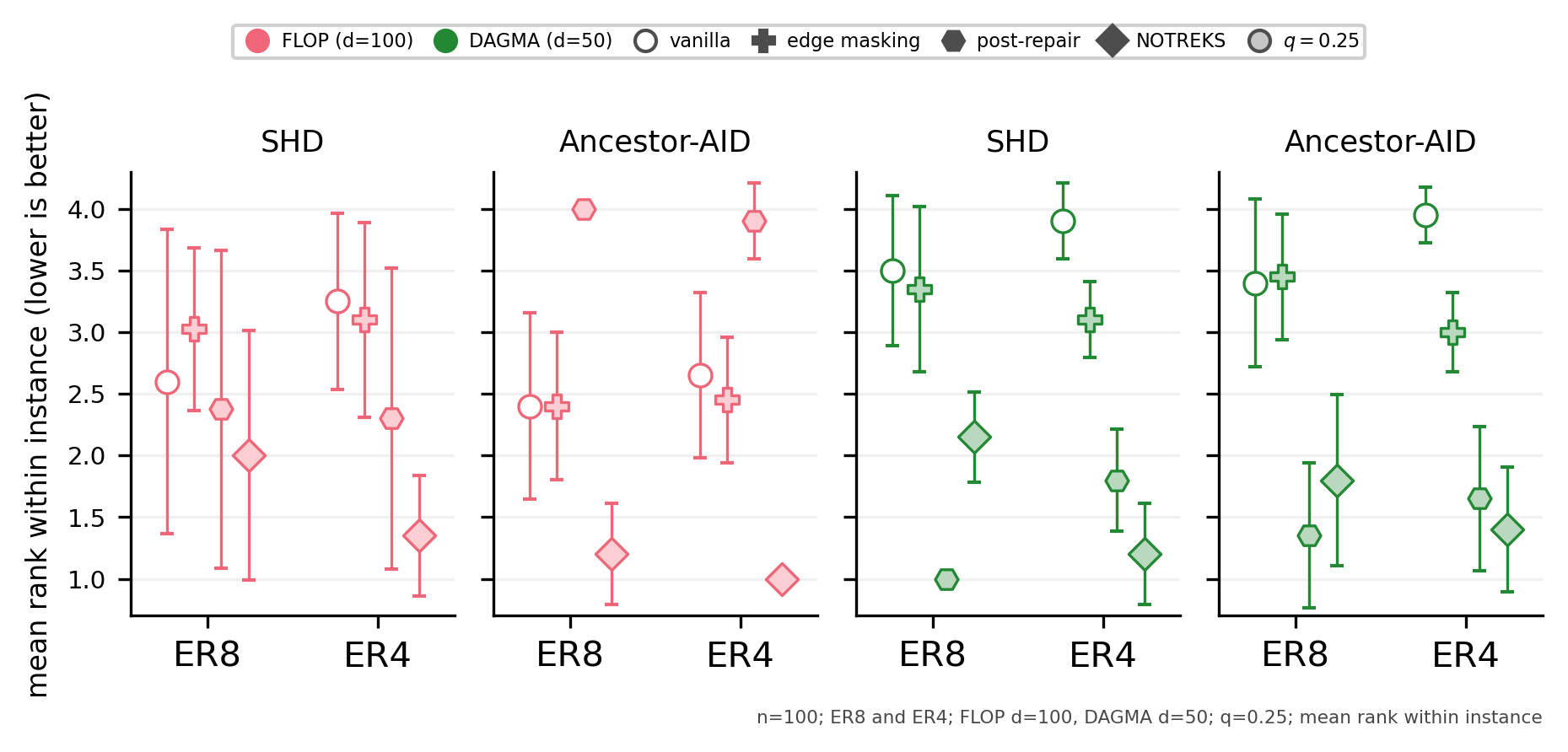}
  \end{minipage}%
  \begin{minipage}{.4\textwidth}
    \centering
    \includegraphics[
      width=\linewidth,
      height=4.35cm,
      keepaspectratio
    ]{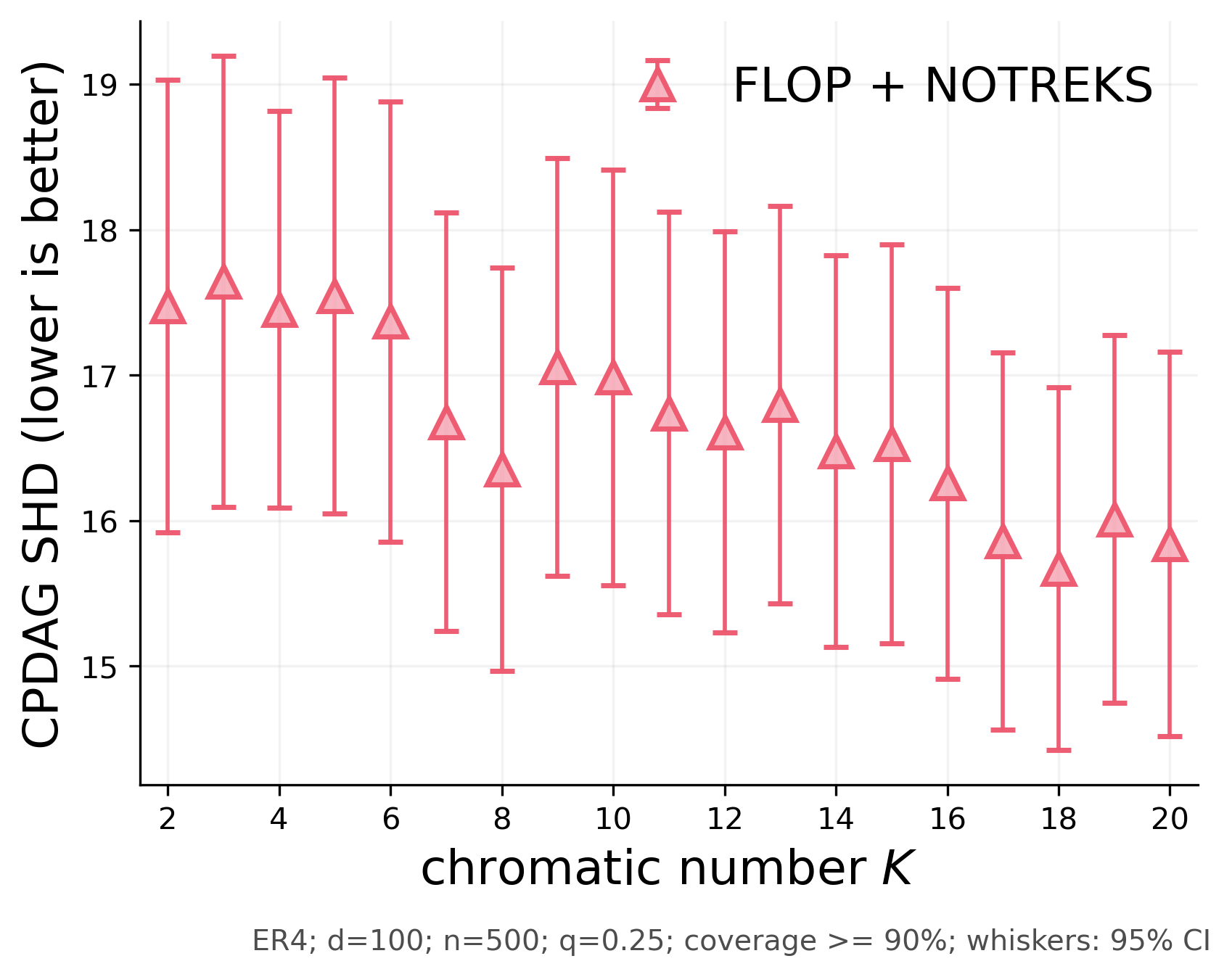}
  \end{minipage}
  \caption{NOTREKS information ablation study. Left: Benchmark against simple baselines of integrating the NOTREKS information at $n=100$. Markers show the mean within-instance rank based on CPDAG SHD; whiskers indicate $\pm 1$ standard deviation across paired benchmark instances. Lower ranks are better. The corresponding paired CPDAG-SHD and Ancestor-AID tests at $n=500$ are reported in \cref{tab:paired_ablation_tests}. Right: FLOP structural recovery by chromatic number of NOTREKS information; markers show the mean CPDAG SHD over graph-level means, and whiskers show two-sided $95\%$ Student-$t$ confidence intervals for that mean across the represented graphs at each $K$. The DAGMA trend is reported separately in \cref{fig:prior-structure-chromatic-trends}. Only target values represented for at least $90\%$ of generated graphs are included.}
  \label{fig:ablation_rank_&_prior-structure-chromatic-number}
\end{figure}

\medskip\noindent\textit{Code availability.} The implementations and experiment workflows used in this paper are available at \href{https://anonymous.4open.science/r/notreks-E13C/}{anonymous.4open.science/notreks}.

\subsection{Results}\label{subsec:experimental-results}
\textit{Statistical recovery and optimization.} Correct NOTREKS information improves both methods relative to their vanilla counterparts and never leads to significantly worse recovery (\cref{fig:main-statistical-recovery}; paired tests in \cref{tab:additional_synthetic_notreks_vanilla}). For DAGMA, the improvement is significant in every setting and for both metrics, and it grows with the sample size; while vanilla DAGMA barely beats the diagnostic baselines $\mathrm{Var}$-SortnRegress and $R^2$-SortnRegress, DAGMA+NOTREKS clearly outperforms them. FLOP already performs very well without prior information, yet FLOP+NOTREKS still achieves significant gains in all sparse settings at $n=100$, with the largest gains in Ancestor-AID and in higher dimensions (e.g., a mean reduction of $286$ in Ancestor-AID for $k=4$ and $d=100$). As expected, these gains diminish as $n$ grows and the data alone become more informative. The NOTREKS variants show a larger BIC gap to the ground truth (\cref{fig:main-statistical-recovery}C). Since the true graph is NOTREKS-feasible, this indicates that the constrained searches do not always find the best feasible graph, so better optimization could increase the benefit further.

\medskip\noindent\textit{Comparison with simpler prior integrations.} We compare the NOTREKS variants with two simpler uses of the same information: edge masking, which only forbids direct edges between pairs in $\mathcal I$ (treks of length one), and post-selection, which applies the feasible post-selection of \cref{alg:feasible-postselection} to the output of the vanilla method. \Cref{fig:ablation_rank_&_prior-structure-chromatic-number} (left) shows ranks at $n=100$, and \cref{tab:paired_ablation_tests} reports paired tests at $n=500$. FLOP+NOTREKS is never significantly worse than any of these baselines. It is on par with post-selection in CPDAG SHD but substantially better in Ancestor-AID in every setting, suggesting that removing treks after the search corrects the skeleton but not the ancestral relations that matter for adjustment. DAGMA+NOTREKS significantly outperforms vanilla DAGMA and edge masking in all settings, and post-selection in CPDAG SHD for the sparser graphs ($k=2,4$); only for dense graphs ($k=8$), where few NOTREKS pairs are available, DAGMA+NOTREKS loses to post-selection, which achieves significantly lower CPDAG SHD and Ancestor-AID. Overall, enforcing the full no-trek restriction during the search pays off, most clearly for effect identification.

\medskip\noindent\textit{Prior-structure effects.} The lower bound on the number of feasible DAGs in \cref{subsec:Marginal Independence and Treks} depends on the NOTREKS information only through its chromatic number $\chi(H)$, so the structure of $H$, and not only its size, determines how restrictive it is. To examine whether this also matters for recovery, we generate NOTREKS information for ER4 graphs ($n=500$) with a fixed number of pairs, one quarter of the expected number of trek-free pairs, and a prescribed chromatic number $K=\chi(H)$ (construction in \cref{subsec:Structure of NOTREKS Information}). For FLOP ($d=100$), CPDAG SHD decreases significantly with $K$ (slope $-0.10$ per unit of $K$, 95\% CI $[-0.13,-0.07]$; \cref{fig:ablation_rank_&_prior-structure-chromatic-number}, right), in line with the combinatorial analysis. For DAGMA ($d=50$), the trend is reversed (slope $0.47$, 95\% CI $[0.38,0.56]$; \cref{fig:prior-structure-chromatic-trends}), suggesting that constraints with higher chromatic number make the continuous optimization harder. Thus, how NOTREKS information is structured, and not only how much of it is supplied, affects recovery, in a method-dependent way.

\medskip\noindent\textit{Further experiments.} The benefits of NOTREKS persist under model misspecification (\cref{tab:additional_synthetic_notreks_vanilla} and \cref{subsec:Additional Synthetic results}). With nonlinear mechanisms or Laplace noise, both NOTREKS variants improve significantly on their vanilla counterparts in most settings and are never significantly worse. We also evaluate all methods on the semi-synthetic manufacturing benchmarks CausalMan \citep{tagliapietra2025causalman} and CausalAssembly \citep{gobler2024texttt} (\cref{app:subsec:In-Silico Benchmarks: Assembly Lines}). Here, the data-generating mechanisms depart strongly from the assumed models, structural recovery remains difficult for all methods, and the available NOTREKS information does not yield consistent improvements.

\section{Conclusion and Outlook}
We introduced NOTREKS, a way to incorporate prior knowledge of marginal independence into score-based causal discovery by translating it, under faithfulness, into the absence of treks. For gradient-based methods, the PSTrek penalty encodes these constraints as a differentiable function of the weight matrix; for order-based search, a trek-aware parent search restricts FLOP to feasible graphs. Our combinatorial analysis shows that NOTREKS information can shrink the search space considerably, and that its structure, through the chromatic number of $H$, and not only its size, determines how restrictive it is.

In our synthetic experiments, correct NOTREKS information improved both DAGMA and FLOP and never led to significantly worse recovery. The gains were large for DAGMA throughout and, for FLOP, most pronounced at small sample sizes and for adjustment-based effect identification. Enforcing the constraints during the search generally outperformed simpler uses of the same information, such as edge masking or post-selection, in particular for effect identification, and the benefits persisted under nonlinear mechanisms and non-Gaussian noise.

Several directions remain open. Our experiments used correct, oracle NOTREKS information, whereas in practice prior knowledge may be incomplete or partly wrong. Since a false hard constraint excludes the true graph while a soft penalty can trade violations against data fit, a study with controlled false constraints would clarify when each form is preferable. On semi-synthetic manufacturing benchmarks with strongly misspecified models, NOTREKS did not yield consistent improvements, which motivates combining it with more flexible score functions. Finally, the larger BIC gap of the constrained searches suggests that better optimization could increase the benefit further; for FLOP, relaxing the constraint to ancestry within a bounded distance and reducing the dependence on the node processing order are natural next steps. When the prior knowledge has a simple form, such as known source nodes, direct edge restrictions may already suffice.

\acks{The authors acknowledge support from the European Research Council (ERC) under the European Union’s Horizon 2020 research and innovation programme (grant agreement No 883818).
Fabian Bleile acknowledges institutional support from the TUM Georg Nemetschek
Institute for Artificial Intelligence for the Built World and the Munich Center for Machine Learning.
Sarah Lumpp was further supported by the Munich Data Science Institute as well as the DAAD programme Konrad Zuse Schools of Excellence in Artificial Intelligence, sponsored by the Federal Ministry of Research, Technology and Space. The authors gratefully acknowledge the computational and data resources provided by the Leibniz Supercomputing Centre (\href{www.lrz.de}{www.lrz.de}).}

\bibliography{literature}

@article {bongers:2021,
    AUTHOR = {Bongers, Stephan and Forr\'e, Patrick and Peters, Jonas and Mooij, Joris M.},
     TITLE = {Foundations of structural causal models with cycles and latent
              variables},
   JOURNAL = {The Annals of Statistics},
  FJOURNAL = {The Annals of Statistics},
    VOLUME = {49},
      YEAR = {2021},
    NUMBER = {5},
     PAGES = {2885--2915},
      optISSN = {0090-5364,2168-8966},
   optMRCLASS = {62A09 (68T30 68T37)},
  optMRNUMBER = {4338897},
optMRREVIEWER = {Mario\ Alviano},
       optDOI = {10.1214/21-aos2064},
       optURL = {https://doi.org/10.1214/21-aos2064},
}

@article {shi:2022,
    AUTHOR = {Shi, Hongjian and Drton, Mathias and Han, Fang},
     TITLE = {On the power of {C}hatterjee's rank correlation},
   JOURNAL = {Biometrika},
  FJOURNAL = {Biometrika},
    VOLUME = {109},
      YEAR = {2022},
    NUMBER = {2},
     PAGES = {317--333},
      optISSN = {0006-3444,1464-3510},
   optMRCLASS = {62C20 (62G10)},
  optMRNUMBER = {4430960},
       optDOI = {10.1093/biomet/asab028},
       optURL = {https://doi.org/10.1093/biomet/asab028},
}

@book {handbook,
     TITLE = {Handbook of Graphical Models},
    optSERIES = {Chapman \& Hall/CRC Handbooks of Modern Statistical Methods},
    EDITOR = {Maathuis, Marloes and Drton, Mathias and Lauritzen, Steffen and Wainwright, Martin},
 PUBLISHER = {CRC Press, Boca Raton, FL},
      YEAR = {2019},
     PAGES = {xviii+536},
      optISBN = {978-1-4987-8862-5},
   optMRCLASS = {62-00 (62-06 62H99)},
  optMRNUMBER = {3889064},
}

@incollection {spirtes:handbook,
    AUTHOR = {Spirtes, Peter and Zhang, Kun},
     TITLE = {Search for causal models},
 BOOKTITLE = {Handbook of Graphical Models},
    SERIES = {Chapman \& Hall/CRC Handbooks of Modern Statistical Methods},
     PAGES = {439--469},
 PUBLISHER = {CRC Press, Boca Raton, FL},
      YEAR = {2019},
      optISBN = {978-1-4987-8862-5},
   optMRCLASS = {62H99 (62A01 62J02 62J05)},
  optMRNUMBER = {3888007},
}

@article{Drton:Maathuis:2017,
   author = "Drton, Mathias and Maathuis, Marloes H.",
   title = "Structure Learning in Graphical Modeling", 
   journal= "Annual Review of Statistics and Its Application",
   year = "2017",
   volume = "4",
   optnumber = "Volume 4, 2017",
   pages = "365--393",
   optdoi = "https://doi.org/10.1146/annurev-statistics-060116-053803",
   opturl = "https://www.annualreviews.org/content/journals/10.1146/annurev-statistics-060116-053803",
   optpublisher = "Annual Reviews",
   optissn = "2326-831X",
   opttype = "Journal Article",
  }

@inproceedings{varando2020graphical,
  title={Graphical continuous {L}yapunov models},
  author={Varando, Gherardo and Hansen, Niels Richard},
  booktitle={Conference on Uncertainty in Artificial Intelligence},
  pages={989--998},
  year={2020},
  organization={PMLR}
}

@article{zheng2018dags,
  title={{DAG}s with {NO TEARS}: {C}ontinuous optimization for structure learning},
  author={Zheng, Xun and Aragam, Bryon and Ravikumar, Pradeep K. and Xing, Eric P.},
  journal={Advances in Neural Information Processing Systems},
  volume={31},
  year={2018}
}

@article{csiba2017global,
  title={Global Convergence of Arbitrary-Block Gradient Methods for Generalized {P}olyak-{{\L}}ojasiewicz Functions},
  author={Csiba, Dominik and Richt{\'a}rik, Peter},
  journal={arXiv preprint arXiv:1709.03014},
  year={2017}
}

@inproceedings{sinha2021using,
  title={Using a General Prior Knowledge Graph to Improve Data-Driven Causal Network Learning.},
  author={Sinha, Meghamala and Ramsey, Stephen A.},
  booktitle={AAAI Spring Symposium: Combining Machine Learning with Knowledge Engineering},
  year={2021}
}

@inproceedings{absar2024time,
  title={Time Series Causal Discovery Using a Hybrid Method},
  author={Absar, Saima and Zhang, Lu},
  booktitle={2024 IEEE International Conference on Big Data (BigData)},
  pages={8208--8210},
  year={2024},
  organization={IEEE}
}

@incollection{pearl1994can,
  title={When can association graphs admit a causal interpretation?},
  author={Pearl, Judea and Wermuth, Nanny},
  booktitle={Selecting Models from Data: Artificial Intelligence and Statistics IV},
  pages={205--214},
  year={1994},
  publisher={Springer}
}

@inproceedings{textor2015learning,
  title={Learning from pairwise marginal independencies},
  author={Textor, Johannes and Idelberger, Alexander and Li{\'s}kiewicz, Maciej},
  booktitle={Conference on Uncertainty in Artificial Intelligence},
  year={2015},
  publisher={AUAI Press}
}

@inproceedings{markham2022transformational,
  title={A transformational characterization of unconditionally equivalent {B}ayesian networks},
  author={Markham, Alex and Deligeorgaki, Danai and Misra, Pratik and Solus, Liam},
  booktitle={International Conference on Probabilistic Graphical Models},
  pages={109--120},
  year={2022},
  organization={PMLR}
}

@article{deligeorgaki2024combinatorial,
  title={Combinatorial and algebraic perspectives on the marginal independence structure of {B}ayesian networks},
  author={Deligeorgaki, Danai and Markham, Alex and Misra, Pratik and Solus, Liam},
  journal={Algebraic Statistics},
  volume={14},
  number={2},
  pages={233--286},
  year={2024},
  publisher={Mathematical Sciences Publishers}
}

@inproceedings{wienobst2020recovering,
  title={Recovering causal structures from low-order conditional independencies},
  author={Wien{\"o}bst, Marcel and Li{\'s}kiewicz, Maciej},
  booktitle={Proceedings of the AAAI Conference on Artificial Intelligence},
  volume={34},
  pages={10302--10309},
  year={2020}
}

@article{wille2006low,
  title={Low-order conditional independence graphs for inferring genetic networks},
  author={Wille, Anja and B{\"u}hlmann, Peter},
  journal={Statistical Applications in Genetics and Molecular Biology},
  volume={5},
  number={1},
  pages={1},
  year={2006},
  publisher={De Gruyter}
}

@article{10.1093/bioinformatics/btae411,
    author = {Zhu, Yuehua and Benos, Panayiotis V. and Chikina, Maria},
    title = {A hybrid constrained continuous optimization approach for optimal causal discovery from biological data},
    journal = {Bioinformatics},
    volume = {40},
    number = {Supplement\_2},
    pages = {ii87--ii97},
    year = {2024},

}

@inproceedings{wienöbst2026embracingdiscretesearchreasonable,
      title={Embracing Discrete Search: A Reasonable Approach to Causal Structure Learning},
      author={Wien{\"o}bst, Marcel and Henckel, Leonard and Weichwald, Sebastian},
      booktitle={International Conference on Learning Representations},
      year={2026}
}

@inproceedings{henckel2024adjustment,
  title={Adjustment identification distance: A gadjid for causal structure learning},
  author={Henckel, Leonard and W{\"u}rtzen, Theo and Weichwald, Sebastian},
  booktitle={Conference on Uncertainty in Artificial Intelligence},
  pages={1569--1598},
  year={2024},
  organization={PMLR}
}

@article{perkovic2018complete,
  title={Complete graphical characterization and construction of adjustment sets in {M}arkov equivalence classes of ancestral graphs},
  author={Perkovi{\'c}, Emilija and Textor, Johannes and Kalisch, Markus and Maathuis, Marloes H.},
  journal={Journal of Machine Learning Research},
  volume={18},
  number={220},
  pages={1--62},
  year={2018}
}

@article{andrews2023fast,
  title={Fast scalable and accurate discovery of {DAG}s using the best order score search and grow shrink trees},
  author={Andrews, Bryan and Ramsey, Joseph and Sanchez Romero, Ruben and Camchong, Jazmin and Kummerfeld, Erich},
  journal={Advances in Neural Information Processing Systems},
  volume={36},
  pages={63945--63956},
  year={2023}
}

@inproceedings{nazaret2023stable,
  title={Stable differentiable causal discovery},
  author={Nazaret, Achille and Hong, Justin and Azizi, Elham and Blei, David},
  booktitle={International Conference on Machine Learning},
  pages={37413--37445},
  year={2024},
  organization={PMLR}
}

@article{bello2022dagma,
  title={{DAGMA}: Learning {DAG}s via {M}-matrices and a log-determinant acyclicity characterization},
  author={Bello, Kevin and Aragam, Bryon and Ravikumar, Pradeep},
  journal={Advances in Neural Information Processing Systems},
  volume={35},
  pages={8226--8239},
  year={2022}
}

@article{schwarz1978estimating,
  title={Estimating the Dimension of a Model},
  author={Schwarz, Gideon},
  journal={The Annals of Statistics},
  volume={6},
  number={2},
  pages={461--464},
  year={1978}
}

@inproceedings{lee2019scaling,
  title={Scaling structural learning with {NO-BEARS} to infer causal transcriptome networks},
  author={Lee, Hao-Chih and Danieletto, Matteo and Miotto, Riccardo and Cherng, Sarah T. and Dudley, Joel T.},
  booktitle={Pacific Symposium on Biocomputing 2020},
  pages={391--402},
  year={2019},
  organization={World Scientific}
}

@book{meyer2000matrix,
  title={Matrix Analysis and Applied Linear Algebra},
  author={Meyer, Carl D.},
  volume={71},
  year={2000},
  publisher={SIAM}
}

@article{magnus1985differentiating,
  title={On differentiating eigenvalues and eigenvectors},
  author={Magnus, Jan R.},
  journal={Econometric Theory},
  volume={1},
  number={2},
  pages={179--191},
  year={1985},
  publisher={Cambridge University Press}
}

@article{sullivant2010trek,
    AUTHOR = {Sullivant, Seth and Talaska, Kelli and Draisma, Jan},
     TITLE = {Trek separation for {G}aussian graphical models},
   JOURNAL = {The Annals of Statistics},
  FJOURNAL = {The Annals of Statistics},
    VOLUME = {38},
      YEAR = {2010},
    NUMBER = {3},
     PAGES = {1665--1685},
      optISSN = {0090-5364,2168-8966},
   optMRCLASS = {62H99 (05A15 05C90 62-09 62J05)},
  optMRNUMBER = {2662356},
optMRREVIEWER = {Erhard\ A. J. Godehardt},
       optDOI = {10.1214/09-AOS760},
       optURL = {https://doi-org.tum-eaccess.de/10.1214/09-AOS760},
}

@article{chen2016ancestral,
  title={Learning {B}ayesian Networks with Ancestral Constraints},
  author={Chen, Elias Y.-J. and Shen, Yujia and Choi, Arthur and Darwiche, Adnan},
  journal={Advances in Neural Information Processing Systems},
  volume={29},
  pages={2325--2333},
  year={2016}
}

@inproceedings{li2018sideconstraints,
  title={Bayesian Network Structure Learning with Side Constraints},
  author={Li, Andrew C. and van Beek, Peter},
  booktitle={International Conference on Probabilistic Graphical Models},
  pages={225--236},
  year={2018},
  organization={PMLR}
}

@inproceedings{ban2025differentiable,
  title={Differentiable structure learning with ancestral constraints},
  author={Ban, Taiyu and Rong, Changxin and Wang, Xiangyu and Chen, Lyuzhou and Wang, Xin and Lyu, Derui and Zhu, Qinrui and Chen, Huanhuan},
  booktitle={Forty-second International Conference on Machine Learning},
  year={2025}
}

@article{guo2026linear,
  title={Linear causal discovery with interventional constraints},
  author={Guo, Zhigao and Dong, Feng},
  journal={Machine Learning},
  volume={115},
  number={3},
  pages={35},
  year={2026},
  publisher={Springer}
}

@article{de2025cluster,
  title={Cluster-{DAG}s as Powerful Background Knowledge For Causal Discovery},
  author={de Vargas, Jan Marco Ruiz and Padh, Kirtan and Kilbertus, Niki},
  journal={arXiv preprint arXiv:2512.10032},
  year={2025}
}

@inproceedings{borboudakis2012incorporating,
  title={Incorporating causal prior knowledge as path-constraints in {B}ayesian networks and maximal ancestral graphs},
  author={Borboudakis, Giorgos and Tsamardinos, Ioannis},
  booktitle={International Conference on Machine Learning},
  year={2012}
}

@book{spirtes2000causation,
  title={Causation, prediction, and search},
  author={Spirtes, Peter and Glymour, Clark N. and Scheines, Richard},
  year={2000},
  publisher={MIT Press}
}

@article{zhang2008completeness,
  title={On the completeness of orientation rules for causal discovery in the presence of latent confounders and selection bias},
  author={Zhang, Jiji},
  journal={Artificial Intelligence},
  volume={172},
  number={16-17},
  pages={1873--1896},
  year={2008},
  publisher={Elsevier}
}

@article{chickering2002optimal,
  title={Optimal structure identification with greedy search},
  author={Chickering, David Maxwell},
  journal={Journal of Machine Learning Research},
  volume={3},
  number={Nov},
  pages={507--554},
  year={2002}
}

@inproceedings{zheng2020learning,
  title={Learning sparse nonparametric {DAG}s},
  author={Zheng, Xun and Dan, Chen and Aragam, Bryon and Ravikumar, Pradeep and Xing, Eric},
  booktitle={International Conference on Artificial Intelligence and Statistics},
  pages={3414--3425},
  year={2020},
  organization={PMLR}
}

@inproceedings{amendola2020structure,
  title={Structure learning for cyclic linear causal models},
  author={Am{\'e}ndola, Carlos and Dettling, Philipp and Drton, Mathias and Onori, Federica and Wu, Jun},
  booktitle={Conference on Uncertainty in Artificial Intelligence},
  pages={999--1008},
  year={2020},
  organization={PMLR}
}

@inproceedings{mooij2013cyclic,
  title={Cyclic causal discovery from continuous equilibrium data},
  author={Mooij, Joris and Heskes, Tom},
  booktitle={Conference on Uncertainty in Artificial Intelligence},
  year={2013},
  publisher={AUAI Press}
}

@inproceedings{sethuraman2023nodags,
  title={{NODAGS-F}low: Nonlinear cyclic causal structure learning},
  author={Sethuraman, Muralikrishnna G. and Lopez, Romain and Mohan, Rahul and Fekri, Faramarz and Biancalani, Tommaso and H{\"u}tter, Jan-Christian},
  booktitle={International Conference on Artificial Intelligence and Statistics},
  pages={6371--6387},
  year={2023},
  organization={PMLR}
}

@article{boege2025conditional,
  title={Conditional independence in stationary distributions of diffusions},
  author={Boege, Tobias and Drton, Mathias and Hollering, Benjamin and Lumpp, Sarah and Misra, Pratik and Schkoda, Daniela},
  journal={Stochastic Processes and their Applications},
  volume={184},
  pages={104604},
  year={2025},
  publisher={Elsevier}
}

@article{waxman2024dagma,
  title={{DAGMA-DCE}: Interpretable, non-parametric differentiable causal discovery},
  author={Waxman, Daniel and Butler, Kurt and Djuri{\'c}, Petar M.},
  journal={IEEE Open Journal of Signal Processing},
  volume={5},
  pages={393--401},
  year={2024},
  publisher={IEEE}
}

@inproceedings{liang2024kernel,
  title={Kernel-based differentiable learning of non-parametric directed acyclic graphical models},
  author={Liang, Yurou and Zadorozhnyi, Oleksandr and Drton, Mathias},
  booktitle={International Conference on Probabilistic Graphical Models},
  pages={253--272},
  year={2024},
  organization={PMLR}
}

@inproceedings{clivio2025learningdefercausaldiscovery,
title={Learning to Defer for Causal Discovery with Imperfect Experts},
author={Oscar Clivio and Divyat Mahajan and Perouz Taslakian and Sara Magliacane and Ioannis Mitliagkas and Valentina Zantedeschi and Alexandre Drouin},
booktitle={Workshop on Reasoning and Planning for Large Language Models},
year={2025}
}

@article{reisach2021beware,
  title={Beware of the simulated {DAG}! {C}ausal discovery benchmarks may be easy to game},
  author={Reisach, Alexander and Seiler, Christof and Weichwald, Sebastian},
  journal={Advances in Neural Information Processing Systems},
  volume={34},
  pages={27772--27784},
  year={2021}
}

@article{reisach2023scale,
  title={A scale-invariant sorting criterion to find a causal order in additive noise models},
  author={Reisach, Alexander and Tami, Myriam and Seiler, Christof and Chambaz, Antoine and Weichwald, Sebastian},
  journal={Advances in Neural Information Processing Systems},
  volume={36},
  pages={785--807},
  year={2023}
}

@article{watts1998collective,
  title={Collective dynamics of `small-world' networks},
  author={Watts, Duncan J. and Strogatz, Steven H.},
  journal={Nature},
  volume={393},
  number={6684},
  pages={440--442},
  year={1998},
  publisher={Nature Publishing Group UK, London}
}

@inproceedings{gobler2024texttt,
  title={{CausalAssembly}: Generating Realistic Production Data for Benchmarking Causal Discovery},
  author={G{\"o}bler, Konstantin and Windisch, Tobias and Drton, Mathias and Pychynski, Tim and Roth, Martin and Sonntag, Steffen},
  booktitle={Conference on Causal Learning and Reasoning},
  pages={609--642},
  year={2024},
  organization={PMLR}
}

@article{tagliapietra2025causalman,
  title={{CausalMan}: A physics-based simulator for large-scale causality},
  author={Tagliapietra, Nicholas and Luettin, Juergen and Halilaj, Lavdim and Willig, Moritz and Pychynski, Tim and Kersting, Kristian},
  journal={arXiv preprint arXiv:2502.12707},
  year={2025}
}

\newpage

\appendix
\crefalias{section}{appendix}
\crefalias{subsection}{subappendix}
\crefalias{subsubsection}{subsubappendix}

\title[\savedtitle]{\savedtitle \\ (Supplementary Material)}
\maketitle

\section{Additional Material}
\subsection{Non-Convexity}\label{app_subsec:Non-Convexity}
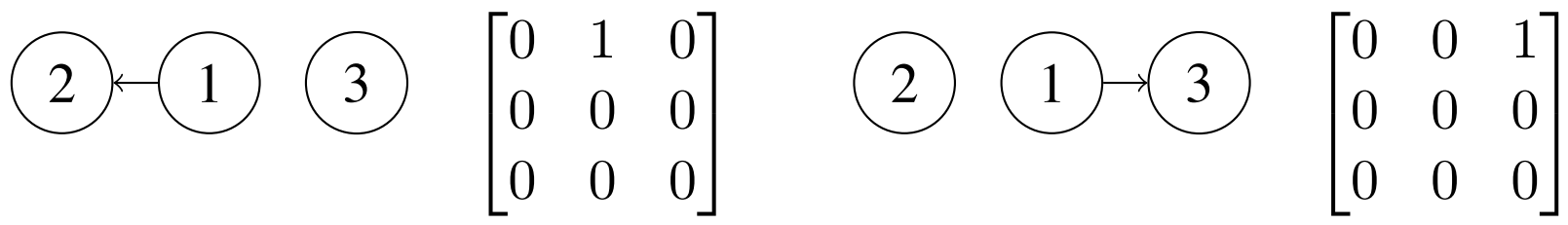
\begin{figure}[ht]
    \centering
    \begin{minipage}{0.2\textwidth}
        \centering
        \begin{tikzpicture}[->, node distance=.5cm]
            \node[circle, draw, minimum size=0.4cm] (1) at (0,0.0) {1};
            \node[circle, draw, minimum size=0.4cm] (2) at (-1, 0) {2};
            \node[circle, draw, minimum size=0.4cm] (3) at (1, 0) {3};
            \draw (1) -> (2);
        \end{tikzpicture}
    \end{minipage}%
    \begin{minipage}{0.15\textwidth}
        \[
        \begin{bmatrix}
        0 & 1 & 0 \\
        0 & 0 & 0 \\
        0 & 0 & 0
        \end{bmatrix}
        \]
    \end{minipage}\hspace{1em}%
    \begin{minipage}{0.2\textwidth}
        \centering
        \begin{tikzpicture}[->, node distance=.5cm]
            \node[circle, draw, minimum size=0.4cm] (1) at (0,0.0) {1};
            \node[circle, draw, minimum size=0.4cm] (2) at (-1, 0) {2};
            \node[circle, draw, minimum size=0.4cm] (3) at (1, 0) {3};
            \draw (1) -> (3);
        \end{tikzpicture}
    \end{minipage}%
    \begin{minipage}{0.15\textwidth}
        \[
        \begin{bmatrix}
        0 & 0 & 1 \\
        0 & 0 & 0 \\
        0 & 0 & 0
        \end{bmatrix}
        \]
    \end{minipage}
    \caption{Example of the non-convexity of the combinatorial constraint of no trek between $2$ and $3$.}
    \label{fig:notreks is nonconvex}
\end{figure}

\begin{figure}[ht]
    \centering
    \includegraphics[width=.7\linewidth]{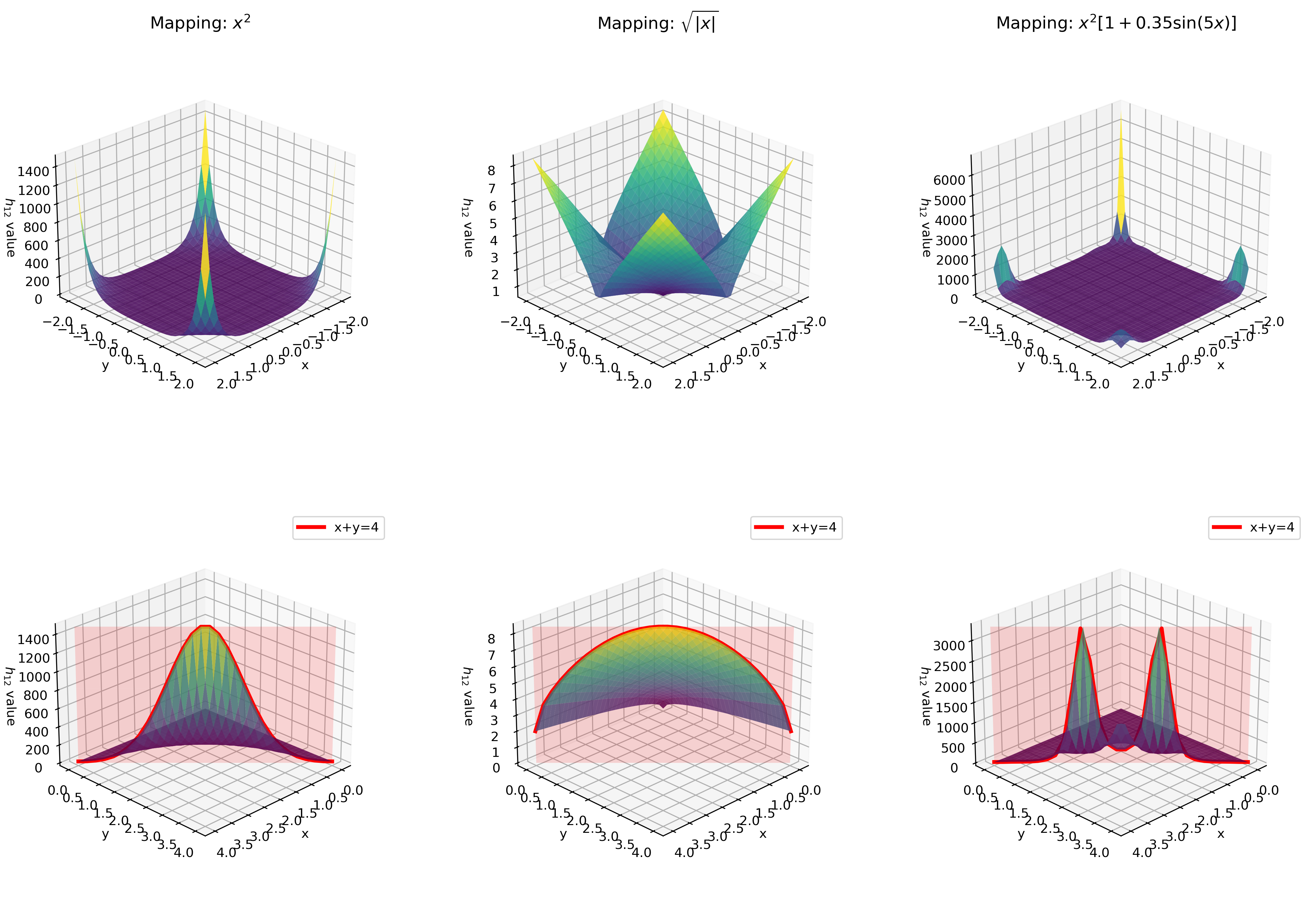}
    \label{fig:surface_adjacency_mappings}
    \caption{First row displays the landscape of $h(W)_{12}$ for $W$ defined as in \cref{eq:example_nonconvex_notreks}. Columnwise different mappings $\phi$ apply. Specifically, the second row shows a slice of that landscape demonstrating visually that PSTrek is not convex in $W$, even if $\phi$ itself is convex.}
\end{figure}

We first illustrate that the hard no-trek constraint is non-convex. The two displayed adjacency matrices contain no trek between nodes $2$ and $3$, but every non-trivial convex combination contains the trek $2\leftarrow1\to3$ (\cref{fig:notreks is nonconvex}).
The smooth PSTrek penalties are not convex either. To illustrate this, let
\begin{equation}\label{eq:example_nonconvex_notreks}
    W = \begin{pmatrix}
    0&x\\y&0
\end{pmatrix},
\end{equation} 
with $x,y\in\R$. \Cref{fig:surface_adjacency_mappings} shows the resulting surfaces for different mappings $\phi$. Non-convexity of these penalties does not preclude the additional optimization properties established below.


\subsection{Binary trek-matrix example}
\label{sec:additional-trek-constraint-material}

Let $B\in\{0,1\}^{d\times d}$ be the adjacency matrix of a DAG and define the path matrix by
\[
    P(B):=\sum_{k=0}^{d-1}B^k.
\]
Because $B$ is acyclic, $P(B)_{ci}$ counts directed paths from $c$ to $i$. The binary Gram-matrix construction of \citet{deligeorgaki2024combinatorial} therefore satisfies
\begin{equation}
    [P(B)^\top P(B)]_{ij}=\sum_{c\in V}P(B)_{ci}P(B)_{cj},
    \label{eq:additional-binary-trek-gram}
\end{equation}
which is positive exactly when $i$ and $j$ have a common ancestor and hence a trek. For example, for
\[
    B=\begin{bmatrix}0&0&1\\0&0&1\\0&0&0\end{bmatrix},
    \qquad P(B)=I+B=\begin{bmatrix}1&0&1\\0&1&1\\0&0&1\end{bmatrix},
\]
we obtain
\[
    P(B)^\top P(B)=\begin{bmatrix}1&0&1\\0&1&1\\1&1&3\end{bmatrix}.
\]
The zero $(1,2)$ entry records the absence of a trek between vertices $1$ and $2$, whereas the $(3,3)$ entry counts three common-source path pairs.

\subsection{NOTREKS Combinatorial Viewpoint}\label{subsec:NOTREKS Combinatorial Viewpoint}

\subsubsection{Coloring lower bound}
The main text derives a lower bound from one independent color class. Here we use all color classes simultaneously to obtain a stronger bound for NOTREKS information \(H=(V,\cal{I})\).

Let \(a_m\) denote the number of labeled DAGs on \(m\) vertices. We can bound it as follows
\begin{equation}\label{eq:number of DAGs}
    2^{\binom{m}{2}} \leq a_m \leq m!\,2^{\binom{m}{2}}.
\end{equation}
The lower bound follows by fixing one topological order and freely choosing any subset of its \(\binom{m}{2}\) admissible forward edges, while the upper bound allows all \(m!\) topological orders and may therefore count the same DAG multiple times. 

\begin{figure}[htbp]
    \centering
    \tikzset{
        vertex/.style={
            circle,
            draw,
            minimum size=0.48cm,
            inner sep=0pt,
            font=\small
        },
        classone/.style={
            vertex,
            fill=blue!18,
            draw=blue!70!black
        },
        classtwo/.style={
            vertex,
            fill=orange!25,
            draw=orange!80!black
        },
        notrek/.style={
            draw=notrekpurple,
            line width=1.1pt 
        },
        remaining/.style={
            draw=gray!45,
            line width=0.45pt
        }
    }

    \begin{minipage}[t]{0.22\textwidth}
        \centering
        \vspace{0pt}
        \begin{tikzpicture}[->, >=stealth]
            \node[vertex] (1) at (-1.0, 0.65) {1};
            \node[vertex] (2) at (-1.0,-0.65) {2};
            \node[vertex] (3) at ( 0.0, 0.65) {3};
            \node[vertex] (4) at ( 0.0,-0.65) {4};
            \node[vertex] (5) at ( 1.0, 0.65) {5};
            \node[vertex] (6) at ( 1.0,-0.65) {6};

            \draw (1) -> (2);
            \draw (3) -> (4);
            \draw (5) -> (6);
        \end{tikzpicture}

        \smallskip
        {\small\textbf{(a)} Original DAG \(G\)}
    \end{minipage}
    \hfill
    \begin{minipage}[t]{0.25\textwidth}
        \centering
        \vspace{0pt}
        \begin{tikzpicture}
            \node[vertex] (1) at (-1.0, 0.65) {1};
            \node[vertex] (2) at (-1.0,-0.65) {2};
            \node[vertex] (3) at ( 0.0, 0.65) {3};
            \node[vertex] (4) at ( 0.0,-0.65) {4};
            \node[vertex] (5) at ( 1.0, 0.65) {5};
            \node[vertex] (6) at ( 1.0,-0.65) {6};

            \draw[remaining] (1) to[bend left=12]  (4);
            \draw[remaining] (2) to[bend right=12] (3);
            \draw[remaining] (3) to[bend left=12]  (6);
            \draw[remaining] (4) to[bend right=12] (5);
            
            \draw[remaining] (1) to[bend left=28]  (5);
            \draw[remaining] (2) to[bend right=28] (6);
            
            \draw[notrek] (1) -- (3);
            \draw[notrek] (3) -- (5);
            \draw[notrek] (2) -- (4);
            \draw[notrek] (4) -- (6);
            \draw[notrek] (1) -- (6);
            \draw[notrek] (2) -- (5);
        \end{tikzpicture}

        \smallskip
        {\small\textbf{(b)} \(T(G)^c\) and
        \(\mathcal I\subseteq E(T(G)^c)\)}
    \end{minipage}
    \hfill
    \begin{minipage}[t]{0.22\textwidth}
        \centering
        \vspace{0pt}
        \begin{tikzpicture}
            \node[classone] (1) at ( 0.00, 1.00) {1};
            \node[classtwo] (3) at ( 0.87, 0.50) {3};
            \node[classone] (5) at ( 0.87,-0.50) {5};
            \node[classtwo] (2) at ( 0.00,-1.00) {2};
            \node[classone] (4) at (-0.87,-0.50) {4};
            \node[classtwo] (6) at (-0.87, 0.50) {6};

            \draw[notrek] (1) -- (3);
            \draw[notrek] (3) -- (5);
            \draw[notrek] (5) -- (2);
            \draw[notrek] (2) -- (4);
            \draw[notrek] (4) -- (6);
            \draw[notrek] (6) -- (1);
        \end{tikzpicture}

        \smallskip
        {\small\textbf{(c)} Proper coloring of
        \(H=(V,\mathcal I)\)}
    \end{minipage}
    \hfill
    \begin{minipage}[t]{0.25\textwidth}
        \centering
        \vspace{0pt}
        \begin{tikzpicture}[->, >=stealth]
            \node[classone] (1) at (-0.85, 0.75) {1};
            \node[classone] (5) at (-1.25,-0.35) {5};
            \node[classone] (4) at (-0.45,-0.35) {4};

            \node[classtwo] (3) at ( 0.85, 0.75) {3};
            \node[classtwo] (2) at ( 0.45,-0.35) {2};
            \node[classtwo] (6) at ( 1.25,-0.35) {6};

            \draw (1) -> (5);
            \draw (1) -> (4);
            \draw (5) -> (4);

            \draw (3) -> (2);
            \draw (3) -> (6);
            \draw (2) -> (6);

            \draw[
                dashed,
                rounded corners,
                blue!70!black
            ] (-1.58,-0.70) rectangle (-0.12,1.15);

            \draw[
                dashed,
                rounded corners,
                orange!80!black
            ] (0.12,-0.70) rectangle (1.58,1.15);

            \node[
                blue!70!black,
                font=\small
            ] at (-0.85,1.35) {\(C_1\)};

            \node[
                orange!80!black,
                font=\small
            ] at (0.85,1.35) {\(C_2\)};
        \end{tikzpicture}

        \smallskip
        {\small\textbf{(d)} A feasible DAG \(\widetilde G\)}
    \end{minipage}

    \caption{
        Illustration of the coloring lower bound.
        The original DAG \(G\) in \textbf{(a)} has three disconnected
        components. In \textbf{(b)}, the gray edges form \(T(G)^c\),
        while the red edges constitute the selected NOTREKS information
        \(\mathcal I\). The graph \(H=(V,\mathcal I)\) is properly
        colored in \textbf{(c)}, with
        \(C_1=\{1,4,5\}\) and \(C_2=\{2,3,6\}\).
        Placing an arbitrary DAG on each color class and adding no edges
        between the classes produces a DAG satisfying all constraints in
        \(\mathcal I\), as illustrated by complete DAGs in \textbf{(d)}.
        It need not satisfy all no-trek relations in \(T(G)^c\):
        for example, \(1\to5\) introduces a trek between a pair in
        \(E(T(G)^c)\setminus\mathcal I\).
    }
    \label{fig:notreks-coloring-lower-bound}
\end{figure}
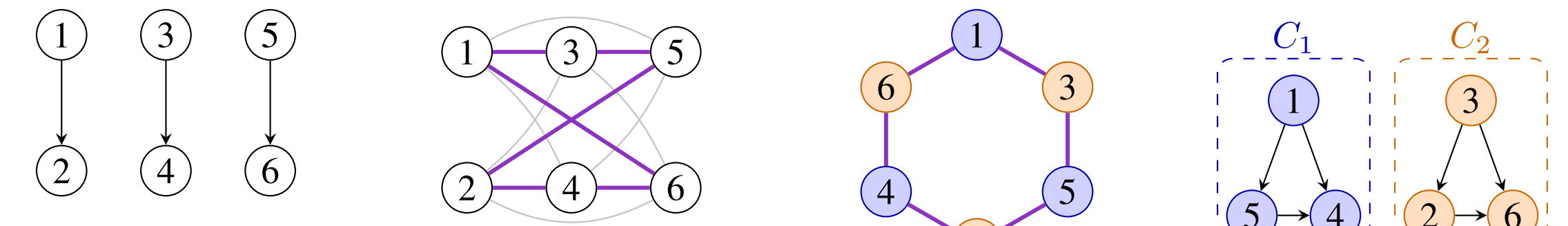

For any proper $r$-coloring \(V=C_1\dot\cup\cdots\dot\cup C_r\) of $H$, arbitrary DAGs may be placed on each of the color classes, without adding edges between them. Adjacent nodes in $H$ are given different colors and it follows that such graphs do not introduce treks between \(H\)-adjacent nodes. Hence, they are feasible with respect to the NOTREKS information. This is visualized in \cref{fig:notreks-coloring-lower-bound}. Consequently,
\begin{equation}\label{eq:log NH lower bound}
    \begin{aligned}
        \log_2 N(H)
        &\geq \log_2 \prod_{\ell=1}^r a_{\lvert C_\ell\rvert}
        \geq \sum_{\ell=1}^r \binom{\lvert C_\ell\rvert}{2}
        = \frac{1}{2}
        \left(
            \sum_{\ell=1}^r \lvert C_\ell\rvert^2-d
        \right)\\
        &\geq \frac{1}{2}
        \left(
            \frac{d^2}{r}-d
        \right),
    \end{aligned}
\end{equation}
where the last inequality follows from Jensen's inequality. For $d \to \infty$ we get $\log_2 N(H) =\Omega(\frac{d^2}{r})$, whereas $\log_2(a_d) =\Omega(d^2)$. Thus, $r=\chi(H)$ yields the strongest lower bound of this coarse form,
\begin{equation}
    \log_2N(H)\geq
    \frac12\left(\frac{d^2}{\chi(H)}-d\right).
    \label{eq:app-combinatorial-chromatic-bound}
\end{equation}
This does not assert that a minimum coloring maximizes the sharper $\log_2 \prod_{\ell=1}^r a_{\lvert C_\ell\rvert}$ in \cref{eq:log NH lower bound}, which also depends on the individual color-class sizes.

\subsubsection{Example: Lower Bound for \texorpdfstring{$k$}{k}-Erd\H{o}s--R\'enyi DAGs}\label{subsubsec:example_lower_bound_for_ER}

We illustrate the size of the previously obtained lower bound on $N(H)$. Let $G_d$ be a $k$-Erd\H{o}s--R\'enyi DAG on $d$ nodes: fix a topological order and include every forward edge independently with~probability
\begin{equation}\label{eq:app-combinatorial-er-probability}
    p=\frac{kd/2}{\binom d2}=\frac{k}{d-1},
\end{equation}
where $k>0$ is fixed such that $p\leq1$.  Thus $G_d$ has $kd/2$ expected edges and expected undirected degree $k$ (and expected average indegree $k/2$). Let $H_d = T(G_d)^c = (V,\mathcal I_d)$, which means that $H_d$ is the complete NOTREKS information graph. As throughout the paper, this is
\begin{equation}
    \{i,j\}\in\mathcal I_d
    \quad\Longleftrightarrow\quad
    \text{there is no trek between $i$ and $j$ in $G_d$}.
    \label{eq:app-combinatorial-complete-information}
\end{equation}
Let $1_i$ indicate that vertex $i$ in the fixed topological order is a source of $G_d$ with $\chi(H_d) = \sum_{i \in V} 1_i$ following from \cref{cor:treks_and_sources}. Due to the construction of ER-graphs, $1_i \sim \mathrm{Ber}((1-p)^{i-1})$ and independent. We calculate,
\begin{equation}
    \begin{aligned}
    \E\left[\chi(H_d)\right] &=\sum_{i=1}^d(1-p)^{i-1}
    =\frac{1-(1-p)^d}{p}
    =\frac{d-1}{k}\left[1-\left(1-\frac{k}{d-1}\right)^d\right].
    \end{aligned}
\end{equation}
It holds
\begin{equation}\label{eq:chromatic_number_bound+conv}
    \frac{1}{d}\E\left[\chi(H_d) \right] \underset{d\to\infty}{\to} \frac{1}{k} \left( 1-e^{-k} \right) =: s_k,
\end{equation}
and hence, for every $\varepsilon>0$, there is a $d_0$ such that for all $d\geq d_0$,
\[
    \left| \frac{1}{d}\E\left[\chi(H_d) \right]-s_k\right|
    \leq \frac{\varepsilon}{2}.
\]
Under this event, using Chebyshev's inequality, one calculates
\begin{equation}
    \begin{aligned}
        \mathbb P\left(
            \left|
                \frac{1}{d}\chi(H_d)-s_k
            \right|>\varepsilon
        \right) &= \mathbb P\left(
            \left|
                \frac{1}{d}\chi(H_d) -\frac{1}{d}\E[\chi(H_d)] + \frac{1}{d}\E[\chi(H_d)] - s_k
            \right|>\varepsilon
        \right)\\
        &\leq \mathbb P\left(
            \left|
                \frac{1}{d}\chi(H_d) -\frac{1}{d}\E[\chi(H_d)]\right| + \left|\frac{1}{d}\E[\chi(H_d)] - s_k
            \right|>\varepsilon
        \right)\\
        &\leq
        \mathbb P\left(
            \left|
                \chi(H_d)-\E[\chi(H_d)]
            \right|>\frac{\varepsilon d}{2}
        \right)\\
        &\leq
        \frac{4\var(\chi(H_d))}{\varepsilon^2d^2}\\
        &\leq
        \frac{4}{\varepsilon^2d}
        \underset{d\to\infty}{\longrightarrow}0,
    \end{aligned}
\end{equation}
where the last inequality followed from the independence of the indicators and because $1_i$ Bernoulli distributed $\var{[1_i]} \leq \E[1_i]$. Specifically, $$\var{[\chi(H_d)]} = \sum_i (1-p)^{i-1} (1-(1-p)^{i-1}) \leq \sum_i (1-p)^{i-1} = \E[\chi(H_d)] \leq d.$$
Thus, we derive convergence in probability, i.e.
$$\frac{1}{d}\chi(H_d) \xrightarrow{\P} s_k.$$
Now fix $\varepsilon\in(0,1-s_k)$. Then, with probability tending to one, $\chi(H_d)\leq(s_k+\varepsilon)d$. Using \cref{eq:log NH lower bound}, we obtain on this event
\begin{equation}\label{eq:lower_bound_sk}
    \begin{aligned}
        \log_2N(H)
        &\geq
        \frac{1}{2}
        \left(
            \frac{d^2}{\chi(H_d)}-d
        \right)
        \geq
        \frac{d}{2}
        \left(
            \frac{1}{s_k+\varepsilon}-1
        \right).
    \end{aligned}
\end{equation}
Since $s_k+\varepsilon<1$, the constant on the right-hand side is strictly positive. Formally, we write $\log_2N(H)=\Omega(d)$ with probability tending to one.

\subsubsection{Example: Upper Bound for \texorpdfstring{$k$}{k}-Erdős--Rényi Graphs}
Again let $H_d = T(G_d)^c$, and $G_d$ is the $k$-ER graph. For a DAG $\widetilde G = (V, \widetilde E)$ every edge introduces a trek of length one, which means $\operatorname{skel}(\widetilde G)\subseteq T(\widetilde G)$. $\widetilde G$ is feasible with respect to the complete NOTREKS information if and only if $T(\widetilde G) \subseteq T(G_d)$. We conclude that for any feasible DAG it holds $\operatorname{skel}(\widetilde G)\subseteq T(G_d)$.

Let $E(\cdot)$ denote the edge set of the graph-argument. For every skeleton \(F\subseteq E(T(G_d))\), there are at most \(2^{\abs F}\) orientations to construct a DAG with that skeleton. Consequently,
$$N(H) \leq \sum_{F\subseteq E(T(G_d))}2^{\abs F} = 3^{\abs{E(T(G_d))}}.$$
This count includes cyclic orientations and orientations that create additional forbidden treks, so the upper bound is generally not tight.

It remains to control the size of \(T(G_d)\). For that we fix a topological order of $V$ and use the node index as the position in the causal order. We count how many paths we expect to be created in the construction of an $k$-ER graph. Let \(Q_i\) denote the number of directed paths starting at node \(i\), including the path of length zero. Writing \(B_{ij}\) for the indicator of \(i\to j\), every nontrivial path has a unique first edge, and hence
$$Q_i=1+\sum_{j=i+1}^dB_{ij}Q_j.$$
Every edge of \(T(G_d)\) is witnessed by two distinct directed paths starting at a common ancestor. Therefore,
\begin{equation}\label{eq:trek_count_upper_bound}
    \abs{E(T(G_d))} \leq \sum_{i=1}^d\binom{Q_i}{2} \leq \frac{1}{2}\sum_{i=1}^dQ_i^2.
\end{equation}

\paragraph{Calculating $\E[Q_i^2]$.}
Due to topological ordering of $B$ and its acyclicity with $i < j$, \(B_{ij}\) is independent from \(Q_j\) which gives
\begin{equation}
    \begin{aligned}
        \E[Q_i] &= 1+p\sum_{j=i+1}^d\E[Q_j] = 1 + p\E[Q_{i+1}] + (\E[Q_{i+1}] - 1)= (1+p) \E[Q_{i+1}].
    \end{aligned}
\end{equation}
Here $p$ is the probability of \(i\to j\) existing \cref{eq:app-combinatorial-er-probability}.
By backward induction with $\E[Q_d] = 1$, we get $\E[Q_i] = (1+p)^{d-i}$. Using $1 + x \leq e^x$ and substituting $p = \frac{k}{d-1}$, we get
\begin{equation}
     \E[Q_i] = (1+p)^{d-i} \leq (1+p)^{d-1} \leq \exp{\left((d-1)p\right)} = e^{k}.
\end{equation}
Now, for \(\E[Q_i^2]\), expanding the square and using the basic fact that $2Q_jQ_l \leq Q_j^2 + Q_l^2$, gives 
\begin{equation}
    \begin{aligned}
        \E[Q_i^2]
        &= \E\left[1 + 2\sum_{j=i+1}^dB_{ij}Q_j + \sum_{j=i+1}^d \sum_{l=i+1}^d B_{ij} B_{il} Q_j Q_l \right]\\
        &\leq 2 \E[Q_i] + \E\left[ \sum_{j=i+1}^d  B_{ij}^2 Q_j^2\right] + \E\left[ \sum_{j=i+1}^d \sum_{l > j}^d B_{ij} B_{il} (Q_j^2 + Q_l^2)\right]\\
        &= 2 \E[Q_i] + \left(p+p^2(d-i-1)\right) \sum_{j=i+1}^d\E[Q_j^2]\\
        &\leq 2e^{k} + \frac{k+k^2}{d-1} \sum_{j=i+1}^d\E[Q_j^2].
    \end{aligned}
\end{equation}
Backward induction with $\E[Q_d^2] = 1$ yields
\begin{equation}\label{eq:second_moment_path_count}
    \E[Q_i^2] \leq 2e^{k} \left( 1+\frac{k+k^2}{d-1} \right)^{d-i} \leq 2e^{2k+k^2}.
\end{equation}
With this we jump back to \cref{eq:trek_count_upper_bound}. Taking expectations and using \cref{eq:second_moment_path_count} gives $\E[\abs{E(T(G_d))}] \leq d\;e^{2k+k^2}$. For $a>0$, Markov's inequality implies
\begin{equation}
    \P(\abs{E(T(G_d))} > a d)
    \leq
    \frac{\E[\abs{E(T(G_d))}]}{ad}
    \leq
    \frac{e^{2k+k^2}}{a}.
\end{equation}
Hence, for every $\delta \in (0,1)$ and $d \in \mathbb{N}_{>0}$, choosing $a=\delta^{-1}e^{2k+k^2}$ gives
$$
\P\left(
    \frac{1}{d}\abs{E(T(G_d))}
    >
    \delta^{-1}e^{2k+k^2}
\right)
\leq
\delta.
$$
We denote this with $\abs{E(T(G_d))}=O_{\mathbb P}(d)$. Consequently, $$\log_2N(H) \leq \log_2(3)\abs{E(T(G_d))} = O_{\mathbb P}(d).$$
To bring the lower bound into a similar form, for $\eps > 0$ sufficiently small, choose $0<c_k<\frac{1}{2}\left(\frac{1}{s_k + \eps}-1\right)$. For every $\delta\in(0,1)$ and all sufficiently large $d$, combining the two
bounds gives
\begin{equation}
    \mathbb P\left(
        c_k\leq\frac1d\log_2N(H_d)\leq C_{k,\delta}
    \right)\geq1-\delta-o(1),
    \label{eq:app-combinatorial-probability-combination}
\end{equation}
where $C_{k,\delta}:=\delta^{-1}\log_2(3)e^{2k+k^2}$. Therefore, $\log_2N(H_d)=\Theta_{\mathbb P}(d).$

\section{Proofs and Additional Results}

\subsection{Treks and Sources}
\begin{corollary}\label[corollary]{cor:treks_and_sources}
    Let $G$ be a DAG, and $H=(V, \cal{I}) \subseteq T(G)^c$. Then it holds $\chi(H) \leq \abs{\operatorname{src}(G)}.$
    Moreover, sufficient for equality is $H=T(G)^c$.
\end{corollary}
\begin{proof}
    Note that $\operatorname{src}(G) = \{v \in V \mid \pa{v} = \emptyset\}$.
    A coloring of $H$ with $\abs{\operatorname{src}(G)}$ is constructed by coloring all sources of $G$ distinctly and assigning every other node the color of one of its source ancestors. Specifically, assign $v$ a color from $\{c(w) \mid w \in \operatorname{src}(G) \cap \an{v}\}$, where $c$ denotes the color map. Consequently, for any $v \neq w$ with $\{v,w\} \in \cal I$, then $\an{v} \cap \an{w} = \emptyset$ and $c(v) \neq c(w)$.
    
    Conversely, the sources of \(G\) form a clique in \(T(G)^c\), which proves the equality under $H = T(G)^c$.
\end{proof}

\subsection{Trek expansion and characterization}\label{subsec:PSTrek well-defined}
The following expansion establishes well-definedness and will also be used in the proof of the characterization theorem.
On $\mathcal D_{a,\phi}$, the non-negative matrix series
$f_a(\phi(W))=\sum_{k\geq0}a_k\phi(W)^k$ converges absolutely because
$\rho(\phi(W))<R_a$. Entrywise non-negativity then allows us to multiply
the two matrix power series term by term. This gives
\begin{equation}\label{eq:sum_of_treks}
    \begin{aligned}
        h_{a,\phi}(W)_{ij}
    &=\sum_{r,q\geq0}a_ra_q\bigl((\phi(W)^r)^\top \phi(W)^q\bigr)_{ij}\\
    &=\sum_{k=1}^d\sum_{r,q\geq0}
      a_ra_q(\phi(W)^r)_{ki}(\phi(W)^q)_{kj}\\
    &= \sum_{\tau = (\pi_L, \pi_R) \in \cal{T}_{ij}} a_{\abs{\pi_L}}a_{\abs{\pi_R}} \prod_{(v_k, v_{k+1})\in \tau} \phi(W_{v_k,v_{k+1}}),
    \end{aligned}
    \end{equation}
where the second line inherently aggregates treks by the length of their left $\abs{\pi_L}$ and right arms $\abs{\pi_R}$, whereas the last line sums over all individual treks.

\begin{proof}\label{proof:th:PSTrek characterization of treks_revised}
\textbf{ of characterization ~\cref{th:PSTrek characterization of treks}.}
\begin{itemize}
    \item[(i)] 
The power series $f_a$ induces an analytic matrix function on
\[
\mathcal D_{a,\phi}=\set{W\in\R^{d\times d}\mid\rho(\phi(W))<R_a}.
\]
Since $\phi$ is continuously differentiable, the composition
\[
    W\longmapsto f_a(\phi(W))^\top f_a(\phi(W))
\]
is continuously differentiable on $\mathcal D_{a,\phi}$.

For the entrywise derivative, we use the last line of \cref{eq:sum_of_treks}. If the edge $(k,l)$ occurs $m(k,l;\tau)$ times in $\tau$, then $M_\tau(W)$ includes the factor $\phi(W_{kl})^{m(k,l;\tau)}$ and we
\begin{equation}
    \frac{\partial}{\partial W_{kl}}M_\tau(W)
    =
    m(k,l;\tau)a_{\abs{\pi_L}}a_{\abs{\pi_R}}
    \phi^\prime(W_{kl})\phi(W_{kl})^{m(k,l;\tau)-1}
    \prod_{(u,v)\neq(k,l)}
    \phi(W_{uv})^{m(u,v;\tau)}.
\end{equation}
where the product is over the distinct edge pairs occurring in $\tau$ other than $(k,l)$. Summing these derivatives over all treks, which use the $(k,l)$-edge, gives
\begin{equation}\label{eq:gradient PSTrek}
    \frac{\partial}{\partial W_{kl}}h_{a, \phi}(W)_{ij}
    =
    \sum_{\substack{
        \tau\in\mathcal T_{ij}\\
        (k,l)\in \tau
    }}
    \frac{\partial}{\partial W_{kl}}M_\tau(W).
\end{equation}
At $W_{kl}=0$, this expression remains well-defined. For mappings with $\phi(0)=\phi^\prime(0)=0$, every such derivative vanishes.

\item[(ii)] We use the second line representation in \cref{eq:sum_of_treks}: A positive summand consists of a directed
$k$--$i$ walk of length $r$ and a directed $k$--$j$ walk of length $q$;
together they form an $i$--$j$-trek in $G_A=G_W$.  Hence positivity of the
PSTrek constraint implies the existence of a trek.

Conversely, suppose that an $i$--$j$-trek exists.  Removing directed cycles
from each side of the trek gives directed paths from a common source $k$ to
$i$ and $j$, each of length at most $d-1$.  Let their lengths be $r$ and $q$.
All edges on these paths have strictly positive $\phi$-weight and
$a_r a_q>0$.  The corresponding summand in \cref{eq:sum_of_treks} is therefore
strictly positive.  Since every summand in \cref{eq:sum_of_treks} is non-negative, no cancellation is possible, and
$h_{a,\phi}(W)_{ij}>0$.
\end{itemize}
\end{proof}

\subsection{Proof of Theorem~\ref{th:PSTrek constraints are invex}}\label{proof:PSTrek constraints are invex}

We use invexity in the differentiable inequality sense: a function $F$ is invex if there exists a map
$\eta:\mathcal D_{a,\phi}\times\mathcal D_{a,\phi}\to\R^{d\times d}$ such that
\[
    F(U)-F(W)\geq\inner{\grad F(W)}{\eta(U,W)}
    \qquad\text{for all }U,W\in\mathcal D_{a,\phi}.
\]
We use the following standard characterization: a differentiable function is invex for some choice of the map $\eta$ if and only if each of its stationary points is a global minimizer.

We work under the radial admissibility condition on $\phi$ from \cref{th:PSTrek constraints are invex}. This condition rules out positive-valued stationary points without requiring a uniform lower bound on the radial derivative.

\begin{proof}
Using the trek expansion above, consider the contribution of a trek $\tau$:
\[
    M_\tau(W) =a_{\abs{\pi_L}}a_{\abs{\pi_R}}
      \prod_{(u,v)\in \tau}\phi(W_{uv}),
\]
where edges are included according to their multiplicity in $T$. We make use of the product rule in the form of $\frac{\partial}{\partial s} g(sx) = \inner{\grad g(sx)}{x}$, and specifically $\left.\frac{\partial}{\partial s} g(sx)\right|_{s=1} = \inner{\grad g(x)}{x}$. This gives
\[
    \left.{\frac{\partial}{\partial s}M_\tau(sW)}\right|_{s=1}
    =a_{\abs{\pi_L}}a_{\abs{\pi_R}}
      \sum_{e\in \tau}W_e\phi'(W_e)
      \prod_{\substack{\widetilde e\in \tau\\\widetilde e\neq e}}
      \phi(W_{\widetilde e}),
\]
where both the sum and product include edge occurrences according to their
multiplicity.  Every summand is non-negative.  Indeed, radial admissibility
gives $W_e\phi'(W_e)>0$ when $W_e\neq0$, whereas
$\phi'(0)=0$ because zero is a minimum of the differentiable non-negative
function $\phi$.  Moreover, the displayed derivative is strictly positive
whenever $M_\tau(W)>0$.

Write $F(\cdot) = h_{a, \phi}(\cdot)_{ij}$. Local uniform convergence of the power series and its derivative allows us to sum over all treks and obtain
\begin{equation}\label{eq:revised-radial-pstrek}
    F(W)>0
    \quad\Longrightarrow\quad
    \inner{\grad F(W)}{W}
    =\left.\frac{d}{d s}F(sW)\right|_{s=1}>0.
\end{equation}
By \cref{def:PSTrek family} and \cref{th:PSTrek characterization of treks}(ii), $F(W)\geq0$ and
$F(0)=0$. Consequently with \cref{eq:revised-radial-pstrek}, $\grad F(W)=0$ implies $F(W)=0$. Thus every
stationary point is a global minimizer.

Let $F_\cal{I}$ be a non-negative linear combination of $h_{a, \phi}(\cdot)_{ij}$, the radial derivative is the corresponding non-negative linear combination of the individual radial derivatives.  If $F_\cal{I}(W)>0$, at least one entry with a positive coefficient is positive, and the previous proof shows that $\inner{\grad F_\cal{I}(W)}{W}>0$.

The last part of the theorem is \cref{cor:PSTrek + PSTrace is Invex}, for which we first formally define the Power Series Trace constraint.
\end{proof}

\subsubsection{Extension to PSTrace}
\begin{definition}[Power Series Trace constraint]\label[definition]{def:Power Series Trace constraint}\cite[adapted from Def. 3.1]{nazaret2023stable}
Consider the setup of \cref{def:PSTrek family}. Then, on $\mathcal D_{a,\phi}$, the \emph{Power Series Trace} function is defined as 
\begin{equation}\label{eq:def of PSTrace family}
    g_{a, \phi}(W) = \operatorname{trace}\left( f_a(\phi(W))\right) - a_0 d.
\end{equation}
\end{definition}
By absolute convergence of \cref{eq:def of PSTrace family}, we can rewrite it with respect to the cycles $\cal{C}$
\begin{equation}
    g_{a, \phi}(W) = \sum_{C \in \cal{C}} a_{\abs{C}} \prod_{(u,v) \in C} \phi(W_{uv}).
\end{equation}
From here, the same proof applies as before.

\begin{corollary}[PSTrek + PSTrace is Invex]\label[corollary]{cor:PSTrek + PSTrace is Invex}
    Any non-negative linear combination of PSTrek and PSTrace constraints are invex again. This holds for independent choices of sequences and radial admissible adjacency maps, on the intersection of their convergence domains.
\end{corollary}

\subsubsection{Further consequences}
\begin{proposition}[Sharpness of radial admissibility]
\label[proposition]{prop:sharpness-radial-admissibility}
Let $\phi$ be an admissible adjacency mapping and consider PSTrek constraints
whose coefficient sequence satisfies $a_0a_1>0$.  Every such PSTrek
constraint has only globally minimizing stationary points if and only if
$\phi$ is radially admissible.
\end{proposition}

\begin{proof}
Sufficiency follows from the proof of \cref{th:PSTrek constraints are invex}.  For necessity, consider $d=2$ and the graph $1 \rightarrow 2$ with edge weight $x$. The adjacency matrix is $W(x)=xE_{12}$.  Since $\phi(W(x))^2=0$, we have
\[
    h_{a,\phi}(W(x))_{12}=a_0a_1\phi(x).
\]
If $\phi'(x_0)=0$ for some $x_0\neq0$, then $W(x_0)$ is a stationary point with positive constraint value, whereas the constraint vanishes at $W(0)$. Thus $\phi'$ cannot vanish away from zero.

It remains to determine its sign.  For $x>0$, the mean value theorem applied to $\phi$ on $[0,x]$ gives a point $c\in(0,x)$ with $\phi'(c)=\phi(x)/x>0$.
Since $\phi'$ has no zero on $(0,\infty)$, it follows that $\phi'(x)>0$ for all $x>0$.  Similarly, applying the mean value theorem on $[x,0]$ for $x<0$ gives a point with negative derivative, and hence $\phi'(x)<0$ for all $x<0$.  Therefore $x\phi'(x)>0$ for every $x\neq0$.
\end{proof}

\begin{remark}[Weak Polyak--\L{}ojasiewicz inequality]
Suppose, more strongly, that there exists a constant $\kappa_\phi>0$ such
that
\begin{equation}\label{eq:uniform-radial-admissibility}
    x\phi'(x)\geq\kappa_\phi\phi(x)
    \qquad\text{for all }x\in\R.
\end{equation}
Repeating the product-rule argument above gives the quantitative inequality
\[
    \inner{\grad F(W)}{W}
    \geq\kappa_\phi F(W).
\]
Consequently,
\[
    \kappa_\phi
    \bigl(F(W)-F(0)\bigr)
    \leq
    \norm{\grad F(W)}_F\norm{W-0}_F,
\]
which is the weak Polyak--\L{}ojasiewicz inequality with $\mu=\kappa_\phi^2$ \citep{csiba2017global}. Thus the squared parametrization $\phi(x)=x^2$ satisfies the stated weak Polyak--\L{}ojasiewicz inequality with $\mu=4$. This quantitative inequality is stronger than the stationary-point characterization used in the invexity proof and provides a quantitative gradient bound for the penalty itself; it does not hold for an arbitrary invex function.
\end{remark}
\subsection{Convergence to the feasible region}\label{proof:cor:convergence_to_feasible_region}
We analyze the idealized path-following continuation scheme used by \citet{bello2022dagma}.

The finite output of \cref{alg:dagma-notreks} need not be exactly
feasible.  The precise conclusion is an asymptotic one.
\begin{proof}[{Proof of Proposition~\ref{cor:convergence_to_feasible_region}}]
Continuity and compactness imply that $L$ is bounded on $\mathcal X$.  Let $F(W) := g_{a, \phi}(W)\,+\,h_{a, \phi}(W)_{\cal{I}}$. By construction, $F$ is continuous, nonnegative, and satisfies $F(0)=0$. By assumption $W^{(t)}$ is a global minimizer of the full objective,
\[
    \mu^{(t)}L(W^{(t)})+F(W^{(t)})
    \leq \mu^{(t)}L(0).
\]
It follows that
\[
    0\leq F(W^{(t)})
    \leq
    \mu^{(t)}\bigl(L(0)-\inf_{W\in\mathcal X}L(W)\bigr)
    \longrightarrow0.
\]
Let $W^{(t_k)}\to W^\star$ be any convergent subsequence.  By continuity of
$F$, we obtain $F(W^\star)=0$. By the definition of $F$ and the definitions of their respective constraints $W^\star$ is acyclic and has no trek-violations with respect to $\cal{I}$.
\end{proof}

When $\mu=0$, every stationary point of $F$ is a global minimizer and therefore feasible. For $\mu>0$, invexity of $F$ alone gives no analogous guarantee for stationary points of the score-augmented objective $F+\mu L$; the proposition concerns the idealized continuation limit rather than finite inner optimization.

\section{Additional Experimental Details}
\label{sec:appendix_experimental_details}

\subsection{Finite-Sample Motivation for Hard NOTREKS Constraints}
\label{app:notreks-motivation}

We complement the motivating result in the main text (\cref{fig:motivation-notreks-feasible-dags}) with a small globally solvable experiment at $d=10$ and sample sizes $n\in\{100,500\}$. Data are generated from the linear-Gaussian SEM defined in the main text. For each true DAG $G$, we construct the oracle no-trek set
\[
\mathcal I(G)=\left\{(i,j):\operatorname{an}_{G}(i)\cap\operatorname{an}_{G}(j)=\varnothing\right\}.
\]
Five independently ordered draws from $\mathcal I(G)$ define nested prior budgets containing no pair, one pair, $\lceil0.1|\mathcal I(G)|\rceil$ pairs, or all available pairs.

\begin{figure}[t]
\centering
\includegraphics[width=\textwidth]{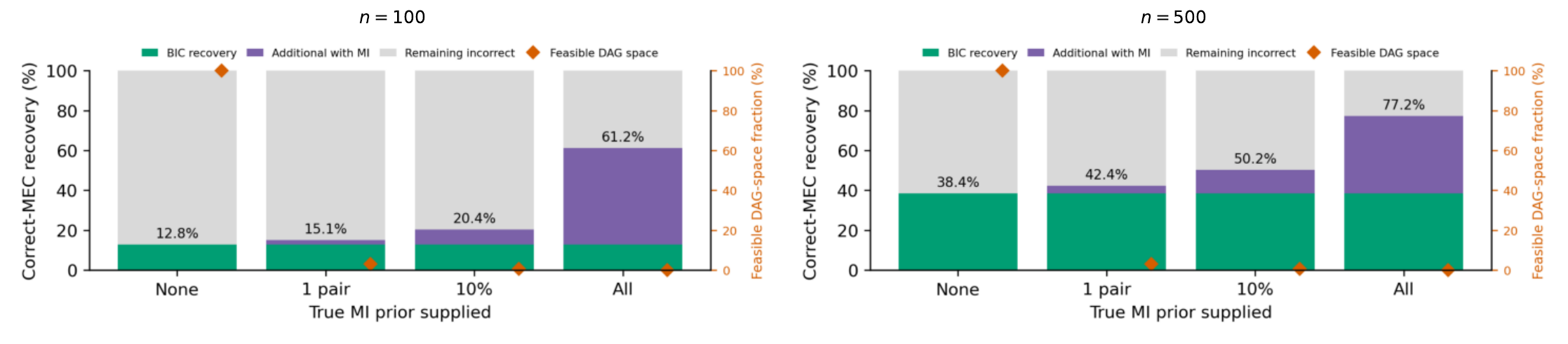}
\caption{Pooled correct-MEC recovery at $n=100$ (left) and $n=500$ (right) for the globally optimized hard-NOTREKS experiment at $d=10$. Prior budgets contain no pair, one pair, $10\%$ of the available oracle no-trek pairs, or all available pairs, using matched graph, noise, and prior-draw seeds. Green denotes recovery already achieved by ordinary BIC, purple denotes additional recovery under the supplied hard no-trek constraints, and gray denotes remaining error. Orange diamonds, read against the right axes, estimate the fraction of uniformly weighted labeled DAGs satisfying the supplied constraints, conditional on a nonempty supplied prior for nonzero budgets. Lower error indicates better recovery; the two sample sizes are shown separately.}
\label{fig:exact-optimization-experiment}
\end{figure}

For each budget, Gaussian-regression BIC is globally minimized over all DAGs subject to the supplied hard no-trek constraints. After optimization, all optimal feasible DAG traces are exhausted and collapsed to distinct CPDAGs. Correct recovery requires every minimizing CPDAG to equal the true Markov-equivalence class. In \cref{fig:exact-optimization-experiment} the paired $n=100$ and $n=500$ experiments are compared. Increasing the sample size improves ordinary-BIC recovery, while correct hard NOTREKS constraints provide additional recovery at both sample sizes. The orange search-space diagnostic estimates the fraction of uniformly weighted labeled DAGs satisfying the supplied constraints.

This deliberately small experiment isolates the finite-sample role of correct hard NOTREKS information from the optimization difficulties of the larger benchmarks.

\subsection{Synthetic Benchmarks}
\medskip\noindent\textit{Experimental protocol.} Unless stated otherwise, each synthetic graph-family/density setting contains $20$ graph replicates, and the same graph and data-generating draws are paired across sample sizes $n\in\{100,500,2000\}$. We use ER and WS graphs with target mean degree $k\in\{2,4,8\}$. The well-specified linear-Gaussian arm is evaluated with FLOP at $d\in\{20,50,100\}$ and with DAGMA at $d\in\{20,50\}$. We compare no prior information ($q=0$), five independently drawn prior subsets containing one quarter of the available ground-truth-derived NOTREKS pairs ($q=\frac14$), and the complete ground-truth-derived set ($q=1$). Thus, uncertainty in the $q=\frac14$ condition includes variation across five prior draws per graph, whereas the $q=1$ condition has one complete prior set per graph. FLOP uses $20$ initial causal-order draws and DAGMA uses $2$ optimization initializations, together with the fixed production settings described in the method sections; these settings are shared across graph and prior conditions rather than tuned separately for each result.

\subsubsection{Scores and hyperparameters}
\label{subsec:scores-hyperparameters}

\medskip\noindent\textit{FLOP score.} For the well-specified linear-Gaussian arm, FLOP uses the decomposable linear-Gaussian BIC criterion of Schwarz \citep{schwarz1978estimating}
\begin{equation*}
\begin{aligned}
\mathcal{L}_{\mathrm{BIC}}(G)
&=n\sum_{j=1}^{d}\log\widehat{\sigma}^{2}_{j\mid\operatorname{pa}_G(j)} +\lambda_{\mathrm{BIC}}|E(G)|\log n,
\qquad \lambda_{\mathrm{BIC}}=2.
\end{aligned}
\end{equation*}
Here, $\widehat{\sigma}^{2}_{j\mid\operatorname{pa}_G(j)}$ is the least-squares residual variance for node $j$ given its parents. This is the ordinary Gaussian-regression BIC used by FLOP, with $\lambda_{\mathrm{BIC}}=2$. FLOP's score is therefore well-specified for our main data-generating model.

\medskip\noindent\textit{FLOP feasibility implementation.} FLOP+NOTREKS stores exact ancestor sets as packed bitsets. A proposed edge addition is checked by bitset intersections over the supplied NOTREKS pairs; after an accepted addition, the affected descendant cone is updated by propagating the source ancestor bitset. Deletions cannot create a NOTREKS violation, but the ancestry table is conservatively rebuilt after accepted deletions because reachability may have alternative supporting paths. With $b=\lceil d/64\rceil$ machine words per ancestor set, these operations cost $O(\abs{\mathcal I}b)$ for a pairwise proposal check and $O(\abs{E}+d^2/64)$ in the worst case for an accepted-addition update, while the overall runtime also includes repeated BIC proposal evaluations.

\medskip\noindent\textit{DAGMA score.} As described in the paper, DAGMA \citep{bello2022dagma} uses an $\ell_2$ score for continuous data. In all of our DAGMA experiments, we instead use the profiled unequal-variance Gaussian criterion. This applies to both vanilla DAGMA and DAGMA+NOTREKS. The criterion follows directly by profiling one Gaussian residual variance for each node:
\begin{equation*}
\begin{aligned}
Q_{\mathrm{DAGMA}}(W)
&=\frac{1}{2}\sum_{j=1}^{d}\log q_j(W),
\\
q_j(W)
&=\widehat{\Sigma}_{jj}-2\widehat{\Sigma}_{:j}^{\mathsf T}w_j +w_j^{\mathsf T}\widehat{\Sigma}w_j,
\end{aligned}
\end{equation*}
where the latter is the profiled residual variance for node $j$, with $w_j$ the $j$th column of $W$ and $\widehat{\Sigma}$ the sample covariance matrix of the standardized data.

\medskip\noindent\textit{DAGMA production settings.} We use $\lambda_1=0.03$ and five continuation stages with
$$\mu=(1,0.1,0.01,0.001,0.0001)\text{ and }s=(1,0.9,0.8,0.7,0.6).$$ We use $30{,}000$ warm-up and $60{,}000$ terminal iterations, learning rate $0.0003$, checkpoint interval $1{,}000$, and Adam parameters $(\beta_1,\beta_2)=(0.99,0.999)$. These are the values used in the DAGMA implementation. For DAGMA+NOTREKS, we additionally use $\lambda_{\mathrm{NT}}=200$, the inverse PSTrek kernel with the fast implementation, and two optimization initializations. The PSTrek penalty uses the same stage value $s_t$ as the DAGMA log-determinant constraint.

\medskip\noindent\textit{DAGMA post-selection.} The optimizer produces a dense weighted matrix without applying a positive optimizer threshold. We first construct candidate supports using the DAG and, for\linebreak DAGMA+NOTREKS, the NOTREKS constraints. We then refit and score these supports with the ordinary Gaussian BIC above and apply deletion-only feasible parent shrinking. Thus, DAGMA post-selection selects among supports generated from its continuous optimizer rather than globally searching all feasible DAGs. The misspecified arms retain these Gaussian criteria. They therefore assess performance under non-Gaussian innovations and nonlinear mechanisms rather than score-matched estimation.

\subsubsection{Metrics}

We write $C=\operatorname{CPDAG}(G)$ and $\widehat C=\operatorname{CPDAG}(\widehat G)$, where $G$ is the ground-truth DAG and $\widehat G$ is the DAG returned by the structure-learning method.

\paragraph{CPDAG SHD.} The primary structural metric is the structural Hamming distance between $C$ and $\widehat C$. It counts false-positive adjacencies, false-negative adjacencies, and incorrect edge marks or orientations. Smaller values indicate better recovery of the conditional-independence structure and its identifiable orientations.

\paragraph{CPDAG Ancestor-AID.} We additionally report Ancestor-AID, an adjustment-identification distance based on the ancestor-adjustment strategy of \citet{henckel2024adjustment}. For every ordered treatment~--~outcome pair $(X_i,X_j)$, the strategy produces an identification conclusion from the estimated\linebreak
CPDAG $\widehat C$. The conclusion may be that the effect is identifiable by ancestor adjustment, together with an ancestor-based adjustment set, or that no such conclusion is available.

\medskip\noindent\textit{Adjustment-set validity.} The validity of the proposed adjustment set is determined using the generalized adjustment criterion (GAC) of \citet{perkovic2018complete}. For a CPDAG and an ordered pair $(X_i,X_j)$, a set $Z$ satisfies the GAC if and only if (i) the graph is adjustment-amenable relative to $(X_i,X_j)$, meaning that every proper possibly causal path from $X_i$ to $X_j$ starts with a directed edge out of $X_i$; (ii) $Z$ contains no possible descendant of $X_i$ that lies on a proper possibly causal path from $X_i$ to $X_j$; and (iii) $Z$ blocks every proper definite-status non-causal path from $X_i$ to $X_j$. Here a path is proper if it contains no vertex of the treatment set other than its starting point, is non-causal if it is not possibly causal from treatment to outcome, and has definite status if every non-endpoint vertex is unambiguously a collider or non-collider from the CPDAG edge marks. Blocking is with respect to the usual collider/non-collider path criterion. We apply the ancestor-adjustment strategy of \citet{henckel2024adjustment} to obtain the pairwise identification conclusion and candidate adjustment set from each CPDAG. The estimated and ground-truth pairwise outputs agree when both report no valid ancestor-adjustment identification, or when both report identification and the estimated candidate set satisfies the GAC in the ground-truth CPDAG $C$.

For each ordered pair, we define
\[ \delta_{ij} = \begin{cases} 0, & \text{if the estimated and ground-truth pairwise outputs agree as defined above,}\\
1, & \text{otherwise}.
\end{cases} \]
The CPDAG Ancestor-AID is
\[ \operatorname{Ancestor\text{-}AID} (C,\widehat C) = \sum_{i\ne j}\delta_{ij}. \]
Smaller values indicate fewer incorrect adjustment-identification conclusions.

\subsubsection{Size of NOTREKS Information}

\begin{figure}[ht]
\centering\includegraphics[width=0.75\linewidth]{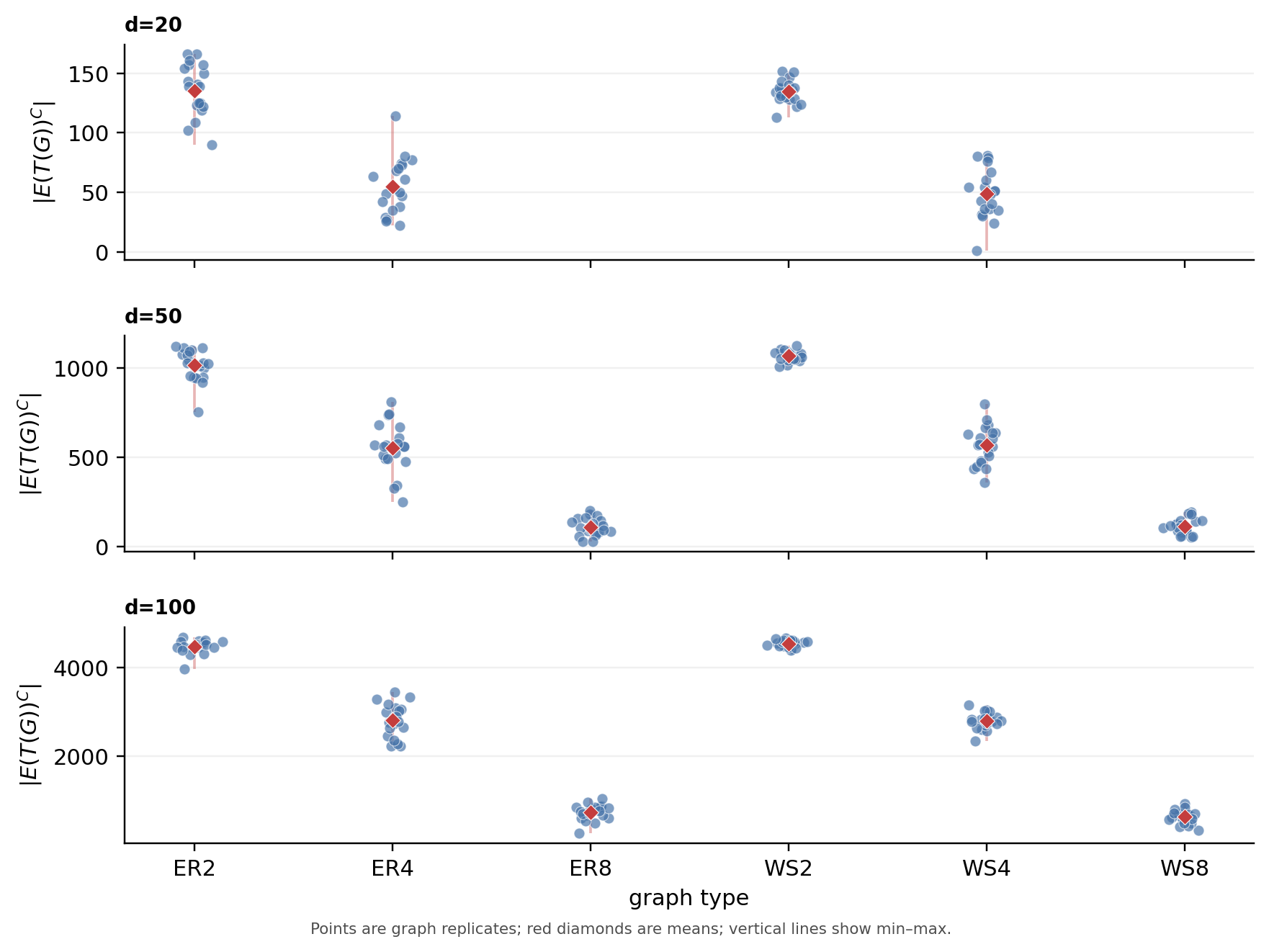}
    \caption{Exact number of available no-trek pairs as a function of graph dimension and density, stratified by graph type. These counts are the denominators used when reporting the NOTREKS information fraction $q$.}
    \label{fig:NOTREKS_information_size}
\end{figure}

\Cref{fig:NOTREKS_information_size} summarizes the exact amount of
available NOTREKS information, $\abs{E(T(G)^c)}$, across the synthetic graph settings.
For a fixed dimension, the number of no-trek pairs exhibits an
approximately linear dependence on graph density over the tested
range. This is an empirical observation for the graph families
considered here, rather than a general relationship between density
and the number of no-trek pairs.

\subsubsection{Paired Ablation Tests}
\medskip\noindent\textit{Test design.} For each method, dimension, and graph family--density setting, we use the $n=500$ graph-level results. For the $q=\frac14$ methods, we average over the five prior draws; vanilla is represented by its single $q=0$ run, while edge masking and post-selection use their matched $q=\frac14$ runs. We then compare the NOTREKS-integrated method with each baseline using two-sided paired Wilcoxon signed-rank tests on graph-level differences. Holm correction is applied separately within FLOP and DAGMA across all baseline, graph-density, and metric tests. \cref{tab:paired_ablation_tests} reports the mean difference and significance marks; a significant negative difference and black star favor the NOTREKS integration, while a significant positive difference and red star indicate deterioration.

\begin{table*}[ht]
\centering
\scriptsize
\setlength{\tabcolsep}{2.5pt}
\renewcommand{\arraystretch}{0.82}
\floatconts{tab:paired_ablation_tests}{\caption{Paired ablation tests at $n=500$ for the $q=\frac14$ NOTREKS integration. Entries are mean graph-level differences $\Delta=\mathrm{NOTREKS}-\mathrm{baseline}$; negative values favor NOTREKS. A black $^{\ast}$ indicates a significant improvement over the named baseline and a red $^{\ast}$ a significant deterioration under a two-sided Wilcoxon signed-rank test; unmarked entries are non-significant. V, E, and P denote vanilla, edge masking, and post-selection. Holm correction is applied across all baseline, graph-density, and metric tests within each method.}}{%
\begin{tabular}{llrccc|ccc}
\toprule
Graph & $k$ & Metric & \multicolumn{3}{c|}{FLOP $d=100$} & \multicolumn{3}{c}{DAGMA $d=50$}\\
 & & & V & E & P & V & E & P\\
\midrule
ER & 2 & CPDAG SHD & $-4.54^{\ast}$ & $-2.48^{\ast}$ & $-0.66$ & $-56.03^{\ast}$ & $-31.77^{\ast}$ & $-8.49^{\ast}$\\
ER & 2 & Ancestor-AID & $-29.17^{\ast}$ & $-19.42$ & $-222.79^{\ast}$ & $-171.78^{\ast}$ & $-97.01^{\ast}$ & $-5.71$\\
\addlinespace
ER & 4 & CPDAG SHD & $-3.33^{\ast}$ & $-3.27^{\ast}$ & $-0.33$ & $-157.86^{\ast}$ & $-94.28^{\ast}$ & $-13.91^{\ast}$\\
ER & 4 & Ancestor-AID & $-77.51^{\ast}$ & $-82.50^{\ast}$ & $-605.39^{\ast}$ & $-496.68^{\ast}$ & $-262.96^{\ast}$ & $-7.23$\\
\addlinespace
ER & 8 & CPDAG SHD & $0.27$ & $0.07$ & $-0.57$ & $-122.49^{\ast}$ & $-89.28^{\ast}$ & $58.49\textcolor{red}{^{\ast}}$\\
ER & 8 & Ancestor-AID & $-6.84$ & $-11.65$ & $-863.43^{\ast}$ & $-190.79^{\ast}$ & $-117.61^{\ast}$ & $130.21\textcolor{red}{^{\ast}}$\\
\addlinespace
WS & 2 & CPDAG SHD & $-6.21^{\ast}$ & $-4.70^{\ast}$ & $-2.11$ & $-61.27^{\ast}$ & $-30.26^{\ast}$ & $-14.25^{\ast}$\\
WS & 2 & Ancestor-AID & $-46.61^{\ast}$ & $-41.32^{\ast}$ & $-227.55^{\ast}$ & $-162.27^{\ast}$ & $-72.82^{\ast}$ & $-14.47^{\ast}$\\
\addlinespace
WS & 4 & CPDAG SHD & $-4.01^{\ast}$ & $-3.32^{\ast}$ & $-1.20$ & $-139.84^{\ast}$ & $-80.09^{\ast}$ & $-12.63^{\ast}$\\
WS & 4 & Ancestor-AID & $-95.02^{\ast}$ & $-66.80^{\ast}$ & $-466.40^{\ast}$ & $-505.78^{\ast}$ & $-273.13^{\ast}$ & $-5.67$\\
\addlinespace
WS & 8 & CPDAG SHD & $-0.30$ & $-0.92$ & $-2.56$ & $-120.34^{\ast}$ & $-93.05^{\ast}$ & $24.78\textcolor{red}{^{\ast}}$\\
WS & 8 & Ancestor-AID & $-11.34$ & $-52.86^{\ast}$ & $-860.42^{\ast}$ & $-215.85^{\ast}$ & $-155.94^{\ast}$ & $95.04\textcolor{red}{^{\ast}}$\\
\bottomrule
\end{tabular}}
\end{table*}

Across the tested settings, FLOP+NOTREKS does not show a significant disadvantage relative to the compared baselines. In the high-density ER8 setting, where relatively few NOTREKS pairs are available, the mean CPDAG SHD difference is slightly positive. The CPDAG SHD differences between FLOP+NOTREKS and post-selection are generally small, whereas the corresponding Ancestor-AID differences are larger in magnitude and favor FLOP+NOTREKS in the tested settings. Thus, the relative comparison depends on the recovery metric: post-selection can produce similar structural skeletons, while the two procedures differ more on ancestor-adjustment identification.

DAGMA+NOTREKS shows significantly lower recovery errors than vanilla DAGMA and edge masking in all tested settings. For high-density graphs, where relatively little NOTREKS information is available, post-selection shows significantly lower errors than the integrated NOTREKS variant.

\subsubsection{Misspecified Synthetic Results}\label{subsec:Additional Synthetic results}

The misspecified arms use the same generated DAGs, signed edge weights, graph families, densities, dimensions, sample sizes, and NOTREKS pairs as the well-specified linear-Gaussian arm. Thus, the graph and prior-information structure are held fixed while the data-generating mechanisms change. In the linear non-Gaussian arm, each variable is generated by the same linear structural equation as in the well-specified arm, but its independent innovation is Laplace rather than Gaussian. Specifically, if $\varepsilon_j\sim\operatorname{Laplace}(0,1)$ with $\operatorname{Var}(\varepsilon_j)=2$, we use $\widetilde\varepsilon_j=\varepsilon_j/\sqrt{2}$, so that $\operatorname{Var}(\widetilde\varepsilon_j)=1$, and then apply the variable-specific scale $\sigma_j$, with $\log\sigma_j^2\sim\operatorname{Unif}[-\log 2,\log 2]$. In the nonlinear Gaussian arm, the innovations remain Gaussian, while the linear parent signal entering each structural equation is replaced by the smooth bounded additive mechanism $f(s)=\tanh(s)+0.15\sin(s)$. In both arms, the resulting variables are standardized before structure learning. Paired comparisons between the $q=\frac14$ NOTREKS condition and the corresponding vanilla method are reported in \cref{tab:additional_synthetic_notreks_vanilla}.

\begin{table*}[htp]
\centering
\scriptsize
\setlength{\tabcolsep}{2.5pt}
\renewcommand{\arraystretch}{0.74}
\floatconts{tab:additional_synthetic_notreks_vanilla}{\caption{Paired comparison of the $q=\frac14$ NOTREKS condition with the vanilla method. Entries are mean graph-level differences $\Delta=\mathrm{NOTREKS}-\mathrm{vanilla}$; negative values favor NOTREKS. A black $^{\ast}$ marks a Holm-adjusted two-sided Wilcoxon signed-rank test with $p<0.05$ and a negative mean difference; a red $^{\ast}$ would mark a significant positive difference. ER and WS graphs are pooled within each density $k$. Each cell has $N=40$ graph-level pairs, except for linear non-Gaussian WS8. Dashes indicate dimension--density combinations that were not run because the resulting graphs are too dense to feature no-trek pairs.}}{%
\begin{tabular}{lllr@{\quad}rrrrr}
\toprule
Arm & $n$ & $k$ & Metric & FLOP $d=20$ & FLOP $d=50$ & FLOP $d=100$ & DAGMA $d=20$ & DAGMA $d=50$ \\
\midrule
\multicolumn{9}{l}{\textit{Well-specified linear-Gaussian}} \\
 & 100 & 2 & CPDAG SHD & $-1.16^{\ast}$ & $-6.61^{\ast}$ & $-23.64^{\ast}$ & $-10.69^{\ast}$ & $-34.86^{\ast}$ \\
 & 100 & 2 & Ancestor-AID & $-6.39^{\ast}$ & $-45.25^{\ast}$ & $-194.08^{\ast}$ & $-20.94^{\ast}$ & $-85.55^{\ast}$ \\
 & 100 & 4 & CPDAG SHD & $-1.44^{\ast}$ & $-4.38^{\ast}$ & $-14.53^{\ast}$ & $-16.42^{\ast}$ & $-79.47^{\ast}$ \\
 & 100 & 4 & Ancestor-AID & $-13.05^{\ast}$ & $-72.17^{\ast}$ & $-286.48^{\ast}$ & $-44.18^{\ast}$ & $-361.13^{\ast}$ \\
 & 100 & 8 & CPDAG SHD & -- & $-2.01$ & $-5.16^{\ast}$ & -- & $-56.85^{\ast}$ \\
 & 100 & 8 & Ancestor-AID & -- & $-47.57^{\ast}$ & $-235.55^{\ast}$ & -- & $-122.65^{\ast}$ \\
 & 500 & 2 & CPDAG SHD & $-0.19$ & $-1.21^{\ast}$ & $-5.37^{\ast}$ & $-16.78^{\ast}$ & $-58.65^{\ast}$ \\
 & 500 & 2 & Ancestor-AID & $-1.51$ & $-9.09^{\ast}$ & $-37.89^{\ast}$ & $-28.87^{\ast}$ & $-167.03^{\ast}$ \\
 & 500 & 4 & CPDAG SHD & $-0.27$ & $-0.52$ & $-3.67^{\ast}$ & $-31.57^{\ast}$ & $-148.85^{\ast}$ \\
 & 500 & 4 & Ancestor-AID & $-2.07$ & $-8.45$ & $-86.27^{\ast}$ & $-66.63^{\ast}$ & $-501.23^{\ast}$ \\
 & 500 & 8 & CPDAG SHD & -- & $-0.44$ & $-0.01$ & -- & $-121.41^{\ast}$ \\
 & 500 & 8 & Ancestor-AID & -- & $-8.97$ & $-9.09$ & -- & $-203.32^{\ast}$ \\
 & 2000 & 2 & CPDAG SHD & $0.08$ & $-0.14$ & $-1.79^{\ast}$ & $-21.91^{\ast}$ & $-80.98^{\ast}$ \\
 & 2000 & 2 & Ancestor-AID & $0.55$ & $-2.77$ & $-23.68^{\ast}$ & $-36.86^{\ast}$ & $-230.53^{\ast}$ \\
 & 2000 & 4 & CPDAG SHD & $-0.23$ & $0.25$ & $-0.26$ & $-42.09^{\ast}$ & $-202.60^{\ast}$ \\
 & 2000 & 4 & Ancestor-AID & $-2.09$ & $-0.55$ & $-9.62$ & $-64.58^{\ast}$ & $-510.86^{\ast}$ \\
 & 2000 & 8 & CPDAG SHD & -- & $0.19$ & $0.04$ & -- & $-156.03^{\ast}$ \\
 & 2000 & 8 & Ancestor-AID & -- & $4.37$ & $1.48$ & -- & $-189.68^{\ast}$ \\
\midrule
\multicolumn{9}{l}{\textit{Linear non-Gaussian}} \\
 & 100 & 2 & CPDAG SHD & $-1.25^{\ast}$ & $-7.31^{\ast}$ & $-24.48^{\ast}$ & $-12.96^{\ast}$ & $-34.30^{\ast}$ \\
 & 100 & 2 & Ancestor-AID & $-7.74^{\ast}$ & $-47.32^{\ast}$ & $-170.08^{\ast}$ & $-21.93^{\ast}$ & $-80.25^{\ast}$ \\
 & 100 & 4 & CPDAG SHD & $-1.00$ & $-3.99^{\ast}$ & $-16.86^{\ast}$ & $-17.54^{\ast}$ & $-85.88^{\ast}$ \\
 & 100 & 4 & Ancestor-AID & $-6.28$ & $-56.34^{\ast}$ & $-323.24^{\ast}$ & $-42.95^{\ast}$ & $-367.96^{\ast}$ \\
 & 100 & 8 & CPDAG SHD & -- & $-2.67$ & $-8.70^{\ast}$ & -- & $-80.55^{\ast}$ \\
 & 100 & 8 & Ancestor-AID & -- & $-58.69$ & $-258.85^{\ast}$ & -- & $-176.15^{\ast}$ \\
 & 500 & 2 & CPDAG SHD & $-0.38$ & $-1.96^{\ast}$ & $-5.76^{\ast}$ & $-19.02^{\ast}$ & $-59.51^{\ast}$ \\
 & 500 & 2 & Ancestor-AID & $-2.94$ & $-15.15^{\ast}$ & $-52.87^{\ast}$ & $-32.86^{\ast}$ & $-152.21^{\ast}$ \\
 & 500 & 4 & CPDAG SHD & $-0.27$ & $-0.79$ & $-2.83^{\ast}$ & $-27.08^{\ast}$ & $-156.94^{\ast}$ \\
 & 500 & 4 & Ancestor-AID & $-2.62$ & $-8.43$ & $-66.50^{\ast}$ & $-56.32^{\ast}$ & $-506.96^{\ast}$ \\
 & 500 & 8 & CPDAG SHD & -- & $-0.48$ & $-1.24$ & -- & $-125.40^{\ast}$ \\
 & 500 & 8 & Ancestor-AID & -- & $-7.64$ & $-46.36$ & -- & $-195.44^{\ast}$ \\
 & 2000 & 2 & CPDAG SHD & $-0.14$ & $-0.21$ & $-0.63$ & $-20.86^{\ast}$ & $-80.28^{\ast}$ \\
 & 2000 & 2 & Ancestor-AID & $-0.94$ & $-3.80$ & $-10.11$ & $-35.90^{\ast}$ & $-204.16^{\ast}$ \\
 & 2000 & 4 & CPDAG SHD & $0.14$ & $-0.14$ & $-0.79$ & $-33.55^{\ast}$ & $-185.20^{\ast}$ \\
 & 2000 & 4 & Ancestor-AID & $1.46$ & $-3.01$ & $-30.09$ & $-58.48^{\ast}$ & $-531.05^{\ast}$ \\
 & 2000 & 8 & CPDAG SHD & -- & $-0.04$ & $-0.12$ & -- & $-150.32^{\ast}$ \\
 & 2000 & 8 & Ancestor-AID & -- & $-2.35$ & $-10.41$ & -- & $-222.65^{\ast}$ \\
\midrule
\multicolumn{9}{l}{\textit{Nonlinear Gaussian}} \\
 & 100 & 2 & CPDAG SHD & $-2.49^{\ast}$ & $-8.65^{\ast}$ & $-27.33^{\ast}$ & $-4.53^{\ast}$ & $-10.98^{\ast}$ \\
 & 100 & 2 & Ancestor-AID & $-19.82^{\ast}$ & $-78.52^{\ast}$ & $-244.66^{\ast}$ & $-8.11^{\ast}$ & $-16.43^{\ast}$ \\
 & 100 & 4 & CPDAG SHD & $-1.73^{\ast}$ & $-9.84^{\ast}$ & $-26.38^{\ast}$ & $-7.25^{\ast}$ & $-15.57^{\ast}$ \\
 & 100 & 4 & Ancestor-AID & $-24.33^{\ast}$ & $-155.20^{\ast}$ & $-531.76^{\ast}$ & $-21.39^{\ast}$ & $-67.70^{\ast}$ \\
 & 100 & 8 & CPDAG SHD & -- & $-6.03^{\ast}$ & $-18.26^{\ast}$ & -- & $-17.78^{\ast}$ \\
 & 100 & 8 & Ancestor-AID & -- & $-144.26^{\ast}$ & $-596.91^{\ast}$ & -- & $-133.93^{\ast}$ \\
 & 500 & 2 & CPDAG SHD & $-0.98^{\ast}$ & $-3.59^{\ast}$ & $-8.20^{\ast}$ & $-10.75^{\ast}$ & $-26.66^{\ast}$ \\
 & 500 & 2 & Ancestor-AID & $-8.15^{\ast}$ & $-37.76^{\ast}$ & $-113.55^{\ast}$ & $-24.02^{\ast}$ & $-74.59^{\ast}$ \\
 & 500 & 4 & CPDAG SHD & $-3.62^{\ast}$ & $-5.91^{\ast}$ & $-8.43^{\ast}$ & $-15.55^{\ast}$ & $-18.30^{\ast}$ \\
 & 500 & 4 & Ancestor-AID & $-33.71^{\ast}$ & $-111.50^{\ast}$ & $-215.70^{\ast}$ & $-57.35^{\ast}$ & $-280.54^{\ast}$ \\
 & 500 & 8 & CPDAG SHD & -- & $-6.28^{\ast}$ & $-8.75^{\ast}$ & -- & $-22.47^{\ast}$ \\
 & 500 & 8 & Ancestor-AID & -- & $-159.72^{\ast}$ & $-377.69^{\ast}$ & -- & $-320.52^{\ast}$ \\
 & 2000 & 2 & CPDAG SHD & $-0.21^{\ast}$ & $-0.73^{\ast}$ & $-3.48^{\ast}$ & $-12.59^{\ast}$ & $-33.60^{\ast}$ \\
 & 2000 & 2 & Ancestor-AID & $-1.09$ & $-8.50$ & $-37.28^{\ast}$ & $-26.51^{\ast}$ & $-88.01^{\ast}$ \\
 & 2000 & 4 & CPDAG SHD & $-0.86^{\ast}$ & $-0.92^{\ast}$ & $-3.15^{\ast}$ & $-18.59^{\ast}$ & $-10.23^{\ast}$ \\
 & 2000 & 4 & Ancestor-AID & $-8.50^{\ast}$ & $-14.97^{\ast}$ & $-104.57^{\ast}$ & $-56.77^{\ast}$ & $-255.14^{\ast}$ \\
 & 2000 & 8 & CPDAG SHD & -- & $-1.16^{\ast}$ & $-1.58^{\ast}$ & -- & $-11.44^{\ast}$ \\
 & 2000 & 8 & Ancestor-AID & -- & $-32.40^{\ast}$ & $-64.95$ & -- & $-236.00^{\ast}$ \\
\bottomrule
\end{tabular}}
\end{table*}

\subsubsection{Structure of NOTREKS Information}\label{subsec:Structure of NOTREKS Information}
\medskip\noindent\textit{Construction.}
To study the role of the NOTREKS information graph structure at fixed information quantity, we construct NOTREKS information graphs with a prescribed target chromatic number \(K\). We first run a bounded randomized greedy clique search on the complete trek-graph complement \(T(G)^c\), using several random restarts, and retain the largest clique found. We select \(K\) vertices from this clique, include all pairs among them in the NOTREKS information graph, and assign these vertices distinct color labels. This planted \(K\)-clique ensures \(\chi(H)\geq K\).

The remaining vertices are then assigned randomly to one of the \(K\) color classes. We retain eligible no-trek pairs whose endpoints lie in different classes, which gives a \(K\)-coloring of the graph and hence \(\chi(H)\leq K\). After including the planted clique, additional admissible cross-class pairs are sampled until the fixed information budget is reached. Thus, the resulting NOTREKS information graph has \(\chi(H)=K\) while the number of supplied NOTREKS pairs remains constant across target values. If the $K$-partition does not provide enough additional admissible cross-class pairs, the construction fails for that graph and target $K$, and another attempt is run.

For comparison, one could select \(K\) source nodes directly: source nodes form a clique in \(T(G)^c\), and \(\chi(T(G)^c)\) equals the number of sources of the ground-truth DAG. We do not use this source-based construction here because it could reduce variation across the prior draws.

For the reported trend analysis, we retain target values represented for at least $90\%$ of the generated graphs; this limits selection from increasingly infeasible high-$K$ constructions while preserving a common graph-level support as far as the experiment permits.

\medskip\noindent\textit{Trend analysis.} To quantify the exploratory chromatic-number trend, we first average CPDAG SHD over the distinct prior sets generated for each ER4 graph and target value $K$. The fixed size of $H$ is roughly $\frac14
\E[\lvert  E(T(G)^c \rvert]$. Each resulting graph--$K$ cell contributes one observation, so graphs with more successful prior constructions do not receive additional weight. For each method, we fit the graph fixed-effects model $\overline{\mathrm{SHD}}_{g,K}=\alpha_g+\beta K+\varepsilon_{g,K}$ using its ER4 graph cohort. Standard errors are clustered by graph, with $G-1$ finite-cluster degrees of freedom, where $G$ is the number of contributing graphs. We test $H_0:\beta=0$ against $H_1:\beta\neq0$ using the two-sided clustered $t$ statistic and apply Holm correction across the two method-specific tests. To limit selection from increasingly infeasible high-$K$ constructions, we retain target values represented for at least $90\%$ of the generated graphs.

The DAGMA test at $d=50$ gives $\hat\beta=0.470$ SHD units per unit increase in $K$ (clustered SE $0.047$, two-sided 95\% CI $[0.376,0.563]$, Holm-adjusted two-sided $p=8.07\times10^{-19}$). The FLOP test at $d=100$ gives $\hat\beta=-0.099$ (clustered SE $0.017$, CI $[-0.132,-0.065]$, Holm-adjusted two-sided $p=2.69\times10^{-8}$). Both tests use 200 contributing ER4 graphs; the regression contains 1,584 graph--$K$ cells for DAGMA and 3,767 for FLOP. Thus, within this protocol, the data support a negative structural-recovery trend for FLOP, while the DAGMA trend is significantly positive. The result remains method-specific and does not establish a universal monotone effect of chromatic number.

\begin{figure}[htbp]
    \centering
    \begin{minipage}[t]{0.49\linewidth}
        \centering
        \includegraphics[width=\linewidth]{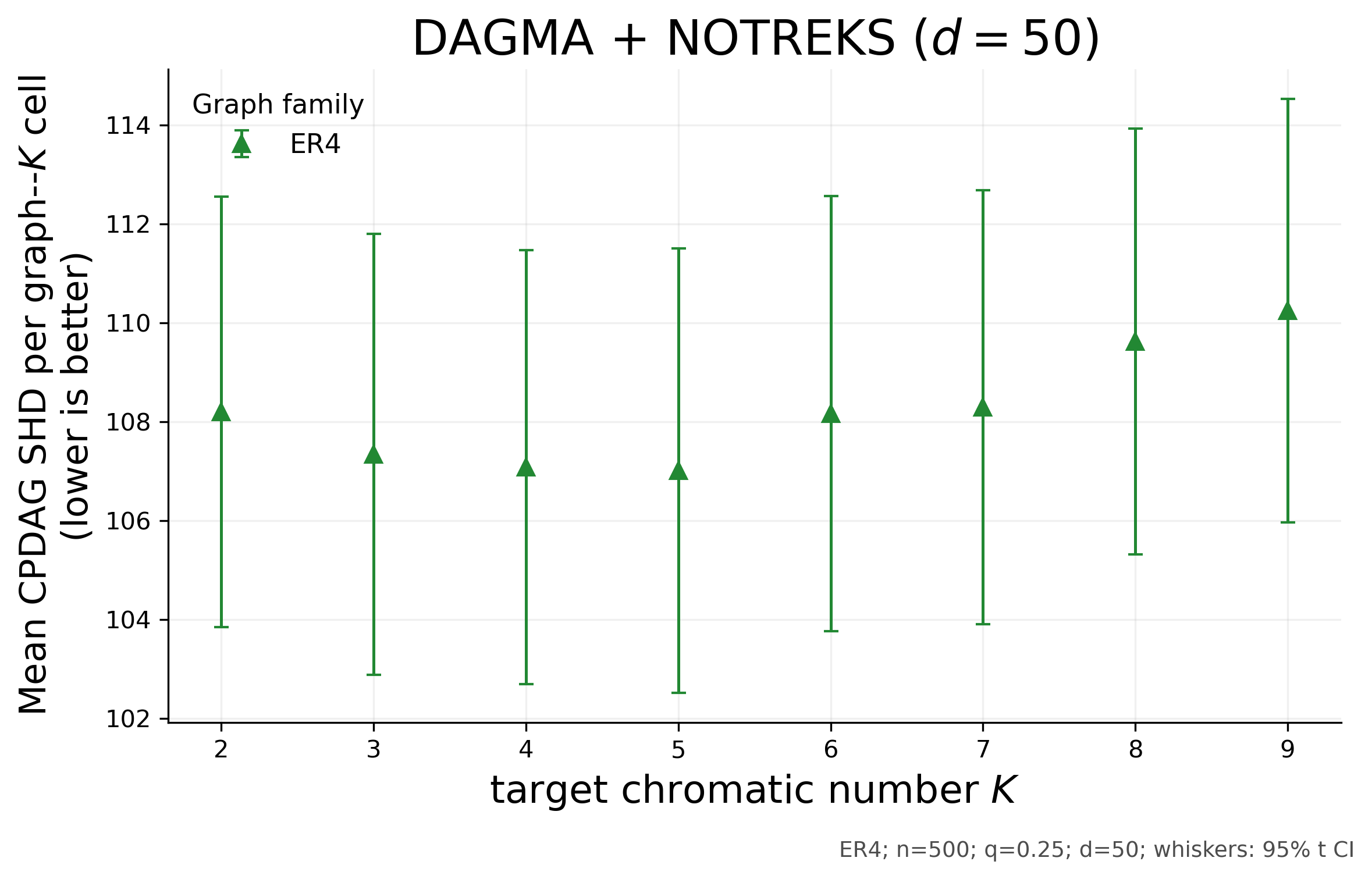}
    \end{minipage}\hfill
    \begin{minipage}[t]{0.49\linewidth}
        \centering
        \includegraphics[width=\linewidth]{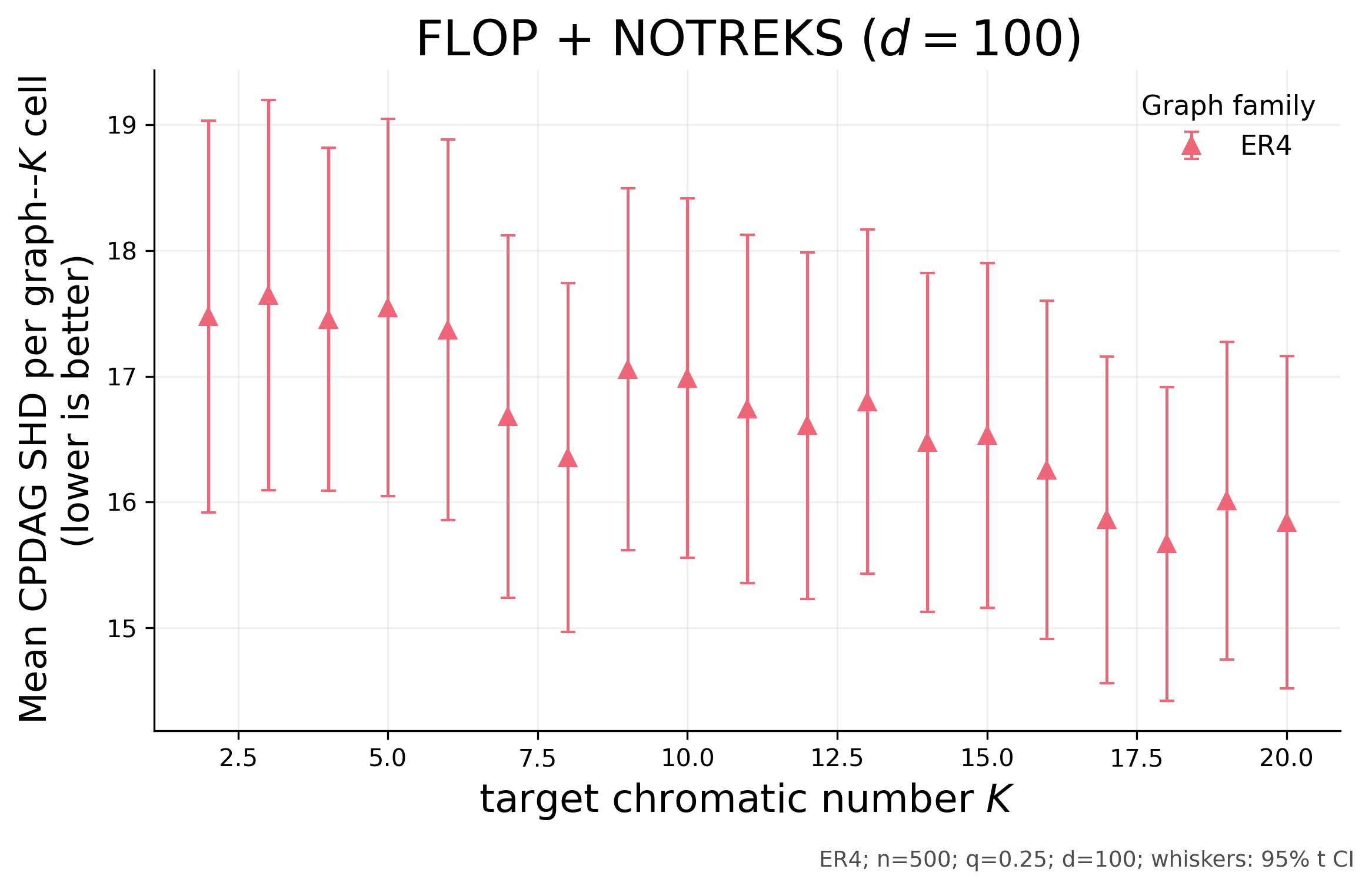}
    \end{minipage}
    \caption{Structural recovery across the target chromatic number of the supplied NOTREKS information for ER4 graphs. Points show means of graph--$K$ means after averaging over the distinct prior sets for each graph; whiskers show two-sided 95\% Student-$t$ confidence intervals across the represented ER4 graphs. Target values represented for fewer than $90\%$ of generated graphs are omitted. The points are shown without connecting trend lines because the target values are discrete and the represented graph cohort can change with $K$.}
    \label{fig:prior-structure-chromatic-trends}
\end{figure}

\subsection{In-Silico Benchmarks: Assembly Lines}\label{app:subsec:In-Silico Benchmarks: Assembly Lines}

Assembly lines provide a natural setting for NOTREKS information when separately generated components have disjoint causal histories before being combined. For example, measurements from two independently produced components may have no common ancestor, even though both subsequently influence the quality of the assembled product.

Such information can be available from knowledge of material flows, component provenance, and process design, without requiring a complete specification of the causal graph. The relevant condition is the absence of shared upstream causes. Shared raw materials, batch conditions, or environmental influences must therefore be considered when
eliciting these constraints.

\subsubsection{CausalMan}

  CausalMan is a physics-based simulator modeled after a real-world manufacturing
  line \citep{tagliapietra2025causalman}. Its structural causal model incorporates
  linear and nonlinear mechanisms, continuous and discrete variables, and
  discrete changes in operating modes. The simulator provides observational data
  together with an underlying causal structure, enabling evaluation in a
  manufacturing setting with known ground truth. We retrieve NOTREKS information
  from this ground-truth graph rather than from an expert of the assembly line.

\begin{figure}[htbp]
    \centering
    \includegraphics[width=\linewidth]{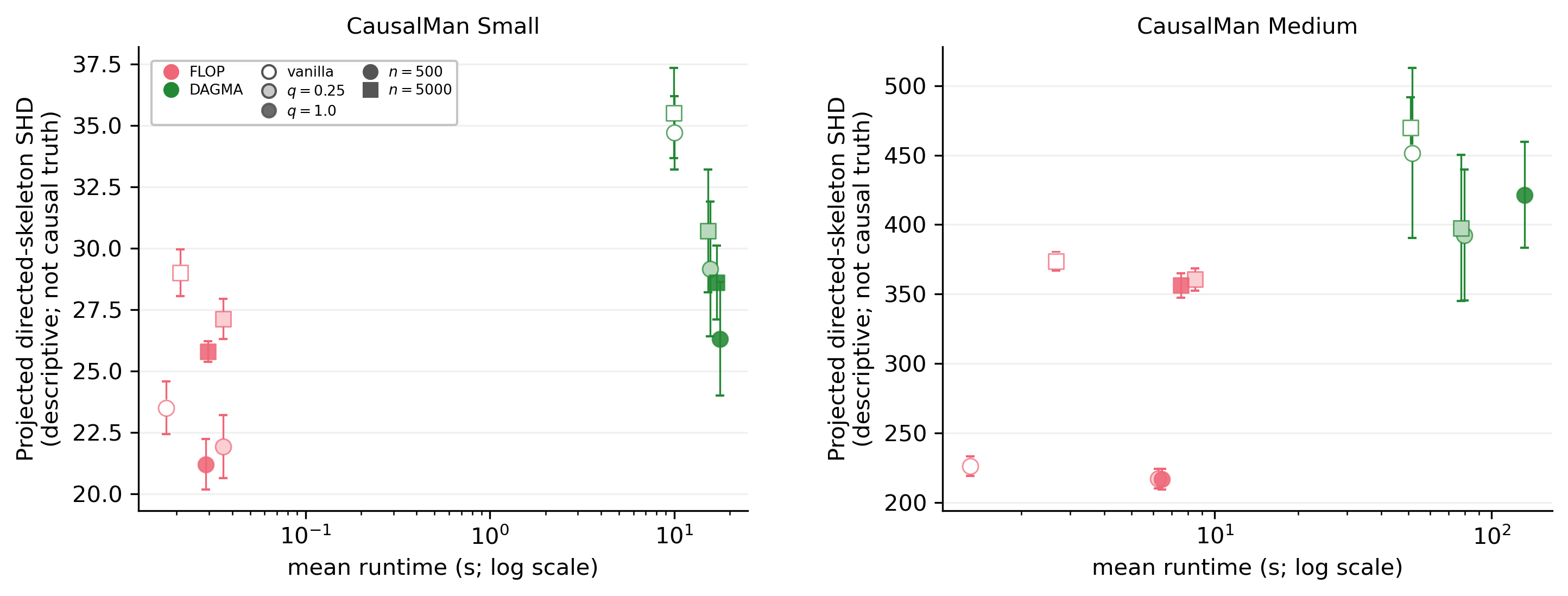}
    \caption{\emph{CausalMan} Small and Medium benchmarks. The horizontal axis reports mean runtime on a logarithmic scale. The vertical axes report projected directed-skeleton SHD. At $q \approx 0.25$, CausalMan Small uses $1/4$ or $4/14$ pairs depending on the retained graph configuration, and CausalMan Medium uses $26/104$ pairs.}
    \label{fig:causalman_small_medium}
\end{figure}

  We consider the Small and Medium CausalMan variants; see
  \cref{fig:causalman_small_medium}. The standard observed-variable CausalMan
  datasets contain latent confounding. After marginalizing latent variables, the
  observed causal structure is therefore an acyclic directed mixed graph (ADMG),
  which may contain both directed and bidirected edges.

  Because the observed CausalMan graph is an ADMG, its bidirected edges represent
  associations induced by latent common causes rather than directed causal
  relations between the retained variables. We therefore do not count bidirected
  edges as directed adjacencies, nor do we convert them into undirected edges.
  This keeps latent-confounding relations separate from directed observed-variable
  relations and makes the comparison with DAG/CPDAG-based estimators explicit.

  Instead, we use two projection-based descriptive diagnostics. Let
  $\mathcal{G}_{\mathrm{LP}}$ denote the latent projection onto the observed
  variables, and let $\mathcal{G}_{\rightarrow}$ be the directed subgraph
  obtained by retaining only its directed edges. First, projected
  directed-skeleton SHD compares the undirected skeleton of
  $\mathcal{G}_{\rightarrow}$ with the skeleton of the estimated graph. It
  counts false-positive and false-negative directed-part adjacencies, while
  ignoring edge orientation and all bidirected confounding edges. This is the
  direct structural comparison available when the latent-projection ADMG is
  compared with DAG/CPDAG-based estimators: the estimators produce
  DAG/CPDAG-like graphs, whereas the correct latent-confounded target is an
  ADMG. Smaller
  values indicate closer agreement with the directed observed-part
  target: fewer adjacency errors for projected directed-skeleton SHD.

\subsubsection{CausalAssembly}

\begin{figure}
    \centering
    \includegraphics[width=1\linewidth]{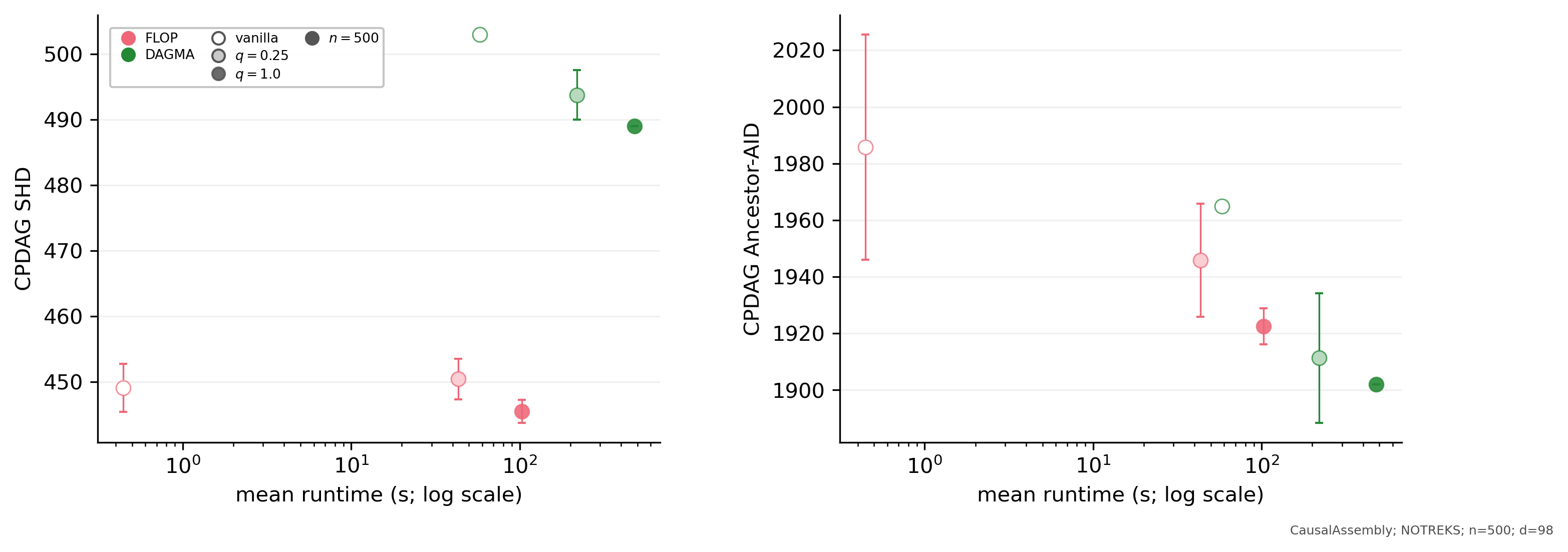}
    \caption{\emph{CausalAssembly} benchmark, at $q \approx 0.25$, CausalAssembly uses $504/2015$ pairs.}
    \label{fig:causal_assembly_oracle}
\end{figure}

The semi-synthetic benchmark \texttt{CausalAssembly} \citep{gobler2024texttt} is based on measurements collected along a real manufacturing assembly line. Physical process knowledge establishes a partial causal structure, which is completed using nonparametric regression. Distributional random forests estimate the conditional distributions associated with the resulting graph. Sampling from their product yields data that are Markov with respect to the specified ground-truth DAG. This construction combines mechanisms estimated from production data with a known graphical evaluation target. It therefore provides a complementary setting for studying NOTREKS constraints beyond fully synthetic models. We retrieve NOTREKS information from the ground-truth graph, and not from an expert of this assembly line. Results are reported in \cref{fig:causal_assembly_oracle}.

\section{Trek-Cycle Coupling Constraint}\label{subsec:Trek-Cycle Coupling constraint}

\subsection{Trek-Cycle Graph}
\begin{figure}[htbp]
    \centering
    \begin{minipage}{0.30\columnwidth}
        \centering
        \begin{tikzpicture}[->, node distance=.5cm]
            \node[circle, draw, minimum size=0.4cm] (1) at (0,1.2) {1};
            \node[circle, draw, minimum size=0.4cm] (3) at (0,0.0) {3};
            \node[circle, draw, minimum size=0.4cm] (2) at (0,-1.2) {2};

            \draw (3) -> (1);
            \draw (3) -> (2);

            \node[below] at (0,-1.8) {\small (a)};
        \end{tikzpicture}
    \end{minipage}
    \hspace{1.5em}%
    \begin{minipage}{0.6\columnwidth}
        \centering
        \begin{tikzpicture}[->, node distance=.5cm]

            \node[circle, draw, minimum size=0.4cm] (1) at (0,1.2) {$1$};
            \node[circle, draw, minimum size=0.4cm] (3) at (0,0.0) {$3$};
            \node[circle, draw, minimum size=0.4cm] (2) at (0,-1.2) {$2$};

            \node[circle, draw, minimum size=0.4cm] (1') at (2.6,1.2) {$1'$};
            \node[circle, draw, minimum size=0.4cm] (3') at (2.6,0.0) {$3'$};
            \node[circle, draw, minimum size=0.4cm] (2') at (2.6,-1.2) {$2'$};

            \draw (3) -> (1);
            \draw (3) -> (2);

            \draw (1') -> (3');
            \draw (2') -> (3');

            \draw[densely dashed] (3') -> (3);
            \draw[densely dashed] (1') -> (1);
            \draw[densely dashed] (2') -> (2);

            \draw[dotted, line width=0.6pt] (1) to[bend left=10] node[pos=0.5, above] {$w$} (2');

            \node[below] at (1.3,-1.8) {\small (b)};

        \end{tikzpicture}
        \vspace{0pt}
    \end{minipage}
    \caption{Example of a graph $G$ (a) and its trek-cycle graph $\mathcal{K}(w)$ (b) for assessing existence of a $1$--$2$ trek. Connecting edges are drawn dashed, while the added edge $1\to 2'$ is dotted. Note that $1 \to 2'$ completes a cycle together with the path $2' \to 3' \to 3 \to 1$ in $\mathcal{K}$, which corresponds to the trek $1 \leftarrow 3 \to 2$ in $G$.}
    \label{fig:example_trek_graph_K}
\end{figure}
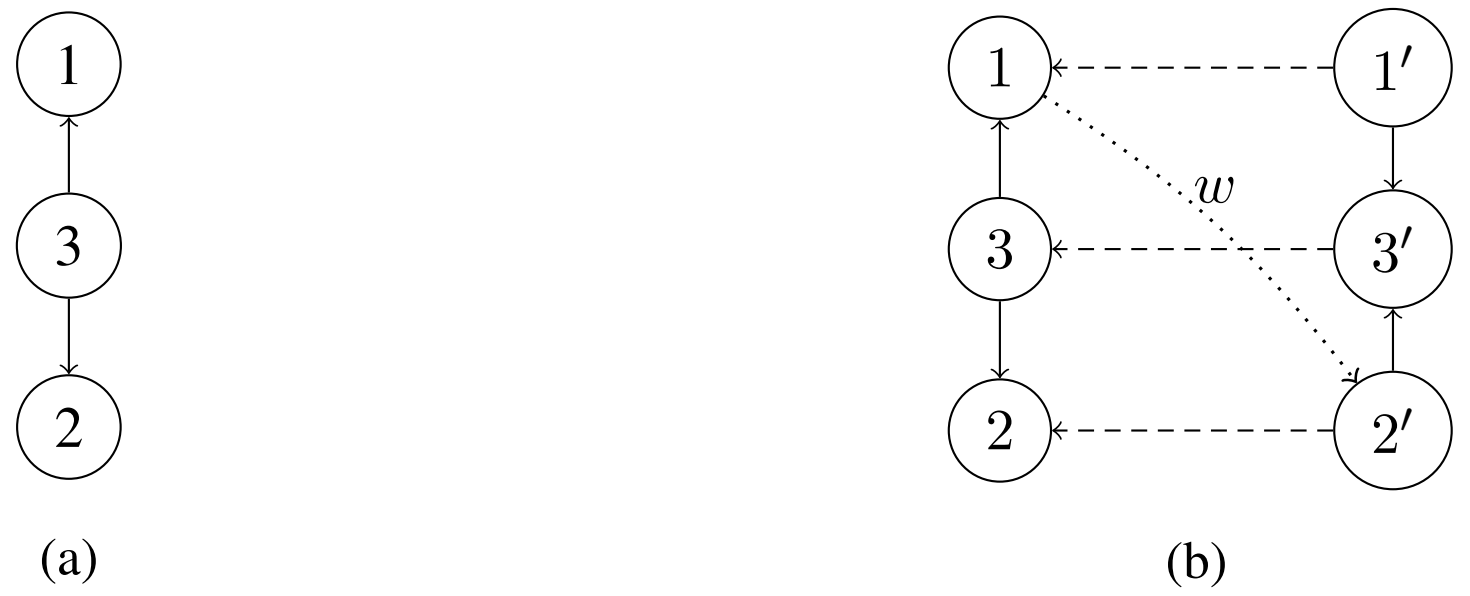

To detect $i$-$j$-treks through an acyclicity constraint, we transform trek detection into cycle detection in an auxiliary graph. The \emph{trek-cycle graph} contains the original graph $G$, a copy $G'$ with reversed edge directions, connectors $v'\to v$, and one added edge $i\to j'$. An example is shown in \cref{fig:example_trek_graph_K}; the construction is formalized below.

\begin{definition}[Trek-cycle graph $\mathcal{K}(w)$]
    Consider a graph $G$ with adjacency matrix $W \in \R^{d \times d}$ and a node pair $(i,j)$.
    The corresponding \emph{trek-cycle graph} $\mathcal{K}(w)$ for a weight $w \in \R$ is defined by the adjacency matrix
    \begin{equation}\label{eq:trek-cycle graph adjacency matrix}
        W_{\mathcal{K}(w)} = \begin{pmatrix}
            W & w E_{ij} \\ I_d & W^T
        \end{pmatrix},
    \end{equation}
    where $E_{ij}$ is the indicator matrix at the entry $(i,j)$.
\end{definition}
The construction without the upper block matrix $E_{ij}$ has been proposed by \citet[Prop.~3.5]{sullivant2010trek}, where they used it to turn treks in $G$ into directed walks in $\mathcal{K}(0)$. We add the edge which closes the directed walk to a directed cycle. We illustrate the idea in \cref{fig:example_trek_graph_K}.
The trek-cycle graph exhibits the following properties.

\begin{lemma}[Properties of the trek-cycle graph]\label[lemma]{lem:Properties of the trek-cycle graph}
The trek-cycle graph $\mathcal{K}(w)$ of a graph $G$ and a node pair $(i,j)$ for a weight $w \in \R$ exhibits the following properties.
    \begin{enumerate}
        \item For $w \neq 0$, the nodes $i$ and $j'$ are strongly connected in $\mathcal{K}(w)$ if and only if there is a $i$-$j$-trek in $G$.
        \item Specifically, $\mathcal{K}(0)$ has no directed $i$-$j'$-walk. In general, there are no directed left-to-right-walks in $\mathcal{K}(0)$.
        \item The trek-cycle graph $\mathcal{K}(0)$ is a DAG if and only if $G$ is a DAG. Moreover, for $w \neq 0$, $\mathcal{K}(w)$ is a DAG if and only if $G$ is a DAG and there is no $i$-$j$-trek in $G$.
    \end{enumerate}
\end{lemma}
\begin{proof}\label{proof:lem:Properties of the trek-cycle graph}
    \begin{enumerate}
        \item An $i$-$j$-trek in $G$ exists if and only if $\operatorname{an}_G(i) \cap \operatorname{an}_G(j)$ is non-empty. A directed $j'$-$i$-walk exists in $\mathcal{K}(0)$ if and only if
    \begin{equation}
        \{i\} \cap \{\bigcup_{v \in \operatorname{an}_G(j)} \operatorname{de}_G(v)\}
    \end{equation}
    is non-empty, which means that there is a $v$ such that $v \in \operatorname{an}_G(j)$, and $i \in \operatorname{de}_G(v)$. This is again the trek condition.

    As a consequence there is an $i$-$j$-trek in $G$ if and only if a directed $j'$-$i$-walk exists in $\mathcal{K}(0)$. The added edge constructs a directed $i$-$j'$-walk.
    \item Specifically, the edge $i \to j'$ is the only such walk, as there is no other way of changing from the original to the flipped graph.
    \item Any cycle in $G$ is copied to $\mathcal{K}(0)$. Moreover, $G$ is a DAG if and only if the flipped graph $G^T$ is a DAG. This is because a directed cycle only changes direction under the flipping of all edges. Due to the prior reasoning there are no directed left-to-right-walks in $\mathcal{K}(0)$, and hence no cycles that include left and right vertices.

    The second part follows from 1.
    \end{enumerate}
\end{proof}
We apply an acyclicity function to the trek-cycle graph. For a DAG, $g\left(W_{\mathcal{K}(w)}\right)$ detects both cycles already present in $G_W$ and the cycle created by an $i$-$j$-trek. For cyclic graphs, subtracting $g\left(W_{\mathcal{K}(0)}\right)$ removes the contribution already present at zero coupling. This motivates the following definition.

\begin{definition}[Trek-cycle constraint]\label[definition]{def:Trek-cycle Constraint}
    Let $G$ be a graph with non-negative adjacency matrix\linebreak
    $W_{\geq 0} \in \R^{d \times d}$. For a node pair $(i,j)$ and $w>0$, let $\mathcal{K}(w)$ be the corresponding trek-cycle graph with adjacency matrix $W_{\mathcal{K}(w)}$. We define the \emph{trek-cycle} constraint for an acyclicity constraint $g$ as
    \begin{equation}\label{eq:Trek-cycle Constraint}
        h_g(W) = g(W_{\mathcal{K}(w)}) - g(W_{\mathcal{K}(0)}).
    \end{equation}
\end{definition}
We seek a constraint that is non-negative and vanishes when no $i$-$j$-trek is present. One possible choice is a power-series trace function (\cref{def:Power Series Trace constraint}); here we study the spectral radius $\rho$, which has different differentiability and stability properties.

\subsection{Spectral Trek-Cycle Constraint (TCC)}
\subsubsection{Basic properties}

Let $g$ be the spectral radius $\rho$. Then \cref{def:Trek-cycle Constraint} becomes the following expression:
\begin{equation}\label{eq:Spectral Trek-cycle Constraint}
    h_\rho(W) = \rho(W_{\mathcal{K}(w)}) - \rho(W_{\mathcal{K}(0)}).
\end{equation}
The spectral radius is generally non-convex and is not differentiable when the dominant eigenvalue is not simple. Since $W_{\mathcal{K}(w)} \geq W_{\mathcal{K}(0)} \geq 0$ entrywise, monotonicity gives $h_\rho(W)\geq0$. The constraint characterizes $i$-$j$-treks in acyclic and strongly connected graphs for every $w>0$. For graphs that are neither acyclic nor strongly connected, we establish an asymptotic characterization as $w\to\infty$; for finite $w$, the added edge need not affect the dominant strongly connected component. The results are summarized in \cref{eq:Spectral Trek constraint properties}.

\subsubsection{Supporting lemmas}

A matrix is \emph{irreducible} if and only if the induced graph is strongly connected. The following Lemma shows that under positive perturbations that lead to an irreducible matrix the spectral radius increases strictly. This will be a key result to show that the spectral trek constraint is non-zero in the presence of a $i$-$j$-trek.

\begin{lemma}[Strict increase of spectral radius under positive perturbation]\label[lemma]{lem:Strict increase of spectral radius under positive perturbation leading to irreducibility}
Let $B \in \mathbb{R}_{\ge 0}^{d\times d}$ be a nonnegative matrix and let
$P \in \mathbb{R}_{\ge 0}^{d\times d}$ satisfy $P \neq 0$. Define $A := B + P$. Then $\rho(A) \geq \rho(B)$. If $A$ is irreducible, then $\rho(A) > \rho(B)$.
\end{lemma}

In the proof of this Lemma we apply the Perron-Frobenius theorem for non-negative irreducible matrices, which makes statements on the properties of the spectral radius for a non-negative, irreducible matrix. For a basic introduction we refer to \citet[Chapter~8]{meyer2000matrix}.

\begin{lemma}[Perron-Frobenius theorem; \citealp{meyer2000matrix}]\label[lemma]{th:perron-frobenius}
    Let $A \in \R_{\geq 0}^{d \times d}$ be non-negative and irreducible. Then each of the following is true.
    \begin{enumerate}
        \item $\rho(A)$ is a simple and positive eigenvalue of $A$.
        \item There exists $v>0$ such that $Av = \rho(A)v$, which is called the Perron vector. There are no non-negative eigenvectors for $A$ except for positive multiples of $v$, regardless of the eigenvalue.
        \item The Collatz-Wielandt formula states that $\rho(A) = \max_{x>0} g(x)$, where $g(x) = \min_{i \in [d]} \frac{[Ax]_i}{x_i}$.
        \item For a non-negative matrix $B$, the Collatz-Wielandt inequality applies for every non-negative vector $x$,
    \begin{equation}
        \min_{i \in [d]; x_i \neq 0} \frac{[Bx]_i}{x_i} \leq \rho(B) \leq \max_{i \in [d]; x_i \neq 0} \frac{[Bx]_i}{x_i}.
    \end{equation}
    \end{enumerate}
\end{lemma}

\begin{proof}[Proof of Lemma~\ref{lem:Strict increase of spectral radius under positive perturbation leading to irreducibility}]
The first statement is a corollary of Gelfand's formula, which states that for any matrix norm $\norm{\cdot}$, for a matrix $M \in \R^{d \times d}$ it holds that $\rho(M) = \lim_{k \to \infty} \norm{M^k}^{1/k}$.
Observe that from the entrywise inequality, it follows that $\norm{A^k}_\infty \geq \norm{B^k}_\infty$ for all $k \in \mathbb{N}$. We conclude
\begin{equation}
    \rho(B) = \lim_{k \to \infty} \norm{B^k}^{1/k} \leq \lim_{k \to \infty} \norm{A^k}^{1/k} = \rho(A).
\end{equation}

The second statement is more intricate. Fix an entry of $P$ such that $P_{ij} > 0$. Due to the irreducibility of $A$ (strongly connected), for every ordered pair of nodes there is a walk that uses the edge $(i,j)$. A simple path from the first node to $i$, followed by $(i,j)$ and a simple path from $j$ to the second node, has length at most $2d-1$. This walk has an additional positive contribution in $A^\ell$ which is absent from $B^\ell$. Let
\[
    f(x) = \sum_{\ell=1}^{2d-1} x^\ell.
\]
Consequently, the following inequality holds entrywise:
\begin{equation}\label{eq:f(B) < f(A)}
    0 \le f(B) < f(A).
\end{equation}
Since $f(A)$ is non-negative and irreducible, we conclude by the Perron-Frobenius theorem, that there exists a vector $v_{f(A)} > 0$ (Perron vector) such that $f(A)v_{f(A)} = \rho(f(A))v_{f(A)}$. Using the Collatz-Wielandt inequality for non-negative matrices, specifically for $f(B)$, it holds that
\begin{equation}
    \rho(f(B)) \leq \max_{i \in [d]} \frac{\left[f(B)x\right]_i}{x_i}
\end{equation}
for any $x > 0$. Combining this, applied to the perron vector $v_{f(A)}$, with the strict inequality from \cref{eq:f(B) < f(A)}, we get
\begin{align*}
    \rho(f(B)) &\leq \max_{i \in [d]} \frac{\left[f(B)v_{f(A)}\right]_i}{[v_{f(A)}]_i}\\
    &< \max_{i \in [d]} \frac{\left[f(A)v_{f(A)}\right]_i}{[v_{f(A)}]_i}\\
    &= \rho(f(A)).
\end{align*}
We now seek to relate $\rho(f(A))$ with $\rho(A)$. Let $v_A > 0$ be the Perron vector to $A$, i.e., $Av_A = \rho(A) v_A$.
Observe that $\sum_{\ell=1}^{2d - 1} \rho(A)^\ell$ is an eigenvalue of $f(A)$, with a strictly positive eigenvector $v_A$. This is the Perron eigenvalue of $f(A)$ as well, since $f(A)$ is irreducible and $v_A$ is strictly positive.
From the latter we conclude that
\begin{equation}\label{eq:rho(f(A))}
    \rho(f(A))=\sum_{\ell=1}^{2d - 1} \rho(A)^\ell.
\end{equation}

Let $\lambda$ be an eigenvalue of $B$, then $\sum_{\ell=1}^{2d - 1} \lambda^\ell$ is an eigenvalue of $f(B)$. An application of \cref{eq:max SCC spectra} yields $\rho(B) = \max_{i=1,\dots,k} \rho\!\left(B_{C_i}\right)$, for the SCCs of $B$.
Then Perron-Frobenius can be applied to a dominating SCC $C_\rho$ (associated to $\rho(B)$). We get that $\rho(B)$ is a positive, real eigenvalue of $C_\rho$ -- and of $B$ as well. Hence,
\begin{equation}\label{eq:rho(f(B))}
    \rho(f(B)) \geq \sum_{\ell=1}^{2d - 1} \rho(B)^\ell.
\end{equation}
Since $\rho(f(A)) > \rho(f(B))$, and $\sum_{\ell=1}^{2d - 1} r^\ell$ is strictly increasing in $r$ for $r \geq 0$, we conclude $\rho(A) > \rho(B)$.
\end{proof}

We further provide a primitive lower bound on the spectral radius of a matrix having a cycle.
\begin{lemma}\label[lemma]{lem:lower bound on spectral radius via cycle}
    Let $G$ be a graph with non-negative adjacency matrix $A$, and $\pi = (v_{i_1}, v_{i_2}, \ldots, v_{i_1})$ is a directed cycle in $G$. Then, $\rho(A) \geq \left(\prod_{(v_i, v_j)\in\pi} A_{ij}\right)^{1/\abs{\pi}}$.
\end{lemma}
\begin{proof}[Proof of Lemma~\ref{lem:lower bound on spectral radius via cycle}]
    Let $\ell := \abs{\pi}$ denote the length of $\pi$. By the spectral-radius bound for non-negative matrices, $(A^\ell)_{i_1 i_1} \leq \rho(A)^\ell$. Since the cycle $\pi$ contributes to $(A^\ell)_{i_1 i_1}$ and $A \geq 0$, we obtain
    \[
        \rho(A)^\ell \geq (A^\ell)_{i_1 i_1} \geq \prod_{(v_i, v_j)\in\pi} A_{ij}.
    \]
    Taking the $\ell$-th root proves the result.
\end{proof}

In the following, let $\mathcal{C}(A) = \{C_1, \ldots, C_k\}$ denote the strongly connected components (SCCs) of the directed graph induced by a square matrix $A$. Then, there exists a permutation matrix $P$ such that
\begin{equation}
    P^T A P \;=\;
        \begin{pmatrix}
            A_{C_1} & * & \cdots & *\\
            0 & A_{C_2} & \cdots & *\\
            \vdots & \ddots & \ddots & \vdots\\
            0 & \cdots & 0 & A_{C_k}
        \end{pmatrix}.
\end{equation}
In particular, the spectrum of $W$ is the union of the spectra of the diagonal blocks, and hence
\begin{equation}\label{eq:max SCC spectra}
    \rho(W) \;=\; \max_{i=1,\dots,k} \rho\!\left(W_{C_i}\right).
\end{equation}
Using this and the fact that the eigenvalues of $W$ and $W^T$ are shared, it follows directly that $\rho(W_{\mathcal{K}(0)}) = \rho(W)$.

The following theorem is the counterpart of \cref{th:PSTrek characterization of treks} for the TCC constraint.
\begin{theorem}[Spectral Trek constraint properties]\label[theorem]{eq:Spectral Trek constraint properties}
    Let $G$ be a graph with adjacency matrix $W$. Let $\mathcal{K}(w)$ be the trek-cycle graph of $G$ for the node pair $(i,j)$ and a weight $w \geq 0$. Then each of the following is true.
    \begin{enumerate}
        \item $h_\rho(W) = \rho(W_{\mathcal{K}(w)}) - \rho(W_{\mathcal{K}(0)}) \geq 0$.
        \item $h_\rho(W) = 0$ for all $w > 0$ if there is no $i$-$j$-trek in $G$.
        \item We have $h_\rho(W) > 0$ for all $w > 0$ if and only if \begin{enumerate}
            \item[(i)] there exists an $i$-$j$-trek in $G$ and
            \item[(ii)] the SCC $C$ including $i$ and $j'$ in $\mathcal{K}(w)$ also includes \textit{one} dominating SCC $C^\star$ from $\mathcal{K}(0)$, i.e., $C^\star \subseteq C$ and $\rho(W_{\mathcal{K}(0)\mid C^\star}) = \rho(W_{\mathcal{K}(0)}) = \rho(W)$.
        \end{enumerate}
        \item If 3.(i) holds but 3.(ii) does not, then $h_\rho(W) > 0$ for all sufficiently large $w$. More precisely, let $\pi$ be a directed cycle in the SCC containing $i$ and $j'$ which uses the added edge $i \to j'$, let $\ell := \abs{\pi}$ be its length, and let $c_\pi(W)$ be the product of the weights on the remaining edges of $\pi$. Then
        \[
            w > \frac{\rho(W)^\ell}{c_\pi(W)}
        \]
        is sufficient. Since $c_\pi(W) \geq \min \{1, W_{\min}, W_{\min}^{d-1} \}$, the right-hand side can be bounded using the minimal non-zero entry $W_{\min}$ of $W$.
        \item $h_\rho(W)$ is stable in the sense of \cite{nazaret2023stable}.
    \end{enumerate}
\end{theorem}
\begin{proof}[Proof of Theorem~\ref{eq:Spectral Trek constraint properties}]
    \begin{enumerate}
        \item For $w \geq 0$, we have $W_{\mathcal{K}(w)} \geq W_{\mathcal{K}(0)} \geq 0$. An application of \cref{lem:Strict increase of spectral radius under positive perturbation leading to irreducibility} yields the statement.
        \item If there is no $i$-$j$-trek in $G$ then there is no $j'$-$i$-walk in $\mathcal{K}(w)$, following from \cref{lem:Properties of the trek-cycle graph}. Specifically, we claim that the set of SCCs in $\mathcal{K}(0)$ and $\mathcal{K}(w)$ is equal, i.e., $\mathcal{C}(W_{\mathcal{K}(0)}) = \mathcal{C}(W_{\mathcal{K}(w)})$. If that is true, \cref{eq:max SCC spectra} gives the result. There cannot be more SCCs when an edge is added, as it increases connectivity. Assume there are less SCCs in $\mathcal{K}(w)$ than in $\mathcal{K}(0)$. Then there has been a partial order on the SCCs with $j'$ in the source SCC and $i$ in the sink SCC. But this contradicts the fact that there is no directed $j'$-$i$-walk in $\mathcal{K}(0)$.
        \item By assumption (i) and 1. in \cref{lem:Properties of the trek-cycle graph}, $i$ and $j'$ are strongly connected in $\mathcal{K}(w)$, and not strongly connected in $\mathcal{K}(0)$. As a consequence, we necessarily have less SCCs in $\mathcal{K}(w)$ than in $\mathcal{K}(0)$, i.e., $\abs{\mathcal{C}(W_{\mathcal{K}(w)})} < \abs{\mathcal{C}(W_{\mathcal{K}(0)})}$. Let $C \in \mathcal{C}(W_{\mathcal{K}(w)})$ with $i, j' \in C$.

        Assume first that condition (ii) holds. Define $A:= W_{\mathcal{K}(w) \mid C}$ as the submatrix of $W_{\mathcal{K}(w)}$ restricted to the vertices in $C$, and respectively define $B := W_{\mathcal{K}(0) \mid C}$. For notational ease let $i$ correspond to the first index and $j'$ as the second. We can now apply \cref{lem:Strict increase of spectral radius under positive perturbation leading to irreducibility} with $A$ and $B$ as defined and $P = w E_{ij}$. Note that $A$ is irreducible as it is strongly connected. We get
        \begin{align*}
            \rho(A) &> \rho(B)\\
            &= \max_{C_i \in \mathcal{C}(W_{\mathcal{K}(0)}),\, C_i \subseteq C} \rho(W_{\mathcal{K}(0) \mid C_i})\\
            &= \rho(W_{\mathcal{K}(0)})\\
            &= \rho(W),
        \end{align*}
        where the second equality follows from condition (ii), i.e., $C^\star \subseteq C$ with $\rho(W_{\mathcal{K}(0)\mid C^\star}) = \rho(W_{\mathcal{K}(0)})$.
        Hence $h_\rho(W)>0$ for all $w>0$.

        Conversely, if condition (ii) does not hold, then the SCC $C$ does not contain an SCC of $\mathcal{K}(0)$ attaining $\rho(W_{\mathcal{K}(0)})$. Hence, its spectral radius is strictly smaller than $\rho(W_{\mathcal{K}(0)})$ at $w=0$. By continuity of the spectral radius in the matrix entries, this remains true for all sufficiently small $w>0$, and therefore $h_\rho(W)=0$ for all sufficiently small $w$. This proves the converse implication in 3.
        \item An application of \cref{lem:lower bound on spectral radius via cycle} leads to the result. Observe that $i$ and $j'$ are strongly connected due to (i) and the result in \cref{lem:Properties of the trek-cycle graph}. Choose a directed cycle $\pi$ in this SCC which contains the added edge $i \to j'$, let $\ell := \abs{\pi}$ be its length, and let $c_\pi(W)$ be the product of the weights of the remaining edges of $\pi$. The cycle lemma gives
        \[
            \rho(W_{\mathcal{K}(w)})^\ell \geq w c_\pi(W).
        \]
        Consequently, $w > \rho(W)^\ell/c_\pi(W)$ implies that the spectral radius of this SCC is larger than $\rho(W)$. The eigenvalues of all other SCCs do not change, and hence $h_\rho(W)>0$. The $j'$-$i$-part of the chosen cycle has length at most $2d-3$. However, a trek can always be chosen with arms that are disjoint apart from the top, which gives at most $d-1$ weighted edges and the connector edge with weight $1$. Therefore
        \[
            c_\pi(W) \geq \min \{1, W_{\min}, W_{\min}^{d-1} \}.
        \]
        \item \begin{itemize}
        \item[(i)] We show that $h_\rho(sW) = O_{s \to \infty}(s)$, and hence it does not explode under scaling to infinity.

        It is known that $\rho(sW) = s\rho(W)$. We denote $W_{\mathcal{K}_s(w)} = \begin{pmatrix}
            sW &wE_{ij}\\
            I_n & sW^T
        \end{pmatrix}$. Then, for $s \geq 1$, we have $W_{\mathcal{K}_s(w)} \leq sW_{\mathcal{K}(w)}$.
        Using monotonicity of the spectral radius for nonnegative matrices, we obtain $\rho(W_{\mathcal{K}_s(w)})\leq s\rho(W_{\mathcal{K}(w)})$. For $w=0$, block triangularity gives $\rho(W_{\mathcal{K}_s(0)})=s\rho(W_{\mathcal{K}(0)})$. Therefore,
        \begin{align*}
            h_\rho(sW) &\leq s\left(\rho(W_{\mathcal{K}(w)}) - \rho(W_{\mathcal{K}(0)})\right) \in O_{s \to \infty}(s).
        \end{align*}
        \item[(ii)] We show that $h_\rho(\epsilon W) = \Omega_{\epsilon \to 0^+}(\epsilon)$, and hence it does not vanish under scaling to zero.

        Again by monotonicity of the spectral radius for nonnegative matrices, and for $\eps \leq 1$, we have $\eps W_{\mathcal{K}(w)} \leq W_{\mathcal{K}_\eps(w)}$. For the zero-edge construction, block triangularity gives $\rho(W_{\mathcal{K}_\eps(0)})=\eps\rho(W_{\mathcal{K}(0)})$. Hence,
        \begin{align*}
            h_\rho(\eps W) &\geq \eps\left(\rho(W_{\mathcal{K}(w)}) - \rho(W_{\mathcal{K}(0)})\right) \in \Omega_{\eps \to 0^+}(\eps).
        \end{align*}
    \end{itemize}
    \end{enumerate}
\end{proof}

\subsection{On the Differentiability of the Spectral Radius}
The spectral radius is differentiable at any matrix whose dominant eigenvalue is simple.
In particular, \citet{magnus1985differentiating} shows that all eigenvalues are differentiable functions of the matrix entries whenever the spectrum is simple. If a nonnegative matrix is irreducible, the Perron--Frobenius theorem guarantees that the largest eigenvalue is simple, and its gradient is given by
\begin{equation}\label{eq:spectral radius derivative}
    \frac{\partial \lambda}{\partial A_{ij}}
= \frac{v_i u_j}{u^T v},
\end{equation}
where $u$ and $v$ are the associated left and right eigenvectors.

Regarding the spectral trek constraint \cref{eq:Spectral Trek-cycle Constraint}, $W_{\mathcal{K}(0)}$ has repeated eigenvalues by construction, $W$ is not irreducible in general, and $W_{\mathcal{K}(w)}$ need not have a simple dominant eigenvalue. Two standard regimes are relevant:
\begin{enumerate}
    \item One approach is to argue that $W$ is almost always irreducible \citep{nazaret2023stable}. Indeed, if there is enough degree of freedom in the entries of $W$, this is true.
    \item In structure learning one might impose hard inductive bias by restricting the support of $W$ or fixing certain entries, so that $W$ may be reducible, or eigenvalues may have higher multiplicity. In many settings, $W$ still has almost-always a simple largest-in-magnitude eigenvalue. Using \cref{eq:spectral radius derivative}, the spectral radius is differentiable in every SCC. Moreover, the condition that two SCCs have the same spectral radius, i.e., $\rho\!\left(A_{C_i}\right) = \rho\!\left(A_{C_j}\right)$, has codimension $1$. However, to avoid any issues, \citet{lee2019scaling} enforce irreducibility of $W$ by adding a small $\eps > 0$ on all entries.
\end{enumerate}
\subsection{TCC for a Set of Marginal-Independence Pairs}

A natural question is whether all constraints in $\cal I$ can be enforced \emph{simultaneously} in the spectral TCC construction by replacing $E_{ij}$ with $E_{\cal I}=\sum_{(i,j)\in\cal I}E_{ij}$.
However, this destroys the trek characterization: multiple added connectors can interact and create \emph{spurious cycles}, so that the spectral penalty becomes positive even when no $\cal I$-trek exists in $G_W$.

\begin{figure}[htbp]
    \centering
    \begin{minipage}{0.3\textwidth}
        \centering
        \begin{tikzpicture}[->, node distance=.5cm]
            \node[circle, draw, minimum size=0.4cm] (1) at (0,1.2) {1};
            \node[circle, draw, minimum size=0.4cm] (3) at (0,0.0) {3};
            \node[circle, draw, minimum size=0.4cm] (2) at (0,-1.2) {2};

            \draw (3) -> (1);

            \node[below] at (0.,-2.) {\small (a)}; 
        \end{tikzpicture}
    \end{minipage}
    \hspace{1.5em}%
    \begin{minipage}{0.3\textwidth}
        \centering
        \begin{tikzpicture}[->, node distance=.5cm,
            cycle/.style={red, line width=0.9pt},
            added/.style={dotted, line width=0.9pt},
            conn/.style={densely dashed}
        ]

            \node[circle, draw, minimum size=0.4cm] (1) at (0,1.2) {$1$};
            \node[circle, draw, minimum size=0.4cm] (3) at (0,0.0) {$3$};
            \node[circle, draw, minimum size=0.4cm] (2) at (0,-1.2) {$2$};

            \node[circle, draw, minimum size=0.4cm] (1') at (2.6,1.2) {$1'$};
            \node[circle, draw, minimum size=0.4cm] (3') at (2.6,0.0) {$3'$};
            \node[circle, draw, minimum size=0.4cm] (2') at (2.6,-1.2) {$2'$};

            \draw[cycle] (3) -> (1);

            \draw (1') -> (3');

            \draw[conn] (1') -> (1);
            \draw[conn, cycle] (3') -> (3);
            \draw[cycle, conn] (2') -> (2);

            \draw[cycle, added, dotted] (1) to[bend left=10] node[pos=0.5, above] {$w$} (2');
            \draw[cycle, added, dotted] (2) to[bend right=10] node[pos=0.5, above] {$w$} (3');

            \node[below] at (1.3,-2.) {\small (b)}; 
        \end{tikzpicture}
        \vspace{0.3em}
    \end{minipage}
    \hspace{1.5em}%
    \begin{minipage}{0.3\textwidth}
        \centering
        \begin{tikzpicture}[->, node distance=.5cm,
            cycle/.style={black},
            added/.style={dotted, line width=0.9pt},
            conn/.style={densely dashed}
        ]

            \node[circle, draw, minimum size=0.4cm] (1) at (0,1.2) {$1$};
            \node[circle, draw, minimum size=0.4cm] (3) at (0,0.0) {$3$};
            \node[circle, draw, minimum size=0.4cm] (2) at (0,-1.2) {$2$};

            \node[circle, draw, minimum size=0.4cm] (1') at (2.6,1.2) {$1'$};
            \node[circle, draw, minimum size=0.4cm] (3') at (2.6,0.0) {$3'$};
            \node[circle, draw, minimum size=0.4cm] (2') at (2.6,-1.2) {$2'$};

            \draw[cycle] (3) -> (1);

            \draw (1') -> (3');

            \draw[conn] (1') -> (1);
            \draw[conn, cycle] (3') -> (3);
            \draw[cycle, conn] (2') -> (2);

            \draw[cycle, added, dotted] (1) to[bend left=10] node[pos=0.5, above] {$w$} (2');
            \draw[cycle, added, dotted] (3) to[bend right=10] node[pos=0.5, above] {$w$} (2');

            \node[below] at (1.3,-2.) {\small (c)}; 
        \end{tikzpicture}
        \vspace{0.3em}
    \end{minipage}
    \caption{Example: Spurious Cycles in Trek-cycle graph for multiple added edges}
    \label{fig:I_trek_cycle_counterexample}
\end{figure}
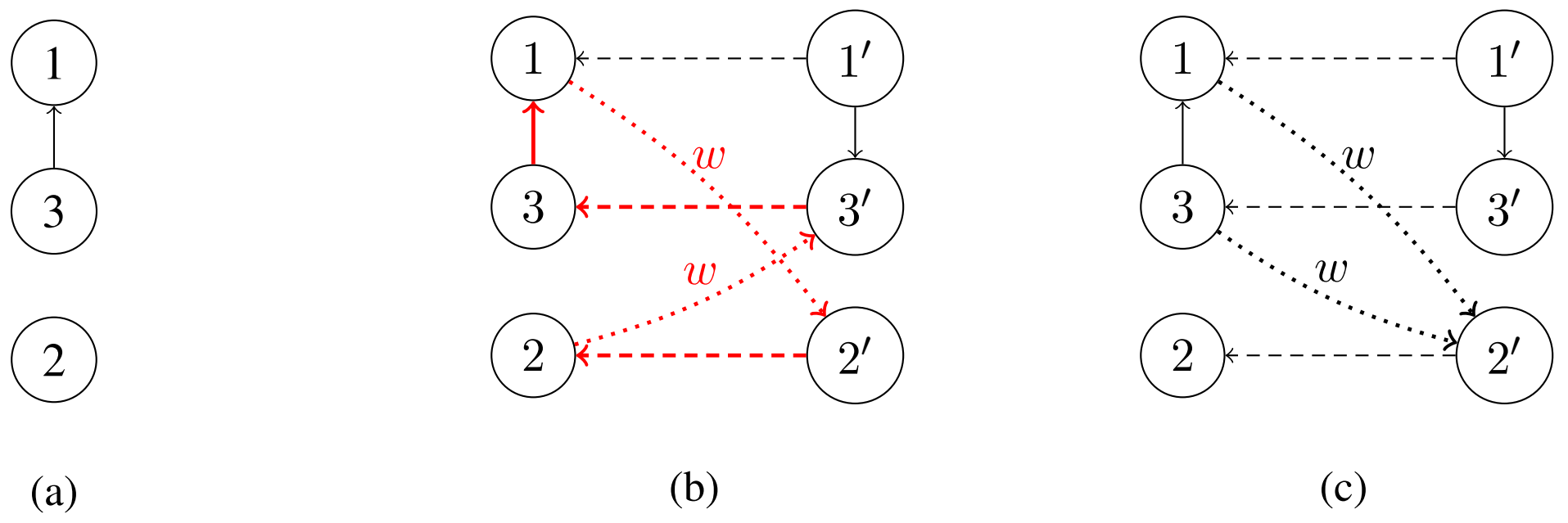

The example in \cref{fig:I_trek_cycle_counterexample} illustrates that the spectral trek construction for a set $\cal{I}$ may introduce spurious cycles -- and hence fails to characterize treks. The base graph (a) contains no trek within the node pairs in $\cal{I} = \{(1,2), (2,3)\}$. However, in the associated trek-cycle graph (b), the highlighted red cycle
$1 \to 2' \to 2 \to 3' \to 3 \to 1$
appears due to the interaction of the added edge $wE_{\mathcal I}$ (dotted) with the identity connectors (dashed) indicated by $I_3$ in the trek-cycle matrix. Consequently, $h_\rho(W_{\mathcal{K}(w;\mathcal I)}) > 0$ although no $\mathcal I$-trek exists in $G_W$. In (c) we switched the connection $2 \to 3'$ to $3 \to 2'$. Observe that (c) is acyclic, hence we get an ambiguity depending on the ordering how one encodes an independence pair of $\cal{I}$ into the trek-cycle graph.

An exception occurs, when no spurious cycles can arise, in the case of $\cal I \subseteq \{i\}\times[d]$ for a fixed node $i$, where adding edges $i\to j'$ does not introduce such artefacts (cf.\ \cref{proof:extension:Properties of the trek-cycle graph for multiple edges}).

It is possible, when avoiding a fully joint construction and instead use stochastic approximations.
At each iteration one can either (i) sample a single pair $(i,j)\in\cal I$ uniformly and evaluate its TCC penalty, or (ii) sample a node $v$ from the multiset of nodes appearing in $\cal I$ and evaluate the penalty for the star-shaped subset of all constraints involving $v$.
Both strategies avoid spurious cycles while keeping per-iteration cost independent of $|\cal I|$. For our experiments we have used the second variant.

\subsection{Extension of Lemma~\ref{lem:Properties of the trek-cycle graph} to Multiple Added Edges}\label{proof:extension:Properties of the trek-cycle graph for multiple edges}
\begin{lemma}[Properties of the trek-cycle graph to multiple connectors]\label[lemma]{lem:Properties of the trek-cycle graph to multiple connectors}
Let $G$ be a graph and let $\cal I \subseteq \{i\}\times[d]$ be a set of marginal-independence pairs for a fixed node $i$. We denote by $\mathcal{K}(w; \cal{I})$ the trek-cycle graph with upper right matrix $E_{\cal I}=\sum_{(i,j)\in\cal I}E_{ij}$ (star-shape), and otherwise analogue to the construction of the trek-cycle graph $\mathcal{K}(w)$, i.e.,
\begin{equation}\label{eq:extended trek-cycle graph adjacency matrix}
    W_{\mathcal{K}(w; \cal{I})} = \begin{pmatrix}
        W & w E_{\cal I} \\ I_d & W^T
    \end{pmatrix}.
\end{equation}
Then the following properties hold in this extended trek-cycle graph:
    \begin{enumerate}
        \item For $w \neq 0$, nodes $i$ and $j'$ with $(i,j) \in \cal I$ are strongly connected in $\mathcal{K}(w; \cal I)$ if and only if there is an $i$-$j$-trek in $G$.
        \item Specifically, $\mathcal{K}(0; \cal I)$ has no directed $i$-$j'$-walk. In general, there are no directed left-to-right walks in $\mathcal{K}(0; \cal I)$.
        \item The trek-cycle graph $\mathcal{K}(0; \cal I)$ is a DAG if and only if $G$ is a DAG. Moreover, for $w \neq 0$, $\mathcal{K}(w; \cal I)$ is a DAG if and only if $G$ is a DAG and there is no $i$-$j$-trek in $G$.
    \end{enumerate}
\end{lemma}
\begin{proof}
    Statements (ii) and (iii) follow from (i) as in the proof of \cref{lem:Properties of the trek-cycle graph}. For (i), the forward implication follows by applying the single-connector result to any $(i,j)\in\cal I$. Conversely, if $i$ and $j'$ are strongly connected for $w\neq0$, take the segment of a simple cycle from $j'$ to the first visit to $i$. It cannot use an added edge, since all added edges originate at $i$; hence it lies in $\mathcal{K}(0;\cal I)$ and corresponds to an $i$-$j$-trek in $G$.
\end{proof}

\subsection{Connecting the PSTrek and Trek-Cycle Constraints}
\label{subsec:Connecting the different Trek regularizer}

To better understand the relationship between the spectral trek-cycle constraint (TCC) and the power-series trek regularizer (PSTrek), we conducted controlled experiments on randomly generated weighted graphs.

\begin{figure}
  \centering
  \begin{minipage}{.48\textwidth}
    \centering
    \includegraphics[
      height=5cm,
      keepaspectratio
    ]{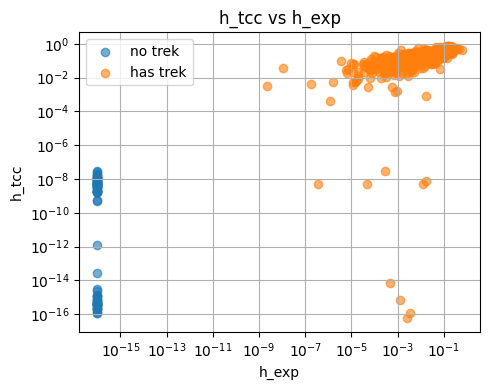}
  \end{minipage}%
  \begin{minipage}{.48\textwidth}
    \centering
    \includegraphics[
      height=5cm,
    ]{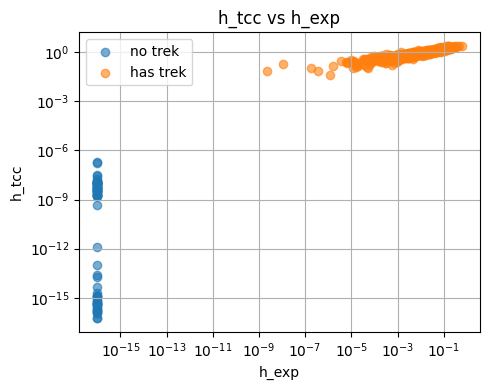}
  \end{minipage}
  \caption{
    Comparison of the spectral trek-cycle constraint (TCC) and the exponential PSTrek regularizer on randomly generated systems.
    Each point corresponds to one sampled weighted graph. The color indicates whether or not there is a $i$-$j$-trek present. We report Spearman's rank correlation $\rho$ for both regimes.
    (a) For moderate coupling weight $w=1$, the two constructions already exhibit strong monotone agreement.
    (b) For large weight $w=100$, the correlation increases further, indicating that the spectral and exponential trek scores become nearly proportional in the high-weight regime. Note that the blue dots are zero for $h_\mathrm{exp}$, but marginally above zero for $h_\mathrm{TCC}$. This might stem from numerical instability of the computation of the spectral radius.
    }
    \label{fig:tcc_vs_exp_combined}
  \end{figure}

\paragraph{Experimental setup.}
For dimensions $d \in \{5,\dots,20\}$, we generated random weighted adjacency matrices $G$ (sparse and Frobenius-normalized) inducing graphs (not necessarily acyclic) with no self-loops and no direct edge between nodes $i$ and $j$. For a fixed edge insertion weight $w$, we evaluated both regularizers and compared their values across random systems. \Cref{fig:tcc_vs_exp_combined} report scatter plots of $h_{\mathrm{TCC}}$ versus $h_{\mathrm{exp}}$.

Across the sampled regimes, the two quantities show a positive rank association, with stronger association at the larger coupling weight. This comparison is descriptive and does not establish equivalence of the two constraints.

\subsection{Synthetic Experiments}
\begin{figure}
    \centering\includegraphics[width=0.5\linewidth]{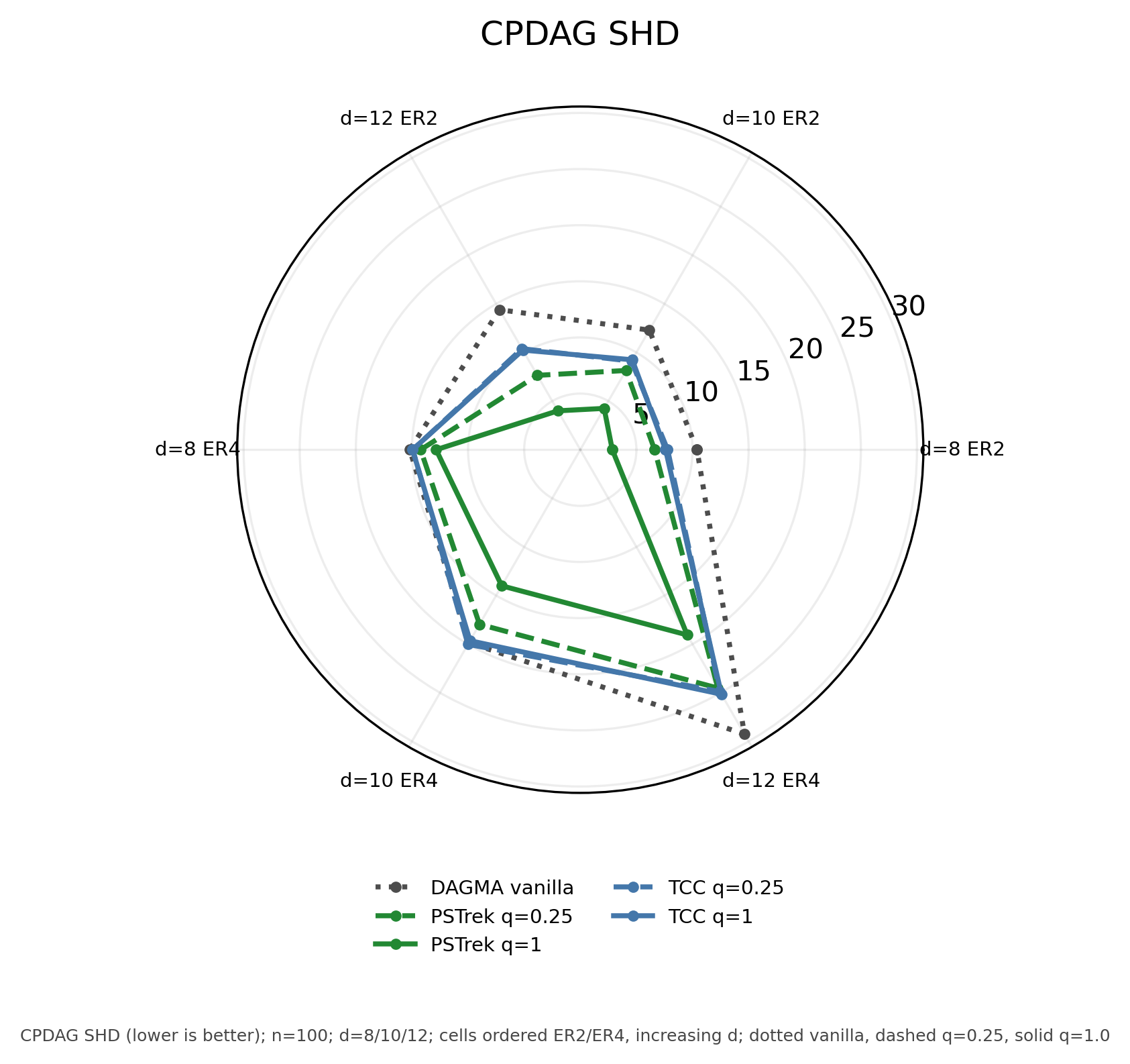}
    \caption{Structural recovery of DAGMA with TCC (blue) and PSTrek constraints (green).}
    \label{fig:pstrek_vs_tcc_radar_SHD}
\end{figure}

We ran synthetic experiments in the same framework as before, testing the spectral trek-cycle constraint (TCC) against the PSTrek extension of DAGMA \cref{fig:pstrek_vs_tcc_radar_SHD}. In the tested settings, TCC is comparable to or better than the vanilla baseline, but increasing the supplied NOTREKS information does not yield a consistent change in structural recovery. PSTrek has lower structural-recovery error and shorter runtime in this comparison.

\end{document}